\documentclass{article}

\usepackage[margin=1in]{geometry}
\usepackage[utf8]{inputenc} 
\usepackage[T1]{fontenc}    
\usepackage{hyperref}       
\usepackage{url}            
\usepackage{booktabs}       
\usepackage{amsfonts}       
\usepackage{nicefrac}       
\usepackage{microtype}      
\usepackage[dvipsnames]{xcolor}         
\usepackage{natbib}
\setcitestyle{authoryear,open={(},close={)}} 
\usepackage{mathrsfs}
\usepackage{bm}
\usepackage[linesnumbered,ruled,vlined]{algorithm2e}
\usepackage{amsmath,amssymb,amsfonts,mathtools}
\usepackage{amsthm}
\usepackage{hyperref}\hypersetup{colorlinks,linkcolor=red,anchorcolor=blue,citecolor=blue}
\usepackage{accents}
\usepackage{enumitem}
\usepackage{xfrac}

\allowdisplaybreaks

\makeatletter
\newtheorem*{rep@theorem}{\rep@title}
\newcommand{\newreptheorem}[2]{%
	\newenvironment{rep#1}[1]{%
		\def\rep@title{#2 \ref{##1}}%
		\begin{rep@theorem}}%
		{\end{rep@theorem}}}
\makeatother

\usepackage{color}
\definecolor{yc}{RGB}{255,69,0}
\definecolor{pv}{RGB}{0,102,204}

\newtheorem{theorem}{Theorem}
\newreptheorem{theorem}{Theorem}
\newtheorem{lemma}{Lemma}

\newreptheorem{definition}{Definition}
\newtheorem{assumption}{Assumption}
\newreptheorem{assumption}{Assumption}
\usepackage{subcaption}

\title{A Multi-Token Coordinate Descent Method for Semi-Decentralized Vertical Federated Learning}

\author{
Anonymous Author(s)\\
Affiliation
\footnote{Full affiliation}
}

\author{%
	Pedro Valdeira\footnote{Department of Electrical and Computer Engineering, Carnegie Mellon University; email: \texttt{pvaldeira@cmu.edu}.} \footnote{Institute for Systems and Robotics, Instituto Superior T\'{e}cnico.}\\
	CMU \& IST
	\and
	Yuejie Chi$^{*}$\\
	CMU
	\and Cl\'{a}udia Soares\footnote{Department of Computer Science, NOVA School of Science and Technology.}\\
	NOVA	
	\and Jo\~{a}o Xavier$^\dagger$\\
	IST
}

\date{September 2023; Revised \today}

\begin{document}

	\maketitle
	
	\footnotetext[4]{A preliminary version of this work was presented at the 2022 NeurIPS Workshop on Federated Learning: Recent Advances and New Challenges~\citep{valdeira2022a}.}
	\renewcommand*{\thefootnote}{\arabic{footnote}}
	
	\begin{abstract}
		
		Most federated learning (FL) methods use a client-server scheme, where clients communicate only with a central server. However, this scheme is prone to bandwidth bottlenecks at the server and has a single point of failure. In contrast, in a (fully) decentralized approach, clients communicate directly with each other, dispensing with the server and mitigating these issues. Yet, as the client network grows larger and sparser, the convergence of decentralized methods slows down, even failing to converge if the network is disconnected.
		This work addresses this gap between client-server and decentralized schemes, focusing on the vertical FL setup, where clients hold different features of the same samples.
		We propose multi-token coordinate descent (\texttt{MTCD}), a flexible semi-decentralized method for vertical FL that can exploit both client-server and client-client links. By selecting appropriate hyperparameters, \texttt{MTCD} recovers the client-sever and decentralized schemes as special cases---in fact, its decentralized instance is itself a novel method of independent interest.
		Yet, by controlling the degree of dependency on client-server links, \texttt{MTCD} can also explore a spectrum of schemes ranging from client-server to decentralized.
		We prove that, for sufficiently large batch sizes, \texttt{MTCD} converges at an $\mathcal{O}(1/T)$ rate for nonconvex objectives when the tokens roam across disjoint subsets of clients.
		To capture the aforementioned drawbacks of the client-server scheme succinctly, we model the relative impact of using client-server versus client-client links as the ratio of their ``costs'', which depends on the application.
		This allows us to demonstrate, both analytically and empirically, that by tuning the degree of dependency on the server, the semi-decentralized instances of \texttt{MTCD} can outperform both client-server and decentralized approaches across a range of applications.
	\end{abstract}
	
	\section{Introduction}
	\label{sec:introduction}
	
	In Federated Learning (FL), a set of clients collaborate to train a model without sharing their local data~\citep{McMahan2017}. Most FL literature focuses on horizontal FL, where the data is distributed by samples and every client holds the same features. However, in many applications of interest, the clients hold different blocks of features for the same set of samples: this setup is known as vertical FL~\citep{liu2022vertical}.
	
	This paper focuses on vertical FL (VFL), which first rose to prominence in cross-silo settings, where we have few clients,
	typically well-resourced and with reliable connectivity, such as companies or organizations. For example, WeBank collaborates with other companies with whom it shares customers to build a joint predictor for risk control~\citep{cheng2020federated}. For cross-silo VFL, the client-server scheme---where clients communicate with a central server, but not with each other---is often well-suited.
	
	Yet, VFL setups also arise in the cross-device settings, where we have a large number of clients, typically with limited computing power and unreliable connectivity.
	For example, in smart city applications, networks of devices collect diverse, sensor- or area-specific data, giving rise to feature-distributed datasets where each sample is a snapshot of these different features at a given timestamp. This has key applications in traffic monitoring and waste management~\citep{sharma2021machine}.
	However, such large numbers of clients can strain server bandwidth and cause latency, creating a bottleneck in training~\citep{Lian2017}. To mitigate this, decentralized approaches can be employed.
	
	
	Decentralized optimization, where clients communicate directly with one another, eliminating the need for a central server, has been widely studied. Earlier work was motivated by applications in wireless sensor networks and multi-agent control~\citep{Nedic2009,Duchi2012,mota2013d,Qu2018,Wu2018f}, while many recent efforts have been driven by FL~\citep{Koloskova2020,li2020communicationefficient,Zhao2022}. These methods avoid the bandwidth bottleneck and single point of failure of client-server setups. However, while they are well-suited for applications with well-connected networks or those lacking client-server links, they often converge slowly in sparse and large networks~\citep{Nedic2018}. Additionally, despite distributing the communication load across the network, they can incur high total communication costs.
	
	To achieve a lower communication cost than the more common consensus-based decentralized methods~\citep{Nedic2009,Duchi2012,mota2013d,Qu2018,Wu2018f,Koloskova2020},
	we can employ token methods~\citep{bertsekas1997,nedic2001,ram2009,johansson2010,Mao2020,Hendrikx2022}.
	%
	In token methods, a ``token'' carrying the current model estimate performs a random walk over a communication graph, undergoing local updates.
	This approach avoids the stale updates of consensus methods, yet it sacrifices their parallelism, which can lead to slower convergence in poorly connected networks~\citep{Hendrikx2022}. Multi-token methods~\citep{Ye2020,Chen2022,Hendrikx2022} navigate this trade-off by running multiple tokens simultaneously and periodically combining them, drawing from the idea that performing multiple random walks in parallel leads to a linear speed-up in cover time~\citep{alon2008many}.
	
\begin{figure*}[!t]
		\centering
		\subfloat[\label{fig:semi_decentralized}]{\includegraphics[width=0.38\textwidth]{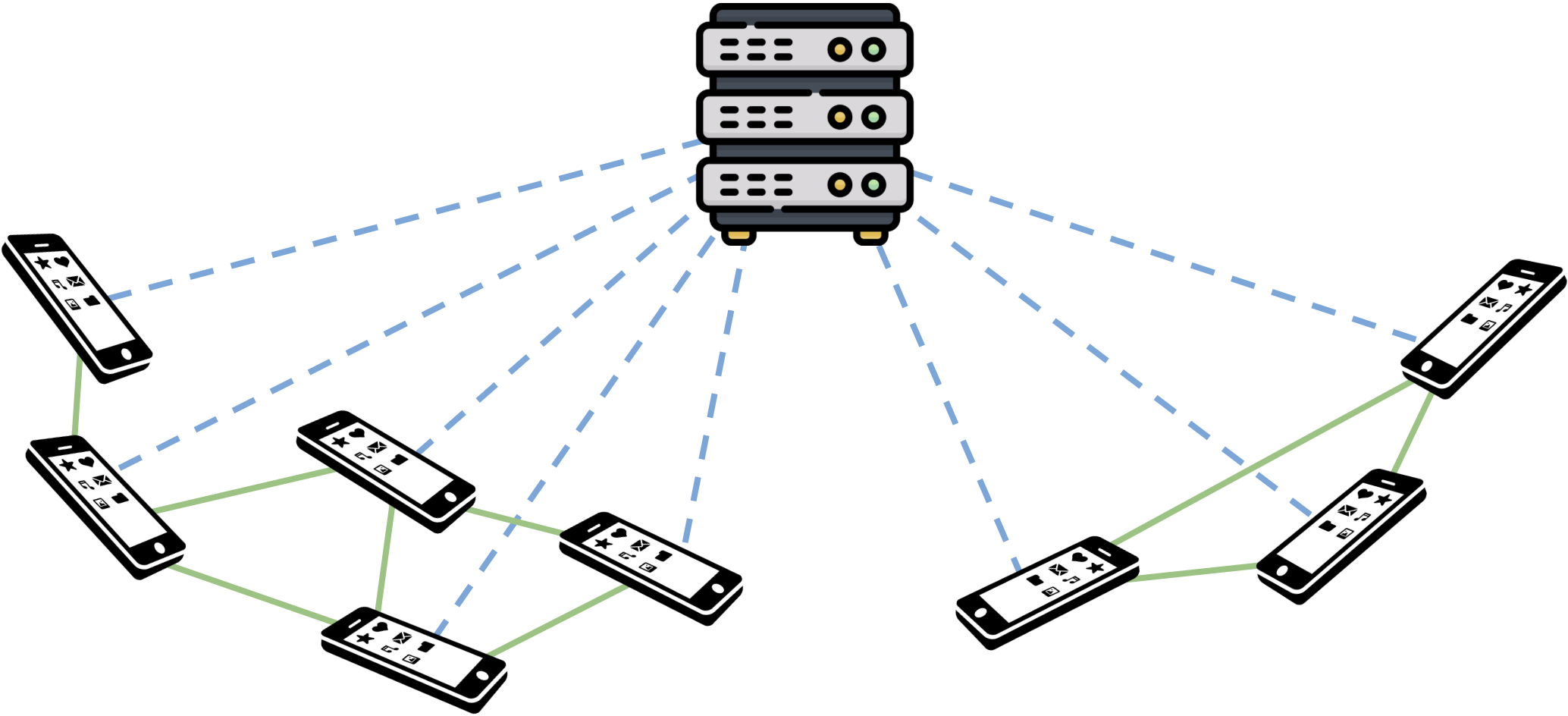}}
		\hfil
		\subfloat[\label{fig:splitNN}]{\includegraphics[width=0.4\textwidth]{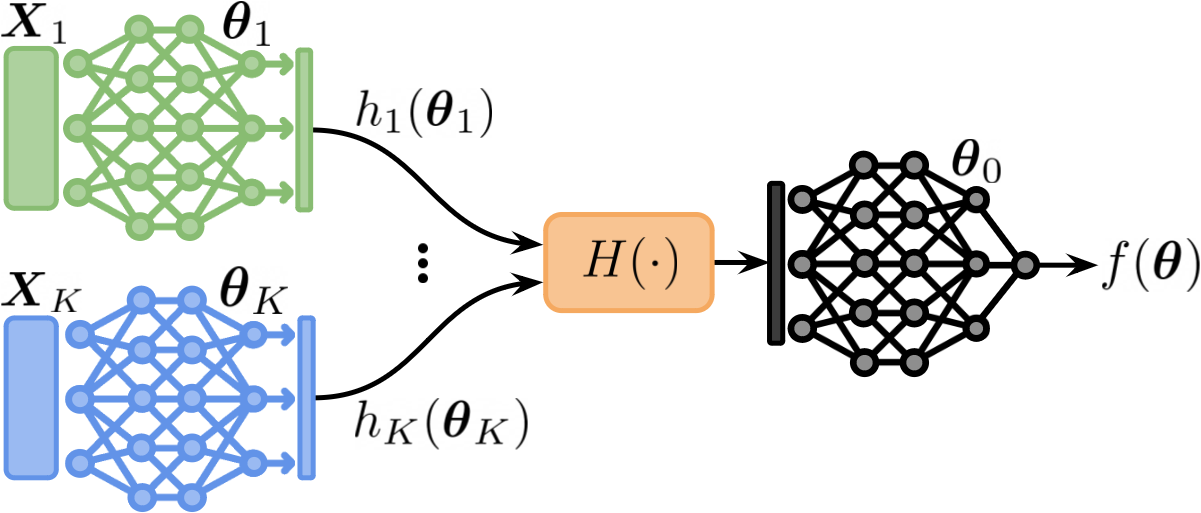}}
		\caption{
			On the left, we illustrate the semi-decentralized setup, where client-server communications are represented by dashed blue lines and client-client communications by solid green lines. On the right, we present a split neural network, where $K$ representations are obtained from neural networks, before an aggregation mechanism $H$ is applied and its result is inputted into a fusion neural network.
		}
		
	\end{figure*}

	To leverage the different strengths that make client-server and decentralized schemes well-suited for different applications, the semi-decentralized approach~\citep{Lin2021} employs both client-client and client-server links (See Figure~\ref{fig:semi_decentralized}). Client-client communications enable information flow without straining the limited bandwidth of the server, while client-server links accelerate convergence in poorly connected networks. In fact, even when the client-client communication graph is disconnected, as long as one client in each cluster can communicate with the server, convergence is still attainable.
	
	To the best of our knowledge, the semi-decentralized setup remains unexplored in VFL, prompting the following question:
	\begin{center}
		\emph{
			Can we design a flexible VFL method that encompasses both client‑server and decentralized schemes, and further operates along the spectrum between them to suit applications where neither extreme is ideal?
		}
	\end{center}
	%
	This paper answers the question affirmatively. Motivated by the observations above, we propose a multi-token coordinate descent (\texttt{MTCD}) method for semi-decentralized (SD) VFL. To the best of our knowledge, \texttt{MTCD} is both the first token method and the first SD method for VFL—where, unlike horizontal FL, the objective function does not decouple as a sum across clients—as well as the first SD multi-token method overall.
	
	\paragraph{Contributions.}
	In more detail, our main contributions are as follows.
	\begin{itemize}[leftmargin=1.0em]
		\itemsep0pt
		\item
		We introduce \texttt{MTCD}, a flexible VFL method that subsumes the client-server and decentralized schemes, and further extends it to the SD scheme. To the best of our knowledge, \texttt{MTCD} is the first token method and first SD method for VFL, as well as the first SD multi-token method overall.
		\item
		We show that, for sufficiently large batch sizes, \texttt{MTCD} converges at an $\mathcal{O}(1/T)$ rate for nonconvex objectives when the tokens roam over disjoint sets of clients. If the objective is further convex, we achieve the same rate while allowing for overlapping token trajectories.
		\item
		We show, both analytically and empirically, that by adjusting the degree of server dependence along the spectrum between client-server and decentralized schemes, \texttt{MTCD} can outperform both schemes across a range of applications.
	\end{itemize}
	
	\subsection{Related work}
	
	\paragraph{Vertical FL.}
	Some VFL works employ primal-dual optimization techniques, such as~\citet{smith2018cocoa}, which focuses on the client-server setting, \citet{He2018}, which generalizes it to the decentralized setting, and \texttt{DCPA}~\citep{Alghunaim2021}, which improves over the convergence rate of~\citet{He2018}. In contrast, some coordinate descent-based methods work in the primal domain, allowing them to train a broader class of models. The work in~\citet{Chen2020} employs asynchronous updates, \citet{Liu2022} resorts to multiple local updates between rounds of communication, \citet{Castiglia2022} introduces the use of lossy compression on the communicated representations, and~\citet{valdeira2025communication} employs error-feedback to speed up convergence of \citet{Castiglia2022}, matching the rate of uncompressed VFL.
	Recent works have extended VFL beyond the assumption that all samples are fully available during training and inference, namely, \citet{ganguli2024faulttolerantserverlessvfl} tackles inference in dynamic networks and~\citet{valdeira2025vertical} addresses missing features during both training and inference. An interesting line of work related to VFL is hybrid FL~\citep{zhang2021hybrid}, which deals with datasets distributed by both samples and features. For a detailed survey of VFL methods, see~\citet{liu2022vertical,yang2023survey}.
	
	\paragraph{Coordinate descent.}
	Coordinate descent methods~\citep{Wright2015}, where (blocks of) coordinates are updated sequentially, rather than simultaneously, are natural candidates for optimization in feature-distributed learning. The block selection is most often cyclic~\citep{Beck2013} or independent and identically distributed at random~\citep{Nesterov2012,Richtrik2012}, yet \citet{Sun2019} considers block selection following a Markov chain.
	Several extensions to coordinate descent have been proposed, such as acceleration and parallelization~\citep{Fercoq2015} and adaptations to the distributed setting~\citep{Liu2022,Chen2022}.
	
	%
	
	\paragraph{Semi-decentralized FL.}
	Recently, SD approaches have been proposed to lower communication costs and deal with data heterogeneity~\citep{Lin2021,guo2022,wang2025communication}, and to handle intermittent connections, latency, and stragglers~\citep{Bian2022,yemini2023robust}. Additionally, other semi-decentralized FL works deal with (multi-layered) hierarchical networks~\citep{Zhang2021,Hosseinalipour2022}. Semi-decentralized FL is sometimes also referred to as hybrid FL; however, we opt for the term semi-decentralized FL to avoid confusion with the data partitioning setting mentioned above. The preliminary version of this work~\citep{valdeira2022a} addressed the special case of generalized linear models, did not provide convergence guarantees, and featured significantly less comprehensive numerical experiments.
	
	\paragraph{Outline.}
	Section~\ref{sec:problem_setup} introduces the problem setup. Section~\ref{sec:stcd} presents the instance of \texttt{MTCD} in the decentralized setting. Section~\ref{sec:sd-setting} extends this to the semi-decentralized setting and provides convergence guarantees. Section~\ref{sec:experiments} presents numerical experiments, demonstrating the performance of our method. Finally, Section~\ref{sec:conclusions} concludes the paper.
	
	\section{Problem setup} \label{sec:problem_setup}
	
	We consider a dataset $\bm{X}\in\mathbb{R}^{N\times d}$ with $N$ $d$-dimensional samples distributed by features across a set of clients $[K]\coloneqq\{1,\dots,K\}$. Client~$k\in[K]$ holds its local data $\bm{X}_k\in\mathbb{R}^{N\times d_k}$ and we have $\bm{X}=[\bm{X}_1,\cdots,\bm{X}_K]$. Note that
	$d_1+\dots+d_K=d$. We consider a broad class of machine learning models, known as split neural networks~\citep{ceballos2020splitnn}, illustrated in Figure~\ref{fig:splitNN}.
	
	In split neural networks, each client~$k$ has a local \textit{representation} model $\bm{h}_k\colon\bm{\Theta}_k\mapsto\mathcal{H}_k$, with parameters $\bm{\theta}_k\in\bm{\Theta}_k$, that extracts an (often lower-dimensional) representation of $\bm{X}_k$. These representations~$\bm{h}_k(\bm{\theta}_k)\coloneqq \bm{h}_k(\bm{\theta}_k;\bm{X}_{k})$ are then aggregated through a mechanism $H\colon \prod_{k=1}^K \mathcal{H}_k \mapsto \mathcal{H}$ to form $H(\{\bm{h}_k(\bm{\theta}_k)\})$, which is used as input to a \textit{fusion} model $\phi\colon\mathcal{H}\times \bm{\Theta}_0\mapsto\mathbb{R}$, parameterized by $\bm{\theta}_{0}\in\bm{\Theta}_0$. While other aggregation mechanisms can be used~\citep{ceballos2020splitnn}, we focus on aggregation by sum and by concatenation, as they are the most common and most general aggregators, respectively.
	Further, while $\bm{\theta}_k$ is associated with and updated at client~$k$, for $k\in[K]$, the fusion model parameters~$\bm{\theta}_0$ may be updated at different entities (clients or server), depending on the communication scheme.
	
	Split neural networks include, for example, generalized linear models, such as linear regression, logistic regression, and support vector machines, where $\bm{h}_k(\bm{\theta}_k)=\bm{X}_{k}\bm{\theta}_k$ for $k\in[K]$ and $\bm{\Theta}_0$ is an empty set.
	Let $f\colon\bm{\Theta}\mapsto\mathbb{R}$ denote our objective function, where
	$\bm{\Theta}\coloneqq \prod_{k=0}^K \bm{\Theta}_k$. We want to solve the following optimization problem:
	\begin{equation} \label{eq:glm}
		\min_{\bm{\theta}\in\bm{\Theta}}
		\;
		\left\{
		f(\bm{\theta})
		\coloneqq
		\ell \Big( \phi
		\left(
		H\big(\{\bm{h}_k(\bm{\theta}_k)\}\big) , \bm{\theta}_0
		\right) 
		\Big)
		\right\}
		,
	\end{equation}
	where the labels are included in the loss function~$\ell$. We assume that $\ell$, like $\phi$ and $H$, is known by all clients. This assumption, known as ``relaxed protocol''~\citep{liu2022vertical}, is often made in VFL~\citep{He2018,Alghunaim2021,Castiglia2022,valdeira2025communication}.\footnote{To ensure every client holds the labels, the server can broadcast them initially or we can start the token roaming from clients holding them.}
	
	Throughout the paper, we consider Problem~\eqref{eq:glm} and assume $f$ is $L$-smooth and has a finite infimum, as formalized below.
	\begin{assumption}[$L$-smoothness and finite infimum] \label{ass:L-smoothness}
		Function~$f\colon\mathbb{R}^d\mapsto\mathbb{R}$ is differentiable and there exists a constant $L\in(0,\infty)$ such that:
		\begin{equation} \label{eq:L-smoothness}
			\forall\;\bm{u},\bm{v}\in\mathbb{R}^d\colon\quad
			\lVert \nabla f(\bm{u}) - \nabla f(\bm{v}) \rVert
			\leq
			L
			\lVert \bm{u} - \bm{v}\rVert.
			\tag{A\ref*{ass:L-smoothness}}
		\end{equation}
		We assume $ f^\star \coloneqq \inf_{\bm{x}} f(\bm{x}) > -\infty $.
	\end{assumption}
	
	
	\section{The proposed method in the decentralized setting}
	\label{sec:stcd}
	
	We start by introducing a simple, special case of our algorithm, which we refer to as single-token coordinate descent (\texttt{STCD}). This method is also a subroutine of our general algorithm. Mathematically, \texttt{STCD} is closely related to~\citet{Sun2019} and the application mentioned therein taken from~\citet{Mao2020}. However, \texttt{STCD} works in the primal domain and on a feature-distributed setting.
	
	In this section, we do not require the existence of a server. We solve Problem~\eqref{eq:glm} in a fully decentralized manner, communicating through channels described by a graph $\mathcal{G}=(\mathcal{V}, \mathcal{E})$, where $\mathcal{V}\coloneqq[K]$ is the vertex set and $\mathcal{E}$ is the edge set. We denote the set of neighbors of client~$k$ by $\mathcal{N}_k\coloneqq \{i\colon \{i,k\} \in \mathcal{E}\}$ and define $\bar{\mathcal{N}}_{k} \coloneqq \mathcal{N}_{k}\cup \{k\}$. In this section only, $\bm{\theta}_0$ is associated with some client~$k$, which is responsible for updating it, as well as the parameters of its local model $\bm{\theta}_k$.\footnote{We do this for alignment with the analysis in~\citet{Sun2019}, where all blocks are selected following a Markov Chain. However, in practice, we may want to update $\bm{\theta}_0$ at each client instead, in which case the analysis would need to be adjusted.}
	
	Since all clients know $\phi$ and $\ell$, if a client knows $\mathcal{Z}\coloneqq\{ H\big(\{\bm{h}_k(\bm{\theta}_k)\}\big) , \bm{\theta}_0 \}$, it can compute $f$. We call $\mathcal{Z}$ our \textit{token}. The size of the token depends on the model being considered. For example, if we have an $E$-dimensional representation per sample, and thus $\mathcal{H}_k\subseteq\mathbb{R}^{NE}$, for all $k$, then aggregation by concatenation leads to a token~$\mathcal{Z}$ of size $K N E + \dim(\bm{\Theta}_0)$, where $\dim(\cdot)$ denotes the dimensionality of a space. Yet, for aggregation mechanisms of the form $H\colon \prod_k \mathbb{R}^a \mapsto\mathbb{R}^a$, such as the sum, the token size is independent of the number of clients. In this case, the token size only exceeds that of a single representation in that it includes the parameters of the fusion model, $\bm{\theta}_0$, which are typically small. (The fusion model is often a linear model, or even nonparameterized~\citep{liu2022vertical}.) In particular, for generalized linear models, $\mathcal{Z}=\{\bm{X}\bm{\theta}\}$ is of size $N$. {(In Section~\ref{sec:sd-setting}, we allow for the use of mini-batches, further dropping the dependency on the number of samples.)}
	
	A client holding $\mathcal{Z}$ can compute $f$. Yet, more importantly, if client~$k$ holds its local data~$\bm{X}_k$ and local parameters~$\bm{\theta}_k$, then holding $\mathcal{Z}$ enables it to compute the partial gradient with respect to $\bm{\theta}_k$, defined as $\nabla_k f(\bm{\theta}) \coloneqq \nabla_{\bm{\theta}_k} f(\bm{\theta})$ and given by:
	\[
	\frac{
		\partial \phi (H(\{\bm{h}_k(\bm{\theta}_k)\}),\bm{\theta}_0)
	}{
		\partial H(\{\bm{h}_k(\bm{\theta}_k)\})
	}
	\cdot
	\frac{\partial H(\{\bm{h}_k(\bm{\theta}_k)\})}{\partial \bm{h}_k(\bm{\theta}_k)}
	\cdot
	\frac{d \bm{h}_k(\bm{\theta}_k)}{d \bm{\theta}_k},
	\]
	where $\mathcal{Z}$ is used to compute the first two terms. Computing $\nabla_k f(\bm{\theta})$ allows client~$k$ to update its local model~$\bm{\theta}_k$.
	
	We now describe \texttt{STCD}, summarized in Algorithm~\ref{alg:stcd}. We index $\bm{\theta}$ and $\mathcal{Z}$ with two counters:
	\textbf{1)} step~$s$ of the token roaming, at which client~$k^s$ is visited, and \textbf{2)} local update~$q\in\{1,\dots,Q\}$ at client~$k^s$. To simplify the algorithm description, we omit $\bm{\theta}_0$ for the rest of Section~\ref{sec:stcd}, as if it were part of the local model of a given client~$k$ where it is updated. Yet, unlike $\bm{\theta}_0$ (which is part of the token), $\bm{\theta}_k$ does not leave $k$.
	
	\paragraph{1) Initialization.} Token $\mathcal{Z}^{s,q}$ must always be in accordance with iterate $\bm{\theta}^{s,q}$. Namely, as the roaming starts at client~$k^0$, this client must know~$\mathcal{Z}^{0,0}=\{ H(\{\bm{h}_k(\bm{\theta}^{0,0}_k)\}), \bm{\theta}^{0,0}_0 \}$. For some models, this can be achieved by initializing $\bm{\theta}^{0,0}_k$ such that $\bm{h}_k(\bm{\theta}^{0,0}_k)$ is independent of the local data~$\bm{X}_k$. When this is not possible, the clients can send their initial representations to $k^0$ as a prelude.
	
	\paragraph{2) Updating the token and the local models.} As explained above, client~$k^s$ can compute the partial gradient with respect to its local model locally, allowing it to perform a coordinate descent step. Further, to lower communication costs, we do $Q$ local steps at each client. That is, for $q=0,\dots,Q-1$:
	\begin{equation} \label{eq:cd_step}
		\bm{\theta}_{k^s}^{s,q+1}
		=
		\bm{\theta}_{k^s}^{s,q} - \eta
		\nabla_{k^s} f(\bm{\theta}^{s,q})
	\end{equation}
	and $\bm{\theta}_k^{s,q+1}=\bm{\theta}_k^{s,q}$ for $k\not=k^s$. We must now update the token accordingly. To compute $\mathcal{Z}^{s,q+1}$, we need $\mathcal{Z}^{s,q}$, $\bm{h}_{k^s}(\bm{\theta}_{k^s}^{s,q+1})$, and $\bm{h}_{k^s}(\bm{\theta}_{k^s}^{s,q})$, all of which are held by $k^s$. For example, for aggregation by sum, let $H^{s,q}\coloneqq H(\{\bm{h}_k(\bm{\theta}^{s,q}_k)\})$, we have:
	\[
	H^{s,q+1}
	=
	H^{s,q} +\bm{h}_{k^s}(\bm{\theta}^{s,q+1}_{k^s})-\bm{h}_{k^s}(\bm{\theta}^{s,q}_{k^s}).
	\]
	This allows us to perform multiple local steps. After these steps, the updated token is communicated to client~$k^{s+1}$. Thus, by induction, we keep the token up-to-date throughout our algorithm.
	
	\paragraph{3) Communicating the token.}
	Client~$k^s$ sends token $\mathcal{Z}^{s,Q}$ to client $k^{s+1}\sim\mathcal{U}( \bar{\mathcal{N}}_{k^s})$, where $\mathcal{U}$ denotes the uniform distribution. Thus, the roaming token performs a random walk over the communication graph, resulting in a Markov chain.
	
	\IncMargin{1.0em}
	\begin{algorithm}[t]
		\DontPrintSemicolon
		\textbf{Input:} initial $\bm{\theta}^{0,0}$ and $k^0$, stepsize~$\eta$, number of hops~$S$, number of local updates~$Q$.\;
		$\mathcal{Z}^{0,0}\gets\{ H(\{\bm{h}_k(\bm{\theta}^{0,0}_k),\}), \bm{\theta}^{0,0}_0 \}$.\;
		\SetKwFunction{TRoam}{TokenRoaming}
		$\bm{\theta}^{S,Q},\mathcal{Z}^{S,Q},k^{S}\gets$ \TRoam{$\bm{\theta}^{0,0},\mathcal{Z}^{0,0},k^0,\bm{X},S,Q$}.\; 
		\SetKwProg{Fn}{Function}{:}{}
		\Fn{\TRoam{$\bm{\theta},\mathcal{Z},k,\bm{D}$,S,Q}}{
			$\bm{\theta}^{0,0},\mathcal{Z}^{0,0}, k^0 \gets \bm{\theta},\mathcal{Z},k$.\;
			\For{$s=0,\dots,S-1$}{
				\For{$q=0,\dots,Q-1$ \textbf{\emph{in client}~$k^s$}}{
					Compute $\bm{\theta}_{k^s}^{s,q+1}$ via~\eqref{eq:cd_step}.\;
					Compute $\mathcal{Z}^{s,q+1}$.\;
				}
				Client~$k^s$ sends $\mathcal{Z}^{s,Q}$ to  $k^{s+1}\sim\mathcal{U}( \bar{\mathcal{N}}_{k^s})$.\;
				$\bm{\theta}^{s+1,0},\mathcal{Z}^{s+1,0} \gets \bm{\theta}^{s,Q},\mathcal{Z}^{s,Q}$.\;
			}
			\KwRet $\bm{\theta}^{S,Q},\mathcal{Z}^{S,Q},k^{S}$.\;
		}
		\caption{
			\texttt{STCD}
		}
		\label{alg:stcd}
	\end{algorithm}
	\DecMargin{1.0em}
	
	To summarize, \texttt{STCD} is a decentralized VFL method that allows us to employ an extension of Markov chain block coordinate descent~\citep{Sun2019} with multiple local updates in feature-distributed settings.
	
	We now discuss the convergence of \texttt{STCD}. We define an auxiliary scalar iteration counter to summarize the tuple $(S,Q)$ used in the description of \texttt{STCD},  $r \coloneqq sQ + q$. This allows us to map \texttt{STCD} to the optimization framework of~\citet{Sun2019}, thereby enabling us to resort to their convergence results to establish the convergence of \texttt{STCD}.
	
	\paragraph{Convergence guarantees.}
	If $f$ is an $L$-smooth function with a finite infimum~\eqref{eq:L-smoothness} and $\{\bm{\theta}^{i}\}^{r}_{i=1}$ is a sequence generated by Algorithm~\ref{alg:stcd}, \citet{Sun2019} gives convergence guarantees for $Q=1$. In particular, under mild assumptions on the Markov chain---for example, being time-homogeneous, irreducible, and aperiodic---we have that $\lim_{r\to\infty}\mathbb{E}\lVert \nabla f(\bm{\theta}^r) \rVert=0$ and
	\begin{equation} \label{eq:mcbcd_bound}
		\mathbb{E}\left[ \min_{i\in[r]}\lVert \nabla f(\bm{\theta}^i) \rVert^2\right]
		\leq
		\frac{(\Omega_1 (\tau-1)^2 + \Omega_2) \Delta}{r},
	\end{equation}
	where $\Delta\coloneqq f \left( \bm{\theta}^{0} \right)-f^\star$, $\Omega_1$ and $\Omega_2$ are constants that depend on the minimum value of the stationary distribution of the Markov chain, $\pi_{\min}$, and $\tau$ denotes the $\frac{\pi_{\min}}{2}$-mixing time of the the Markov chain.
	Although \citet{Sun2019} only considers $Q=1$, we can leverage its analysis to prove convergence for $Q>1$.
	
	Recalling that $\mathcal{G}=(\mathcal{V}, \mathcal{E})$ is the original graph, we consider an auxiliary dynamic graph $\mathcal{G}^r=(\mathcal{V},\mathcal{E}^r)$, where 
	\[
	\mathcal{E}^r
	=
	\begin{cases}
		\mathcal{E} \quad & \text{if } r\text{ mod } Q=0,\\
		\{ \{i,i\}:i \in \mathcal{V} \} & \text{otherwise}.
	\end{cases}
	\]
	Then, running \texttt{STCD} on $\mathcal{G}^r$ with a single local update is equivalent to running \texttt{STCD} on the original graph~$\mathcal{G}$ with $Q$ local updates. Further, this dynamic graph preserves the properties required for the analysis to hold.
	
	To see this, let $\bm{P}$ denote the transition matrix of a random walk over the original graph~$\mathcal{G}$ and let $\bm{P}(r)$ denote the transition matrix of a random walk over the auxiliary graph~$\mathcal{G}^r$. Note that, for $r\mod Q\not=0$, $\bm{P}(r)=\bm{I}$, where $\bm{I}$ denotes the identity matrix, and, for $r\mod Q=0$, $\bm{P}(r)=\bm{P}$.
	At a given iteration~$R$, assumed, for simplicity, to be $R\geq Q$, we have that $\bm{P}(r)\bm{P}(r+1) \ldots \bm{P}(r+R)=\bm{P}^{\lfloor R/Q \rfloor}$. We thus recover the results in~\citet{Sun2019} up to a factor of $Q$ in the mixing time. (That is, in~\eqref{eq:mcbcd_bound}, we replace $\tau$ by $Q\tau$.)
	
	\paragraph{Limitations.}
	The decentralized, single-token method described in this section is appealingly simple and outperforms the baselines across various setups (see Section~\ref{sec:experiments}). However, its convergence slows down as network connectivity decreases. To address this, we introduce \texttt{MTCD}, a multi-token extension to \texttt{STCD} that mitigates this problem.
	
	
	\section{The proposed method in the semi-decentralized setting}
	\label{sec:sd-setting}
	
	In Section~\ref{sec:stcd}, we introduced a (fully) decentralized special case of \texttt{MTCD} where a single token roams over a set of clients. We now present our method in the semi-decentralized setting, which subsumes the setting in Section~\ref{sec:stcd} as a special case. \texttt{MTCD} alternates between a \emph{roaming} step and a \emph{syncing} step. 
	\begin{itemize}[leftmargin=1.0em]
		\itemsep0pt
		\item \textit{Roaming.} We start with multiple, matching tokens at a subset of the clients. As each token performs a different random walk realization, they undergo different sequences of updates, becoming distinct.
		
		\item \textit{Syncing.} To leverage these parallel computations while keeping our model estimates coupled, we periodically sync the roaming tokens at the server, combining the progress of multiple sequences.
	\end{itemize}
	
	By adjusting the number of tokens and syncing frequency, we achieve a flexible degree of parallelization and server dependency. This enables a smooth navigation along the spectrum between client-server and decentralized setups, allowing for communication schemes suitable for different applications.
	
	We further extend \texttt{STCD} to allow for mini-batch estimates. More precisely, if our objective is a finite sum $f(\bm{\theta})=\frac{1}{N}\sum_{n=1}^{N}f_n(\bm{\theta})$, where $f_n(\bm{\theta})$ depends only on sample $n$, and $\mathcal{B}\subseteq[N]$ denotes the indices of a mini-batch of size $B$, then our mini-batch gradient estimate is $ \tilde{\nabla} f (\bm{\theta};\mathcal{B}) \coloneqq \frac{1}{B}\sum_{n\in\mathcal{B}}\nabla f_n(\bm{\theta}) $. We make the following standard assumptions on our mini-batch gradient estimate.
	
	\begin{assumption}[Unbiased]
		\label{ass:unbiased}
		The mini-batch gradient estimate is unbiased. That is, for all $ \bm{\theta} \in \bm{\Theta} $:
		\begin{equation} \label{eq:unbiased}
			{\textstyle
				\mathbb{E} \left[\tilde{\nabla}f(\bm{\theta};\mathcal{B})\right]
				=
				\nabla f(\bm{\theta}),
			}
			\tag{A\ref*{ass:unbiased}}
		\end{equation}
		where the expectation is with respect to mini-batch~$\mathcal{B}$.
	\end{assumption}
	
	\begin{assumption}[Bounded variance]
		\label{ass:bounded_var}
		There exists a positive constant $\sigma$ such that,
		let $B\coloneqq|\mathcal{B}|$,
		we have that, for all $ \bm{\theta} \in \bm{\Theta} $:
		\begin{equation} \label{eq:bounded_var}
			\mathbb{E}
			\left\lVert \tilde{\nabla}f(\bm{\theta};\mathcal{B})-\nabla f(\bm{\theta}) \right\rVert^2
			\leq
			\frac{\sigma^2}{B},
			\tag{A\ref*{ass:bounded_var}}
		\end{equation}
		where the expectation is with respect to mini-batch~$\mathcal{B}$.
	\end{assumption}
	
	In addition to the communication graph $\mathcal{G}$, we now also consider the existence of a central server with links to all clients, as illustrated in Figure~\ref{fig:semi_decentralized}. The existence of a server brings a change to the model partitioning: $\bm{\theta}_{0}$ is now updated at the server. Further, we now have $\Gamma$ tokens roaming simultaneously. Each token $\mathcal{Z}_\gamma$, for $\gamma\in[\Gamma]$, has a model estimate $\bm{\theta}(\gamma)$ associated with it. Thus, during the roaming step, each client~$k$ must now keep up to $\Gamma$ local model estimates $\bm{\theta}_k(\gamma)$ in memory.\footnote{In practice, it suffices to to add a copy of $\bm{\theta}^{t}$ as a new token visits during a given roaming step (with a maximum of $\Gamma$ model estimates at a client), resetting the number of copies when syncing.} (This requirement holds in the general case, but not in the important case discussed below, where the tokens roam over disjoint subsets of clients.) To simplify the description of the algorithm, we consider a token $\gamma=0$ which stays at the server ($k=0$) throughout the roaming step.
	
	We now describe \texttt{MTCD}, summarized in Algorithm~\ref{alg:MTCD}, where $k_\gamma^{t,s}$ denotes the client holding token $\gamma$ after $t$ synchronizations and $s$ hops and $P_\gamma$ denotes the distribution over clients indicating where the roaming begins. (Note that, unlike in \texttt{STCD} where $S$ was the iteration counter, in \texttt{MTCD} it is a fixed hyperparameter.) We index $\bm{\theta}$ and $\mathcal{Z}$ with three counters: synchronization step~$t$, and the two counters used in \texttt{STCD}. For simplicity, we write $\mathcal{Z}^{t}\coloneqq\mathcal{Z}^{t,0,0}_1=\dots=\mathcal{Z}^{t,0,0}_\Gamma$ and $\bm{\theta}^{t}\coloneqq\bm{\theta}^{t,0,0}(1)=\dots=\bm{\theta}^{t,0,0}(\Gamma)$. Lastly, we let $\bm{h}_{k,\mathcal{B}^{t}}(\bm{\theta}^{t}_k)$ denote the mini-batch representations of client~$k$ at time $t$.
	
	\IncMargin{1.0em}
	\begin{algorithm}[t]
		\DontPrintSemicolon
		\textbf{Input:} initial $\bm{\theta}^{0}$, stepsize~$\eta$, number of hops~$S$, number of local updates~$Q$, model estimates combination weights $\{w_{k\gamma}\}$, distributions $\{P_\gamma\}$.\;
		\For{$t=0,\dots,T-1$}{
			\For{$k\in[K]$ \emph{\textbf{in parallel}}}{
				Sample shared mini-batch indices $\mathcal{B}^{t}\subseteq[N]$ via a seed shared.\;
				Send $\bm{h}_{k,\mathcal{B}^{t}}(\bm{\theta}^{t}_k)$ to the server.\;
			}
			Server computes $\mathcal{Z}^{t}\gets\{ H(\{\bm{h}_{k,\mathcal{B}^{t}}(\bm{\theta}^{t}_k)\}), \bm{\theta}^{t}_0 \}$.\;
			\For{$\gamma\in[\Gamma]\cup\{0\}$ \emph{\textbf{in parallel}}}{
				Server sends $\mathcal{Z}^{t}$ to $k^{t,0}_\gamma\sim P_\gamma$.\;
				\SetKwFunction{TRoam}{TokenRoaming}
				$\bm{\theta}^{t,S,Q}(\gamma),\mathcal{Z}^{t,S,Q}_\gamma,k^{t,S}_\gamma\gets$ \TRoam{$\bm{\theta}^{t},\mathcal{Z}^{t},k^{t,0}_\gamma,\bm{B}^t,S,Q$}.\;
			}
			\For{$k\in[K]\cup\{0\}$ \emph{\textbf{in parallel}}}{
				$\bm{\theta}^{t+1}_k=\sum_{\gamma=1}^{\Gamma}w_{k\gamma}\bm{\theta}^{t,S,Q}_{ k}(\gamma)$.\;
			}
		}
		\caption{\texttt{MTCD}}
		\label{alg:MTCD}
	\end{algorithm}
	\DecMargin{1.0em}
	
	\paragraph{1) Initialization.}
	All model-token pairs $(\bm{\theta}(\gamma), \mathcal{Z}_\gamma)$ are initialized to the same values. As in \texttt{STCD}, each token~$\mathcal{Z}_\gamma$ must be in accordance with $\bm{\theta}(\gamma)$.
	
	\paragraph{2) Roaming.}
	When $\mathcal{Z}_\gamma$ is at client $k^{t,s}_\gamma$, it is used to update the block~$k^{t,s}_\gamma$ of parameters~$\bm{\theta}(\gamma)$. Each $\mathcal{Z}_\gamma$ roams for $S$ hops, as in \texttt{STCD}. In parallel, $\bm{\theta}_{0}(0)$ is updated at the server.
	
	\paragraph{3) Syncing.}
	After $S$ hops, each client combines its model estimates, obtaining $\bm{\theta}^{t+1}_k=\sum_{\gamma=0}^{\Gamma}w_{k\gamma}\bm{\theta}^{t,S,Q}_{k}(\gamma)$. (We cover the choice of $\bm{w}_k\coloneqq(w_{k0},\dots,w_{k\Gamma})$, which lies in the $\Gamma$-dimensional probability simplex, later.)
	At the start of the next iteration, each client $k$ samples a set of indices $\mathcal{B}^t$ locally, which is the same for all clients (as they share a seed), and sends its local representation $\bm{h}_{k,\mathcal{B}^{t}}(\bm{\theta}^{t}_k)$ to the server. This allows the server to compute $\mathcal{Z}^t$ and send it to each client~$k^{t,0}_\gamma\sim P_\gamma$, for $\gamma\in[\Gamma]$, to start the next roaming step, with $P_\gamma$ such that $\mathbb{P}(k^{t,0}_\gamma=k)$. (The point mass distribution $P_0$ has support over the server ($k=0$), a node without neighbors.)
	
	\texttt{MTCD} can recover client-server and decentralized setups:
	\begin{itemize}[leftmargin=1.0em]
		
		\item
		If client-server links are not used ($S\to\infty$), \texttt{MTCD} recovers the decentralized setup. Namely, this corresponds to running $\Gamma$ instances of \texttt{STCD} in parallel.
		
		\item
		If the edge set $\mathcal{E}$ is empty, we recover the client-server setup. If $\Gamma=K$ and each $P_\gamma$ has support over a different client, all clients are updated at each step~$t$, whereas if $\Gamma<K$, only a subset of clients is updated at each step~$t$.
	\end{itemize}
	
	\paragraph{On client-server communications.}
	\texttt{MTCD} employs client-server communications twice per iteration $t$: first during the uplink phase (line 5), when each client $k$ sends its representation $\bm{h}_{k,\mathcal{B}^{t}}(\bm{\theta}^{t}_k)$ to the server, and later during the downlink phase (line 8), when the server transmits $\Gamma$ tokens to a subset of clients to initiate the next roaming step.
	
	This scheme entails only $\Gamma$ downlink communications, yet it requires $K$ uplink communications. This contrasts with horizontal FL, where often only a subset of clients participates in both directions. However, in VFL, it is standard for all clients to participate in every round, since computing the loss for even a single sample requires input from every client. An exception to this is the special full-batch case, where only $\Gamma$ uplink communications are needed, as the server can reuse representations from previous iterations to update the token. In this scenario, only clients updating their local model in a given iteration must send new representations to the server.
	
	However, in the general mini-batch case, even if only a subset of clients updates their local models, all clients must transmit their mini-batch representations to the server in the subsequent iteration. This is because each mini-batch representation is associated with both a specific mini-batch and local parameters, and these parameters have generally been updated since the last time that mini-batch was sampled.
	
	
	\paragraph{On the impact of $S$ and $\Gamma$.}
	In general, as we increase the amount of parallel computations (by increasing $\Gamma$) and client-server communications (by lowering $S$), the communication efficiency and the number of iterations needed to converge both decrease. Given this trade-off, we see that our choice of $S$ and $\Gamma$ depends on the application.
	
	Having introduced the general \texttt{MTCD} algorithm, we now go over two instances of it, both for semi-decentralized setups, and provide convergence guarantees for both. Since our convergence results must accommodate $\Gamma > 1$, where different tokens sync only every $S$ hops, the \texttt{STCD} analysis from the previous section does not apply, as it requires $S \to \infty$.\footnote{Note how, for $S \to \infty$, the sequence of iterates in \texttt{MTCD}, $\{\bm{\theta}^t\}$, would consist of a single element, preventing convergence.} Thus, our convergence results for \texttt{MTCD} differ fundamentally from those for $S \to \infty$ and $\Gamma = 1$, as they require a new approach to the proof, addressing the challenges of \textbf{1)} stale information from other tokens and \textbf{2)} the coexistence and periodic combination of different model estimates.
	
	We defer the proofs
	to the appendices.
	
	\subsection{Setting with a token per cluster}
	
	\paragraph{Setup.}
	Consider $C$ disjoint clusters of clients $\{\mathcal{C}_c,\dots,\mathcal{C}_C\}$ and $\Gamma=C$ tokens. Each token $\gamma$ roams $\mathcal{C}_\gamma$, allowing us to use $\gamma$ and $c$ interchangeably. Note that distribution~$P_\gamma$ now has support over $\mathcal{C}_\gamma$, and only over $\mathcal{C}_\gamma$. In this setting, the subsets of the blocks~$\{\bm{\theta}_k\}$ updated by different tokens are disjoint. This allows us to combine the model estimates by simply concatenating the updated blocks. That is, we let $\bm{w}_{k}\in\mathbb{R}^\Gamma$ be the one-hot encoding for the token visiting client~$k$. We also define a cluster containing only the server, $\mathcal{C}_0\coloneqq\{0\}$.
	
	Clusters $\{\mathcal{C}_c\}$ may reflect the natural topology of the communication graph, arising from physical limitations or privacy constraints. For example, a cluster may represent devices within a single household or company. However, we can also partition the graph artificially before training to enable the use of multiple tokens in this token-per-cluster setting, which prevents overlapping trajectories.
	
	The selection of a graph clustering method is beyond the scope of this work, but it is worth noting that there exists an extensive literature addressing this topic. This field, sometimes referred to as community detection, focuses on grouping nodes so that connectivity within the clusters is greater than across clusters, and it includes distributed methods that do not require a centralized orchestrator~\citep{hui2007distributed}.
	
	\paragraph{Convergence results.}
	We denote the number of stale updates between syncing steps by $P\coloneqq SQ$ and the client holding token $c$ at update $p\in[P]$ of iteration $t$ by $i_c^{t,p}$, where $i_c^{t,p}=k_c^{t,s}$ for all $p\in \{sQ, \dots, s(Q + 1) - 1\}$. We define $\bm{p}^t\coloneqq(\bm{p}_1^t,\dots,\bm{p}_C^t)$ where $\bm{p}_c^t=(i_c^{t,0},\dots,i_c^{t,P-1})$, and resort to the set:
	\[
	\mathcal{F}^t\coloneqq
	\{
	\mathcal{B}^0,\bm{p}^0,\dots,\mathcal{B}^{t-1},\bm{p}^{t-1}
	\}
	.
	\]
	We denote the partition across clusters as $ \bm{\theta}=(\bm{\theta}_{1,:},\dots,\bm{\theta}_{C,:}) $, and each partition $\bm{\theta}_{c,:}$ is further partitioned into $K_c\coloneqq|\mathcal{C}_c|$ blocks, $ \bm{\theta}_{c,:}=(\bm{\theta}_{c,1},\dots,\bm{\theta}_{c,K_c}) $.
	%
	Further, in this section, we define $\bm{\theta}^{t,p}(c)$ to refer to the iterate at cluster~$c$, at stale update~$p$ of iteration~$t$.
	We use the following shorthand notation:
	\[
	\mathbb{E}_t[ \cdot ] \coloneqq\mathbb{E}\left[\cdot \mid \mathcal{F}^t\right],
	\quad
	\mathbb{E}_{t^+}[ \cdot ] \coloneqq\mathbb{E}\left[\cdot \mid \mathcal{F}^t, \bm{p}^t\right].
	\]
	Let $\bm{I}$ denote the identity matrix and define $\bm{\Pi}_{c,i}$ as the diagonal matrix whose diagonal entries are $\bm{I}$ for indices corresponding to the partition $\bm{\theta}_{c,i}$ and $0$ elsewhere. It will also be useful to define $\bm{\Pi}_{c}\coloneqq\sum_{i=1}^{K_c}\bm{\Pi}_{c,i}$ and $ \lVert \bm{u} \rVert^2_{\bm{A}}\coloneqq \bm{u}^\top\bm{A}\bm{u}$.
	
	We now present Lemma~\ref{lemma:inner_product_bound} and Lemma~\ref{lemma:norm_bound}, which we use to prove Theorem~\ref{thm:mtcd_tpc_convergence} and Theorem~\ref{thm:mtcd_convergence}.
	\begin{lemma}[Inner product upper bound] \label{lemma:inner_product_bound}
		Let $\{\bm{\theta}^t\}$ be a sequence generated by Algorithm~\ref{alg:MTCD}. If $f$ is $L$-smooth~\eqref{eq:L-smoothness} and our mini-batch gradient estimate is unbiased~\eqref{eq:unbiased}, we have that:
		\begin{equation} \label{eq:inner_product_bound}
			\begin{split}
				&\mathbb{E}_{t^+}
				\left[\nabla_{c,:} f(\bm{\theta}^{t})^\top(\bm{\theta}_{c,:}^{t+1}-\bm{\theta}_{c,:}^{t}) \right]
				\\
				&\leq
				-\eta\left(1-\frac{\eta^2L^2(P-1)^2}{2}\right)
				\lVert\nabla f(\bm{\theta}^{t})\rVert_{\bm{\Pi}_{c,i_c^{t,0}}}^{2}
				\\
				&\quad-
				\frac{\eta}{2} \lVert \nabla f(\bm{\theta}^{t}) \rVert_{\sum_{p=1}^{P-1}\bm{\Pi}_{c,i_c^{t,p}}}^{2}\\
				&\quad
				{
					-
					\frac{\eta}{2}
					(1-\eta^2L^2(P-1)^2)
					\sum_{p=1}^{P-1} \lVert \nabla f(\bm{\theta}^{t,p}(c)) \rVert_{\bm{\Pi}_{c,i_c^{t,p}}}^{2}.
				}
			\end{split}
		\end{equation}
	\end{lemma}
	
	
	\begin{lemma}[Squared norm upper bound] \label{lemma:norm_bound}
		Let $\{\bm{\theta}^t\}$ be a sequence generated by Algorithm~\ref{alg:MTCD}. If our mini-batch gradient estimate is unbiased~\eqref{eq:unbiased} and has a bounded variance~\eqref{eq:bounded_var}, we have that:
		\begin{equation} \label{eq:norm_bound}
			\begin{split}
				&\mathbb{E}_{t^+}
				\left[ ||\bm{\theta}_{c,:}^{t+1}-\bm{\theta}_{c,:}^{t}||^2 \right]
				\leq
				\frac{\eta^2\sigma^2P^2}{B}
				\\
				&\quad
				{
					+
					\eta^2P
					\Big(
					\lVert \nabla f(\bm{\theta}^{t}) \rVert_{\bm{\Pi}_{c,i_c^{t,0}}}^2
					+
					\sum_{p=1}^{P-1}
					\lVert \nabla f(\bm{\theta}^{t,p}(c)) \rVert_{\bm{\Pi}_{c,i_c^{t,p}}}^2
					\Big)
					.
				}
			\end{split}
		\end{equation}
	\end{lemma}
	
	
	{We now make a general assumption on the minimum probability of each client being visited by a token.
		\begin{assumption}[Minimum probability of visit]
			\label{ass:min_prob_token_visit}
			There exists a positive constant $\pi$ such that, for all $c\in[C]$ and $j\in[K_c]$:
			\begin{equation} \label{eq:min_prob_token_visit}
				{\textstyle
					\mathbb{P}\left(\bigcup_{p=0}^{P-1} \{i_c^{t,p}=j\}\right) \geq \pi > 0
					\quad \text{for all} \quad
					t\geq0
					.
				}
				\tag{A\ref*{ass:min_prob_token_visit}}
			\end{equation}
		\end{assumption}
		This can be achieved through a variety of communication schemes. For example, if \textbf{1)} all the clients have a nonzero probability of receiving the token from the server, then this holds for any $P\geq1$, and if \textbf{2)} at least one node in the cluster has a nonzero probability of receiving the token from the server, then this holds if $S$ is greater than or equal to the diameter of the cluster (that is, the greatest distance between any pair of nodes in a graph). In Appendix~\ref{app:lb-pi}, we go over this topic in more detail and provide multiple ways to lower bound $\pi$ for different communication schemes.}
	
	{\begin{theorem}[Main result---token-per-cluster setting] \label{thm:mtcd_tpc_convergence}
			Let $\{\bm{\theta}^t\}$ be a sequence generated by Algorithm~\ref{alg:MTCD} in the setting with a token per cluster. If $f$ is an $L$-smooth function with a finite infimum~\eqref{eq:L-smoothness}, our mini-batch gradient estimate is unbiased~\eqref{eq:unbiased} and has a bounded variance~\eqref{eq:bounded_var}, and each node has a minimum probability $\pi>0$ of being visited by a token at each iteration~\eqref{eq:min_prob_token_visit}, we have that, for a stepsize $\eta \in \big(0,\frac{1}{2LSQ}\big]$:
			\begin{equation} \label{eq:main_theorem}
				\frac{1}{T}
				\sum_{t=0}^{T-1}
				\mathbb{E} \lVert \nabla f(\bm{\theta}^{t}) \rVert_{}^{2}
				\leq
				\frac{2\Delta}{\eta\pi T}
				+\frac{\eta S^2Q^2\sigma^2L}{\pi B },
			\end{equation}
			where $\Delta\coloneqq f(\bm{\theta}^{0}) - f^\star$ and the expectation is with respect to token trajectories $\{\bm{p}^t\}$ and mini-batches $\{\mathcal{B}^t\}$. For the special, full-batch case, we also have that $\nabla f \left( \bm{\theta}^{t}\right)\to0$ almost surely.
	\end{theorem}}
	
	
	\paragraph{Proof sketch.}
	To prove Theorem~\ref{thm:mtcd_tpc_convergence}, we first employ the quadratic upper bound that follows from $L$-smoothness at iterates $t$ and $t+1$. Next, we decouple this bound into cluster specific terms, leveraging the fact that the blocks updated at different tokens are disjoint. We then take the conditional expectation $\mathbb{E}_{t^+}$ and use Lemma~\ref{lemma:inner_product_bound} and Lemma~\ref{lemma:norm_bound}. Next, setting $\eta \in \big(0,\frac{1}{2LP}\big]$ allows us to drop one of the terms in the upper bound. We then take the conditional expectation $\mathbb{E}_{t}$ and use our assumption on the existence of a minimum probability of token visit $\pi>0$ to arrive at a descent lemma (in expectation). This allows us to use the classic result due to Robbins and Siegmund~\citep{robbins1971convergence} to get that $\nabla f \left( \bm{x}^{t}\right)\to0$ almost surely in the full-batch case. Finally, we take the (unconditional) expectation with respect to all the token trajectories and mini-batches and rearrange the terms. Averaging over $t\in\{0,\dots,T-1\}$, and using our finite infimum assumption, we arrive at the result in Theorem~\ref{thm:mtcd_tpc_convergence}.
	
	\paragraph{On the convergence results.}
	When deriving convergence guarantees for VFL methods, it is quite challenging to deal with the presence of stale information in the updates. Even in client-server VFL, methods employing multiple local updates~\citep{Liu2022,Castiglia2022} converge at a $\mathcal{O}( \sfrac{\Delta}{\eta T})$ rate, for a stepsize $\eta\in\mathcal{O}(\sfrac{1}{P})$. (Recall that $P$ is the number of stale updates and $\Delta$ is the suboptimality of the initial iterate.)
	That is, although we can empirically observe a speedup when we increase the number of stale updates, this is not captured in the theoretical results. In fact, increasing $P$ shrinks the upper bound on the stepsize and thus leads to a slower convergence rate.
	
	Similarly, our convergence results achieve a $\mathcal{O}(\sfrac{\Delta}{\eta\pi T})$ convergence rate for $\eta\in\mathcal{O}(\sfrac{1}{P})$, where $\pi$ is the minimum probability of any client being visited by a token during each roaming step. For the client-server setting, under full participation---the setting considered in \citet{Liu2022} and \citet{Castiglia2022}---we have $\pi=1$. Thus, our rate recovers the rate of client-server VFL methods.
	
	\paragraph{On the mini-batch size.}
	Our method converges at a $ \mathcal{O}(1/T) $ rate for a sufficiently large batch size. In particular, exact convergence requires the batch size to be $\Omega (\sigma^2/\epsilon) $, where $ \sigma^2 $ is the upper bound on the gradient variance and $ \epsilon>0 $ is an error level. Or, even more precisely, we have a $\mathcal{O}(1/T) $ rate if
	\[
	B
	\geq
	\frac{\sigma^2}{\epsilon}
	\cdot
	\frac{\eta S^2Q^2L}{\pi}.
	\]
	
	{
		\paragraph{Reduced time and communication complexity.}
		Nevertheless, under some mild assumptions, we can show a reduction in the time and communication complexity of \texttt{MTCD} when compared to client-server VFL methods. To illustrate this, let us go over an example for which we can show that \textbf{1)} the rate of \texttt{MTCD} matches that of client-server VFL methods and that the \textbf{2)} time and \textbf{3)} communication cost per iteration of \texttt{MTCD} are lower than those of client-server VFL methods for a range of applications. Thus, for such applications, we provably achieve better time and communication complexity.
		
		In particular, consider the case where $K$ clients are divided into $C$ clusters, each with $K/C$ clients (assumed divisible) arranged in a path graph. We examine both \texttt{MTCD} and a client-server VFL method (such as \citet{Liu2022} and \citet{Castiglia2022}), each using the same number of stale updates $P$, and thus with a matching upper bound on the stepsize, $\eta\in\mathcal{O}(\sfrac{1}{P})$. \texttt{MTCD} employs $Q=1$ and $S=K/C$ across each path graph, in the token-per-cluster setting, visiting each client in each cluster exactly once per iteration. The client-server method employs $Q=K/C$.
		It follows from the discussion above that both approaches converge at the same rate.
		\begin{itemize}[leftmargin=1.0em]
			\item \textbf{Time complexity:} We model the available bandwidth of each client-server link as $\propto 1/l$, where $l$ is the number of links---a reasonable modeling assumption when a fixed bandwidth budget at the server is shared among the clients communicating with it. Consequently, we model the time needed to send a representation and a token across a client-server link as $\tau_t^h \cdot l$ and $\tau_t^Z \cdot l$, respectively. We further denote the time to send a token across a client-client link by $\tau_s^Z$ and the time to perform a local update by $\tau_q$.\footnote{
				Client bandwidth is typically less significant than server bandwidth, as the number of neighbors of a client usually does not grow with $K$.
			} Thus:
			\begin{itemize}
				\item For client-server VFL:
				\[
				\text{time per iteration}
				=
				\tau_t^h \cdot K + \tau_t^Z \cdot K + Q \cdot \tau_q
				.
				\]
				\item For \texttt{MTCD}:
				\[
				\text{time per iteration}
				=
				\tau_t^h \cdot K + \tau_t^Z \cdot \Gamma + S (\tau_s^Z+ Q \cdot \tau_q)
				.
				\]
			\end{itemize}
			Thus, for $Q$ and $S$ as mentioned above, we want to know whether
			\[
			{\textstyle
				\tau_t^h K + \tau_t^Z K + \frac{K}{C} \cdot \tau_q
				\geq
				\tau_t^h K + \tau_t^Z \Gamma + \frac{K}{C} (\tau_s^Z+ \tau_q).
			}
			\]
			Since $\Gamma=C$, this is equivalent to
			\[
			{
				\tau_t^Z / \tau_s^Z
				\geq
				\frac{K}{\Gamma(K-\Gamma)}
				.
			}
			\]
			Therefore, since typically $K\gg\Gamma$ and $ \tau_t^Z>\tau_s^Z $---given that client-server links often have a greater latency than client-client links~\citep{lin2021semi}---for a variety of applications, \texttt{MTCD} improves over the time complexity of client-server VFL.
			
			\item \textbf{Communication complexity:}
			As noted earlier, to succinctly capture the limitations of the client-server setup—namely, its vulnerability to server bandwidth bottlenecks and its single point of failure—we can quantify the relative impact of using client-server versus client-client links by comparing their communication costs, which vary by application.
			
			\hspace{2mm}
			We denote the cost of transmitting a token over a client-client link and a client-server link by $C_{CC}^Z$ and $C_{CS}^Z$, respectively. Similarly, we denote the cost of transmitting a representation over a client-client link and a client-server link by $C_{CC}^h$ and $C_{CS}^h$, respectively. For the reasons above, we often have $C_{CS}^h > C_{CC}^h$ and $C_{CS}^Z > C_{CC}^Z$.
			
			\hspace{2mm}
			In both \texttt{MTCD} and client-server VFL methods, all clients send $\bm{h}_{k,\mathcal{B}^{t}}(\bm{\theta}^{t}_k)$ to the server at the beginning of each iteration. However, while client-server VFL methods require sending the token to each client after synchronization, \texttt{MTCD} only requires the server to send $\Gamma$ tokens to the clients, which then roam a cluster for $S$ hops.
			Thus:
			
			\begin{itemize}
				\item For client-server VFL:
				\[
				\text{comm. cost per iteration}
				= KC_{CS}^h+KC_{CS}^Z
				.
				\]
				\item For \texttt{MTCD}:
				\[
				\text{comm. cost per iteration}
				= KC_{CS}^h+\Gamma C_{CS}^Z + S\Gamma C_{CC}^Z
				.
				\]
			\end{itemize}
			Thus, the question is whether the following inequality holds:
			\[
			KC_{CS}^h+KC_{CS}^Z
			\geq
			KC_{CS}^h+\Gamma C_{CS}^Z + S\Gamma C_{CC}^Z.
			\]
			Or, equivalently, whether
			\[
			C_{CS}^Z / C_{CC}^Z
			\geq
			\frac{S\Gamma}{(K-\Gamma)}.
			\]
			Since typically $K\gg\Gamma$ and $C_{CS}^Z> C_{CC}^Z$, we thus have that, for a range of applications, \texttt{MTCD} improves over the communication complexity of client-server VFL.
	\end{itemize}}
	
	
	\subsection{An extension with overlapping token trajectories}
	We now extend our convergence guarantees to a setting where we allow for overlapping token trajectories. We propose choosing the convex combination weights to be $\bm{w}_{k}=(\frac{1}{\Gamma},\dots,\frac{1}{\Gamma})\in\mathbb{R}^\Gamma$ for all $k$, thus combining the model estimates by averaging them. To handle this periodic combination of the model estimates in this setting, we now further assume convexity. (We elaborate on this below.)
	
	Note that the cluster-dependent notation---namely $i_\gamma^{t,p}$ and $\bm{p}_\gamma^t$ in the definition of $\bm{p}^t$---as well as the lemmas from the token-per-cluster setting, apply here as well. They differ only in that the ``cluster'' now represents the entire communication graph. The standard definition of convexity is given below.
	
	\begin{assumption}[Convexity] \label{ass:convexity}
		Function~$f\colon\mathbb{R}^d\mapsto\mathbb{R}$ is convex. That is,
		for all $\bm{u},\bm{v}\in\mathbb{R}^d$ and $a\in[0,1]$:
		\begin{equation} \label{eq:convexity}
			f(a\bm{u}+(1-a)\bm{v})
			\leq
			af(\bm{u})
			+(1-a)f(\bm{v})
			.
			\tag{A\ref*{ass:convexity}}
		\end{equation}
	\end{assumption}
	
	\begin{theorem}[Overlapping token trajectories setting] \label{thm:mtcd_convergence}
		Let $\{\bm{\theta}^t\}$ be a sequence generated by Algorithm~\ref{alg:MTCD} in the setting allowing for overlapping token trajectories. If $f$ is a convex~\eqref{eq:convexity}, $L$-smooth function with a finite infimum~\eqref{eq:L-smoothness}, our mini-batch gradient estimate is unbiased~\eqref{eq:unbiased} and has a bounded variance~\eqref{eq:bounded_var}, and each node has a minimum probability $\pi>0$ of being visited by a token at each iteration~\eqref{eq:min_prob_token_visit}, we have that, for a stepsize $\eta \in \big(0,\frac{1}{2LSQ}\big]$:
		\begin{equation} \label{eq:overlapping-tokens-theorem}
			\frac{1}{T}
			\sum_{t=0}^{T-1}
			\mathbb{E} \lVert \nabla f(\bm{\theta}^{t}) \rVert_{}^{2}
			\leq
			\frac{2\Delta}{\eta\pi T}
			+\frac{\eta S^2Q^2\sigma^2L}{\pi B },
		\end{equation}
		where $\Delta\coloneqq f(\bm{\theta}^{0}) - f^\star$ and the expectation is with respect to token trajectories $\{\bm{p}^t\}$ and mini-batches $\{\mathcal{B}^t\}$. For the special, full-batch case, we also have that $\nabla f \left( \bm{\theta}^{t}\right)\to0$ almost surely.
	\end{theorem}
	The proof of Theorem~\ref{thm:mtcd_convergence} follows a similar approach to that of Theorem~\ref{thm:mtcd_tpc_convergence}. However, allowing for overlapping trajectories prevents the quadratic upper bound from decoupling across tokens. We tackle this by resorting to the convexity assumption, which allows us to obtain a decoupled upper bound.\footnote{We discuss the derivation of results in suboptimality in Appendix~\ref{sec:thm2_proof}.}

	
	\begin{figure*}[!t]
		\centering
		\subfloat[Complete graph ($\alpha_{\mathcal{G}}=40$)]{\includegraphics[width=0.24\textwidth]{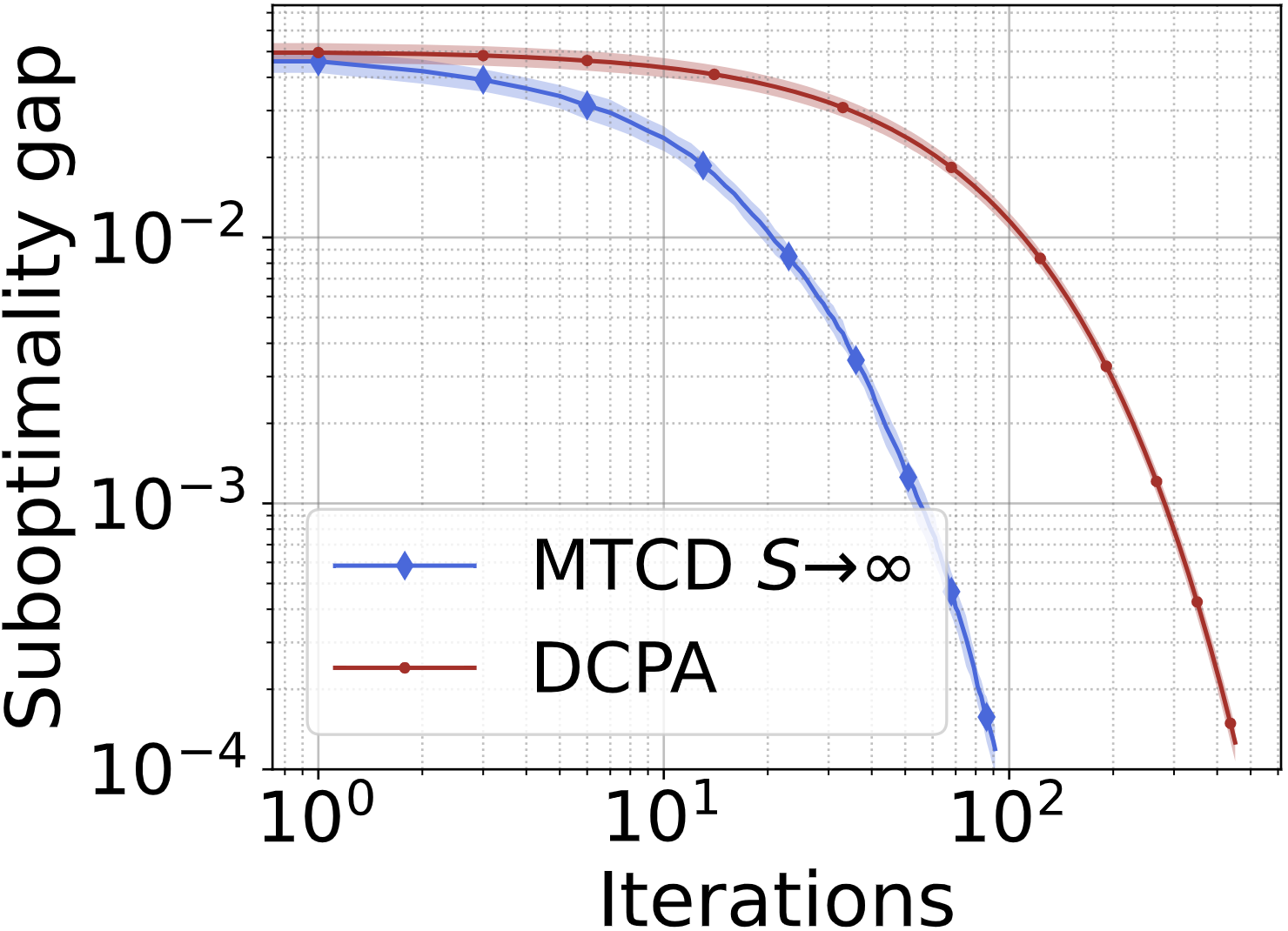}%
			\hspace{1mm}
			\includegraphics[width=0.24\textwidth]{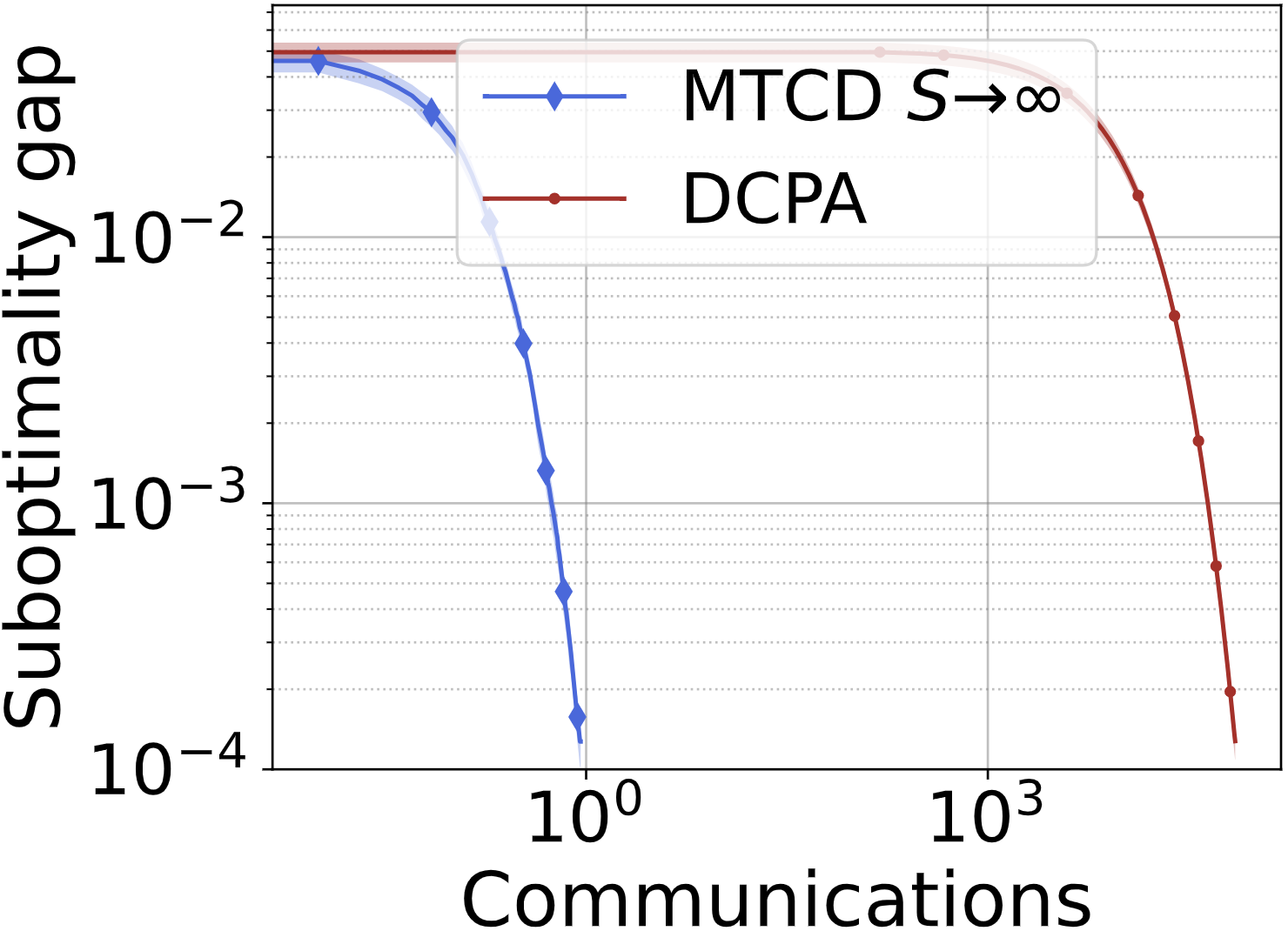}%
			\label{fig:complete}}
		\hfil
		\subfloat[Erdős–Rényi graph with $p=0.4$ ($\alpha_{\mathcal{G}}=7.7$)]{\includegraphics[width=0.24\textwidth]{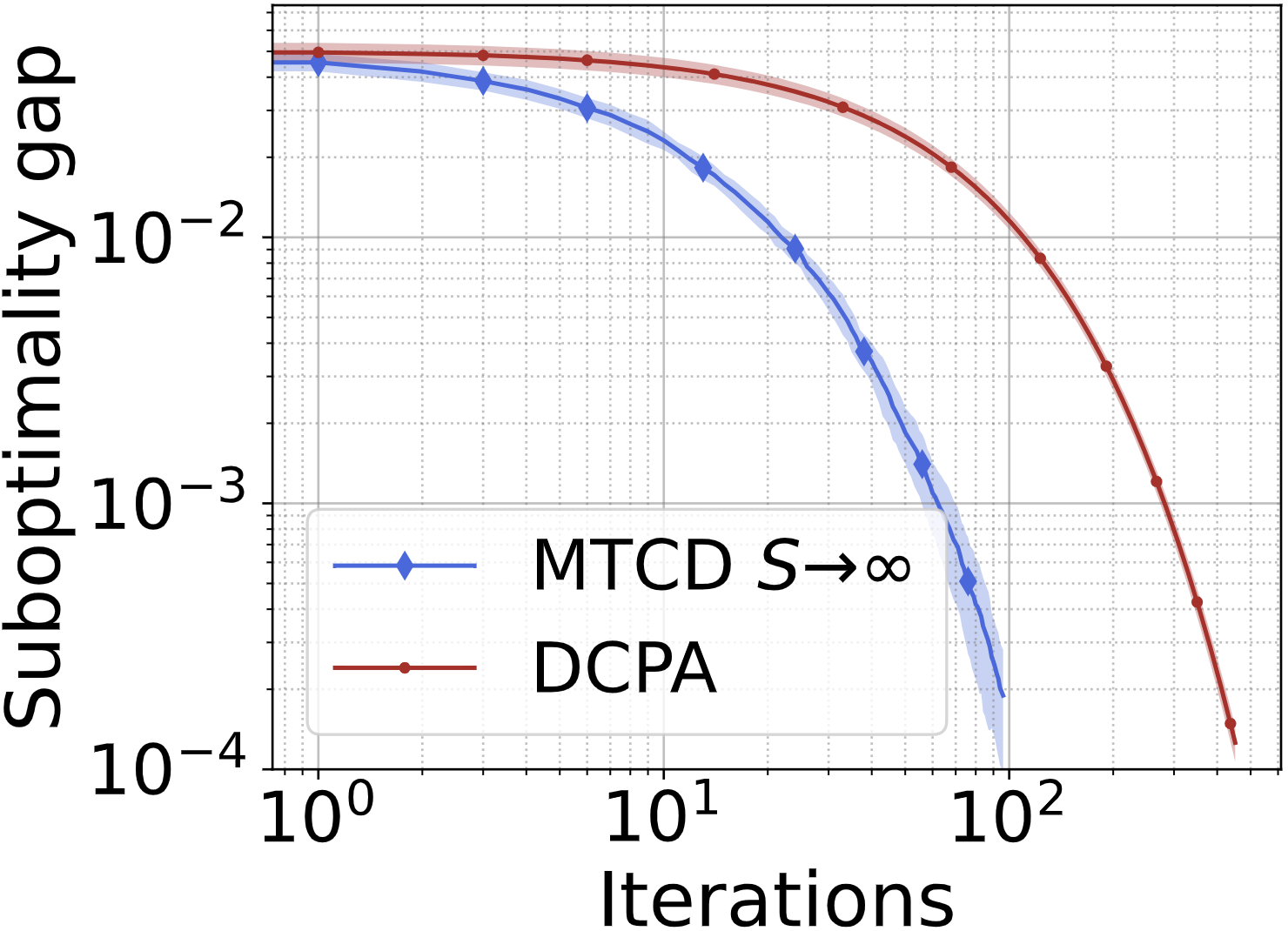}%
			\hspace{1mm}
			\includegraphics[width=0.24\textwidth]{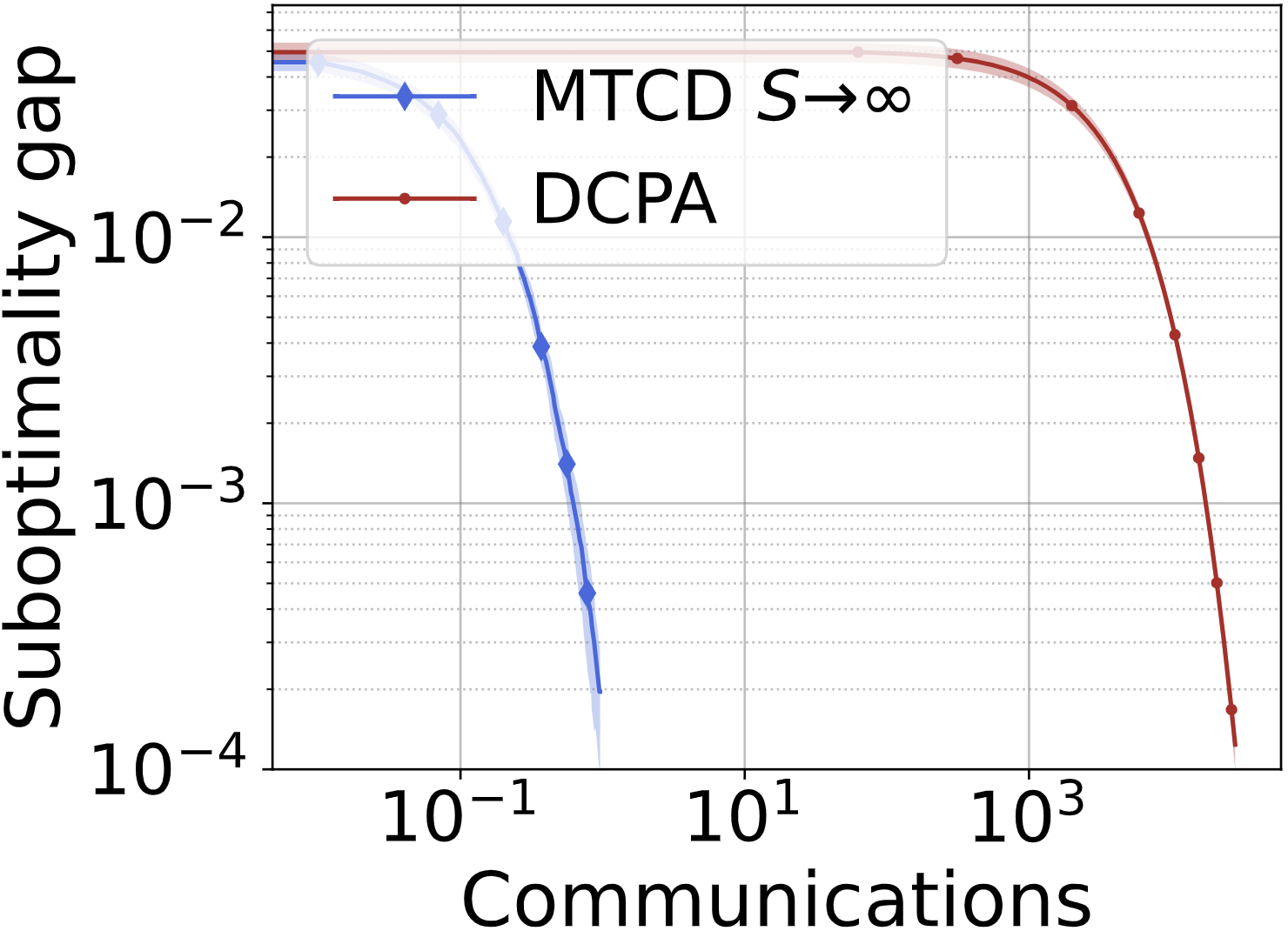}%
			\label{fig:random0.4}}
		
		\subfloat[Erdős–Rényi graph with $p=0.2$ ($\alpha_{\mathcal{G}}=1.4$)]{\includegraphics[width=0.24\textwidth]{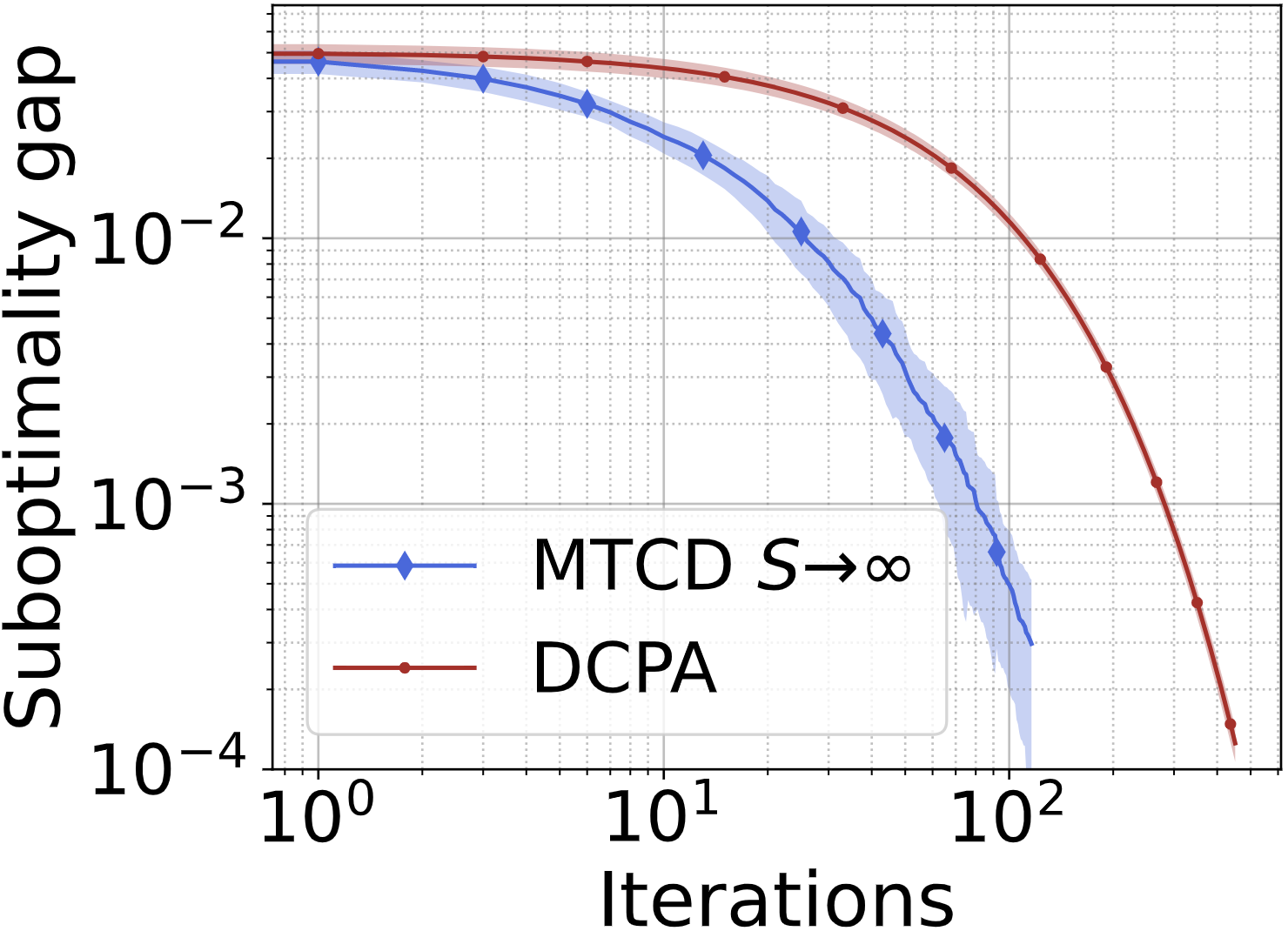}%
			\hspace{1mm}
			\includegraphics[width=0.24\textwidth]{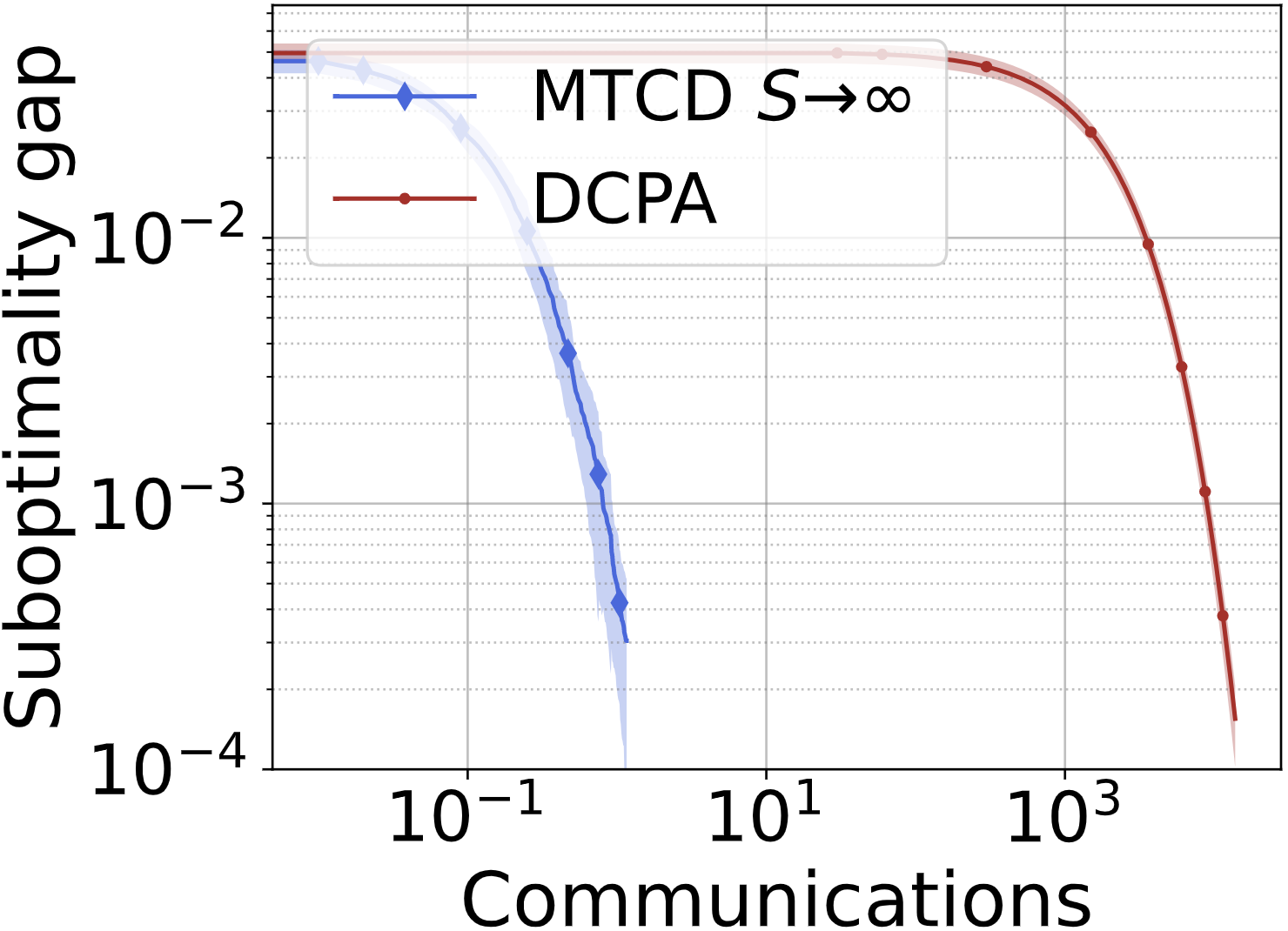}%
			\label{fig:random0.2}}
		\hfil
		\subfloat[Grid graph ($5\times8$) ($\alpha_{\mathcal{G}}=0.15$)]{\includegraphics[width=0.24\textwidth]{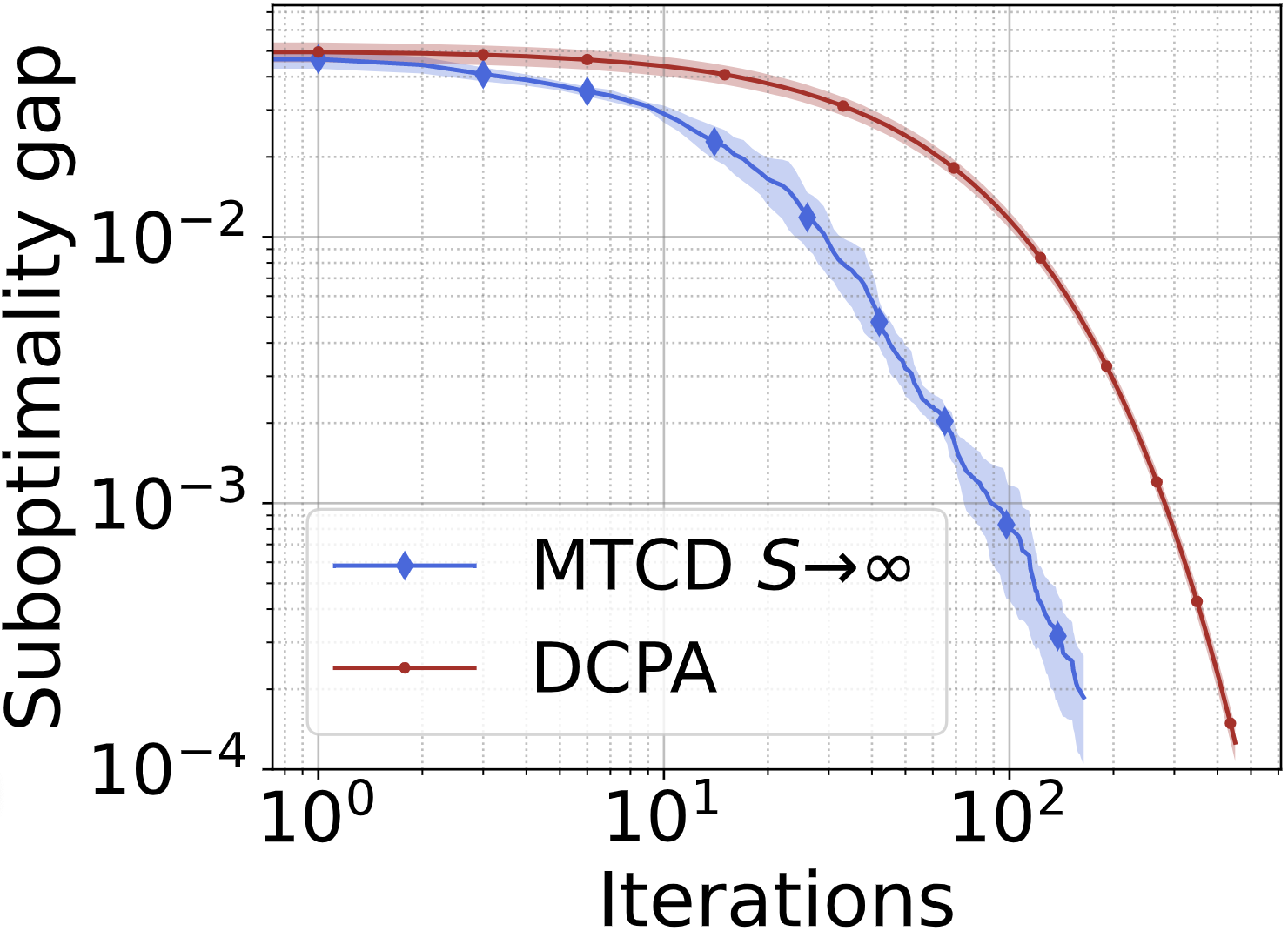}%
			\hspace{1mm}
			\includegraphics[width=0.24\textwidth]{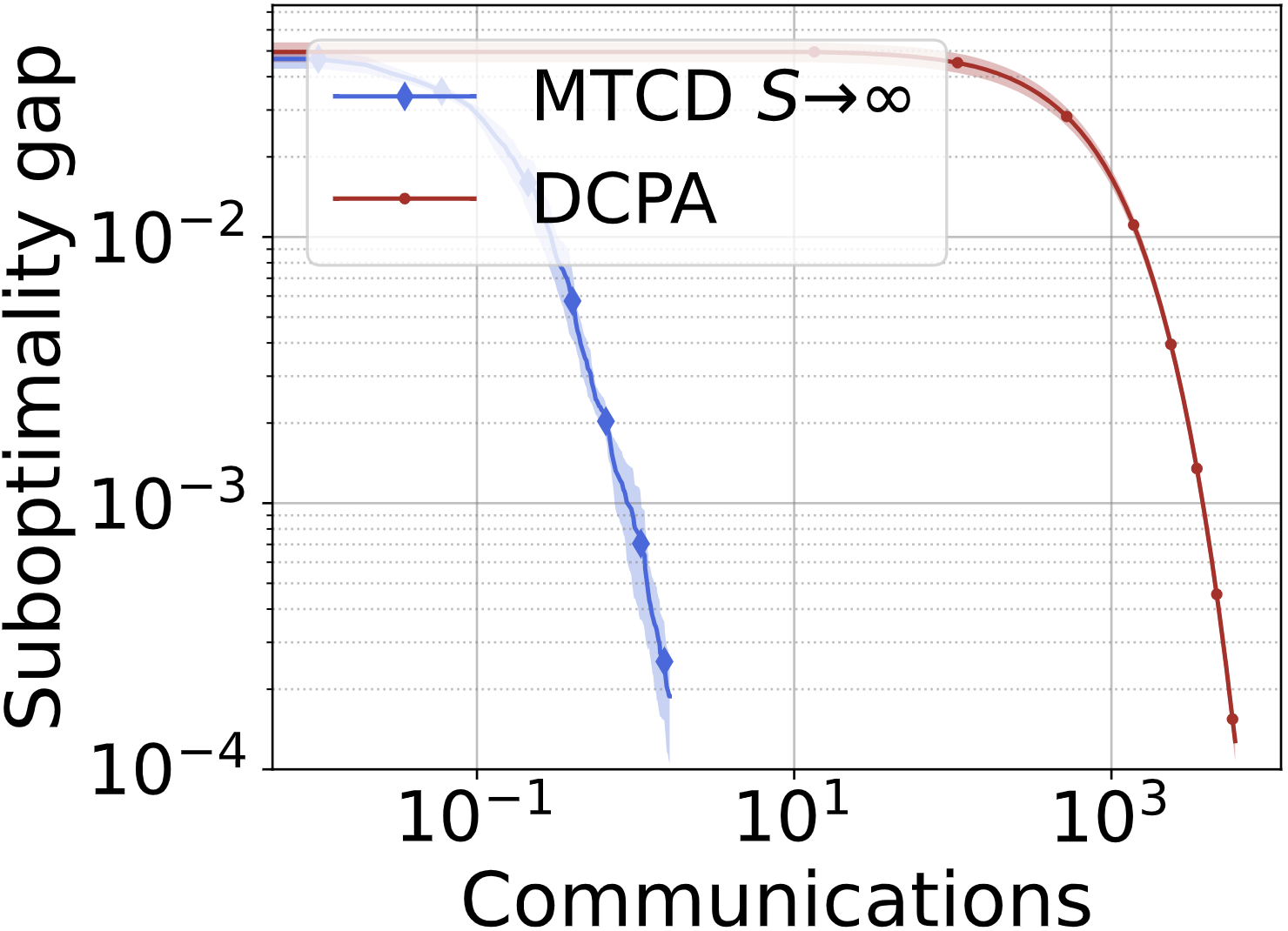}%
			\label{fig:grid}}
		
		\subfloat[Cycle graph ($\alpha_{\mathcal{G}}=0.025$)]{\includegraphics[width=0.24\textwidth]{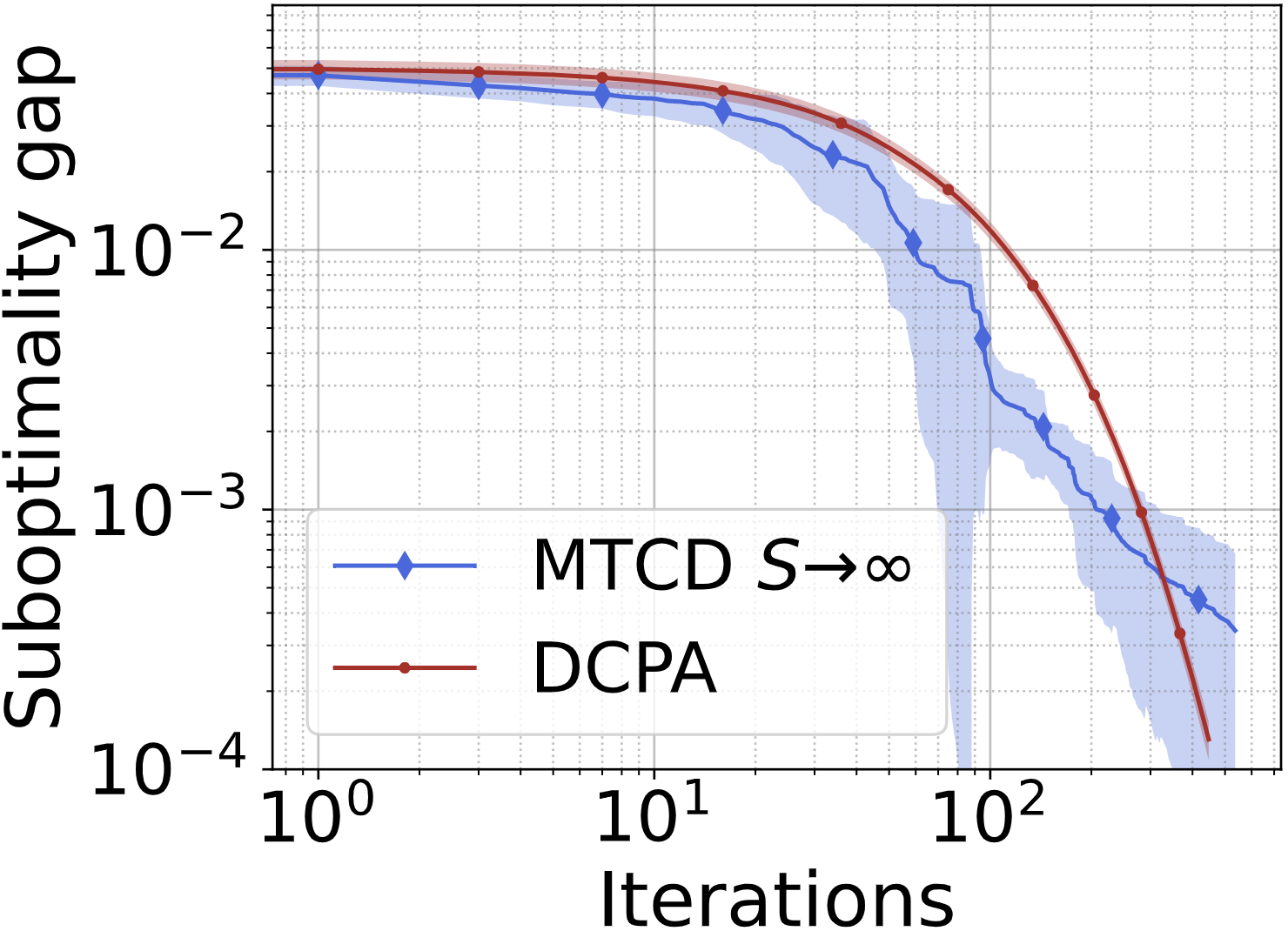}%
			\hspace{1mm}
			\includegraphics[width=0.24\textwidth]{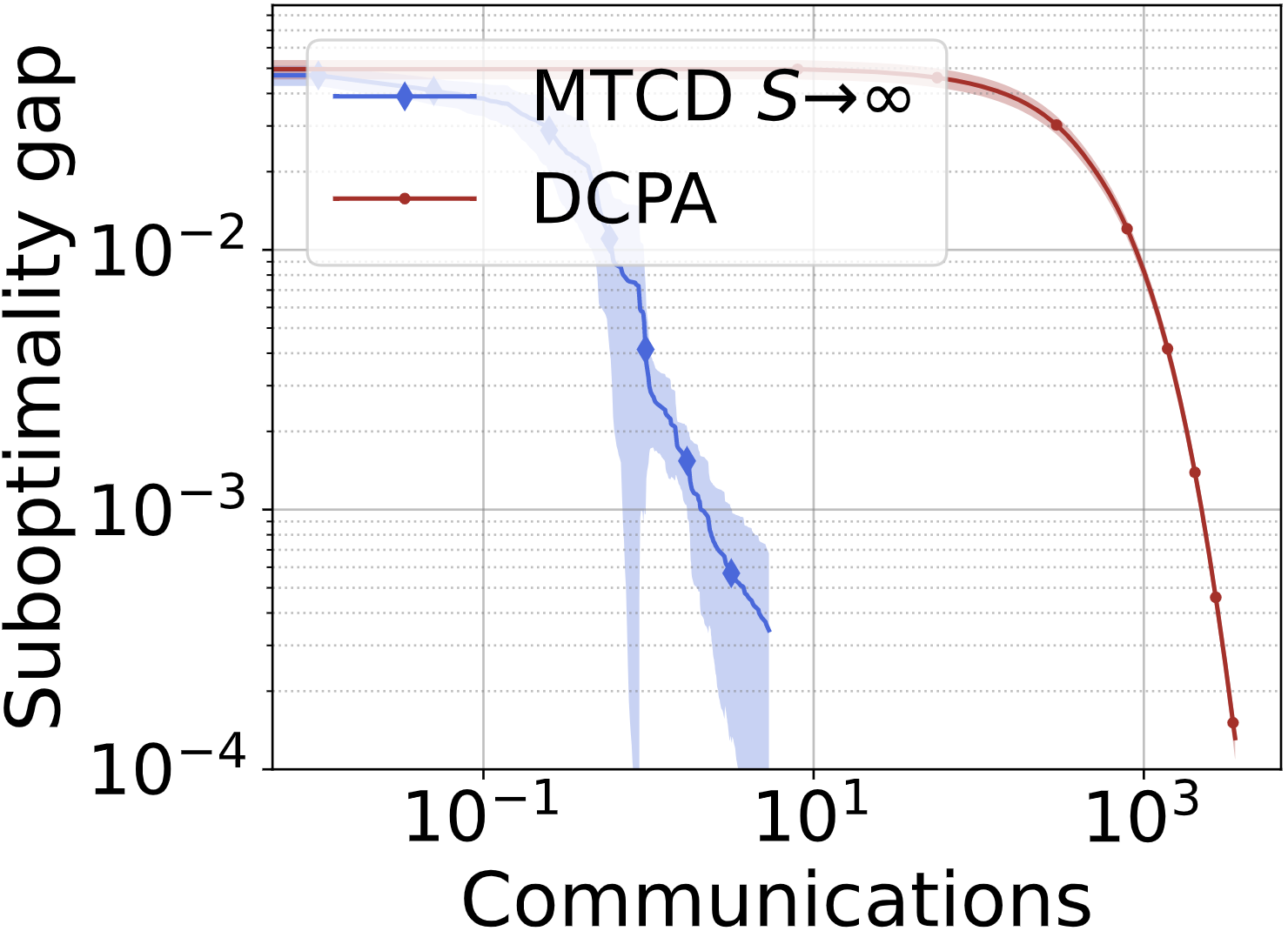}%
			\label{fig:cycle}}
		\hfil
		\subfloat[Path graph ($\alpha_{\mathcal{G}}=0.0061$)]{\includegraphics[width=0.24\textwidth]{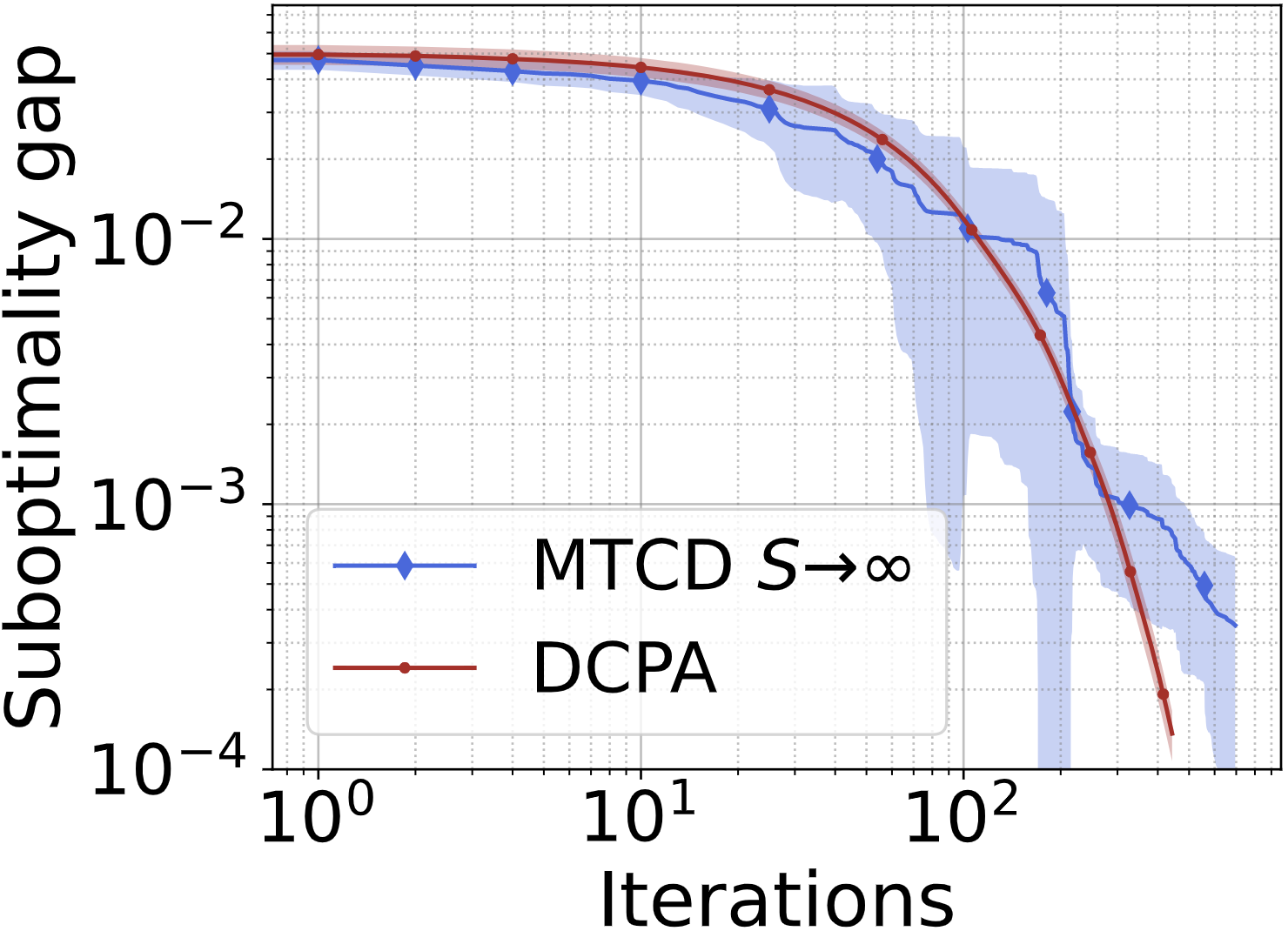}%
			\hspace{1mm}
			\includegraphics[width=0.24\textwidth]{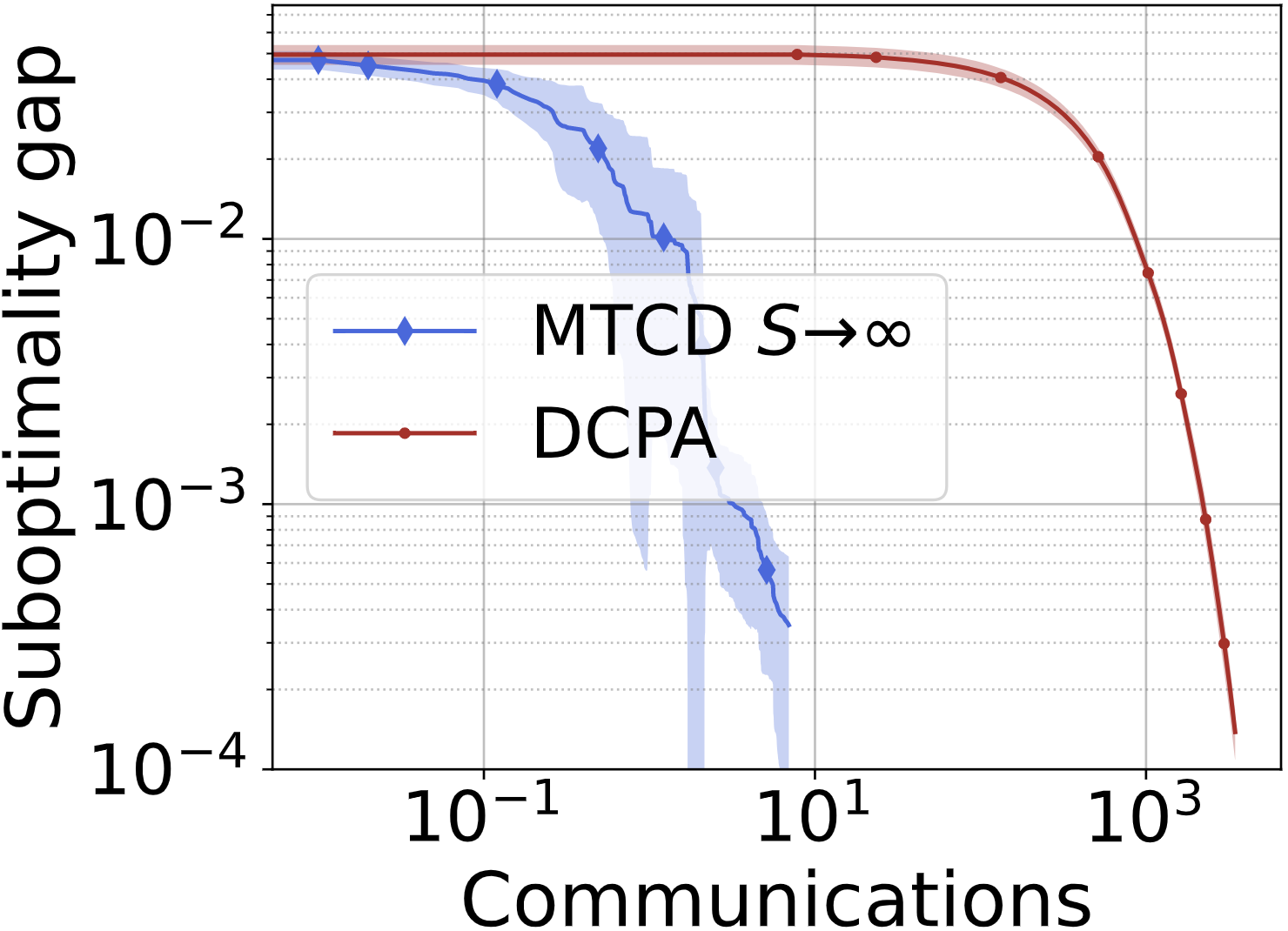}%
			\label{fig:path}}
		
		\caption{Ridge regression with $K=40$ clients, for six different network topologies. The \texttt{MTCD} run has $S\to\infty$ and $\Gamma=1$ (i.e. it corresponds to \texttt{STCD}). Algebraic connectivity, $\alpha_{\mathcal{G}}$, is the second smallest eigenvalue of the Laplacian matrix of a graph $\mathcal{G}$.}
		\label{fig:stcd_dcpa_rr}
	\end{figure*}

	\begin{figure*}[!t]
		\centering
		%
		\subfloat[Erdős–Rényi graph with $p=0.4$ ($\alpha_{\mathcal{G}}=7.7$)]{\includegraphics[width=0.24\textwidth]{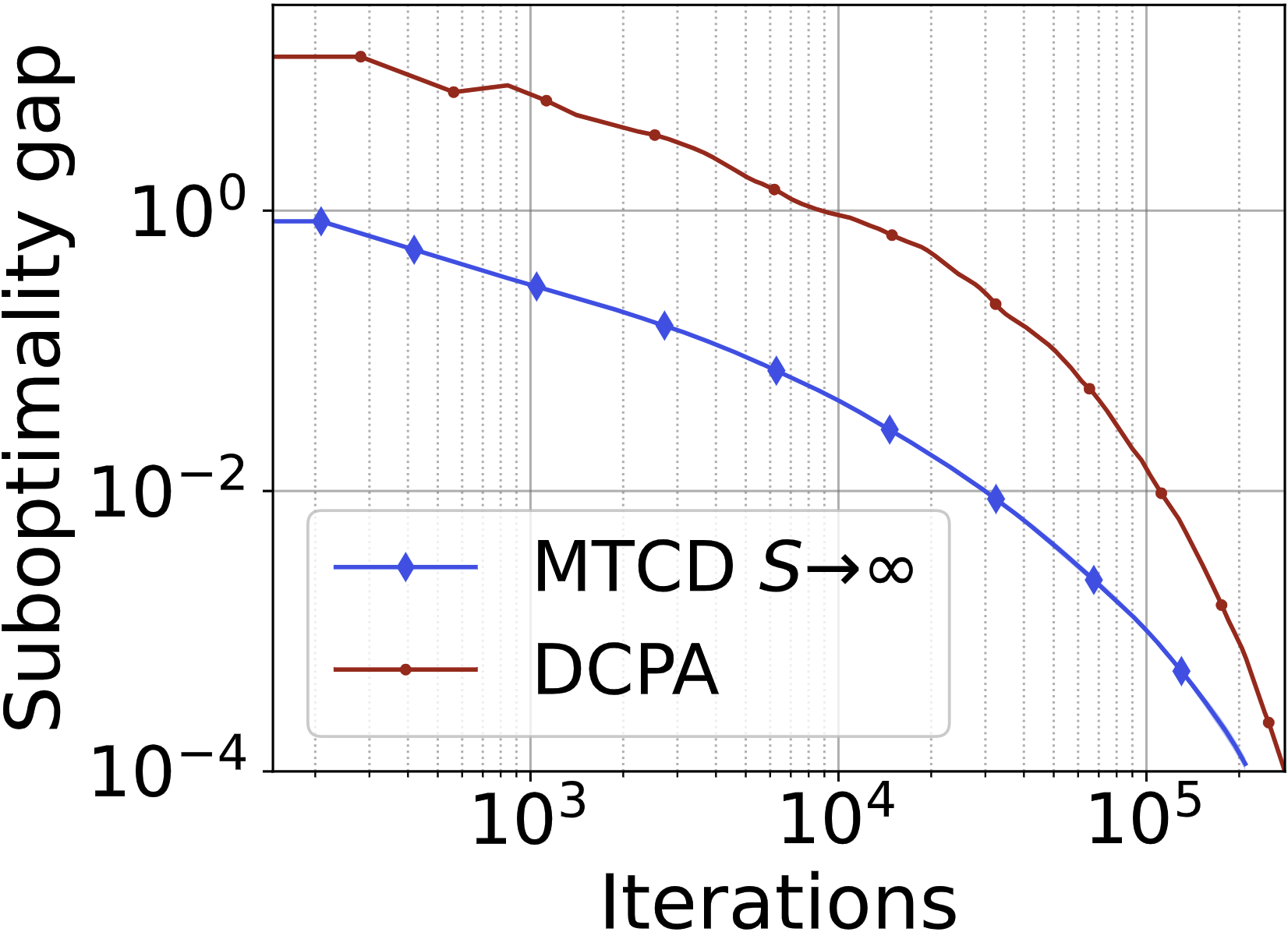}%
			\hspace{1mm}
			\includegraphics[width=0.24\textwidth]{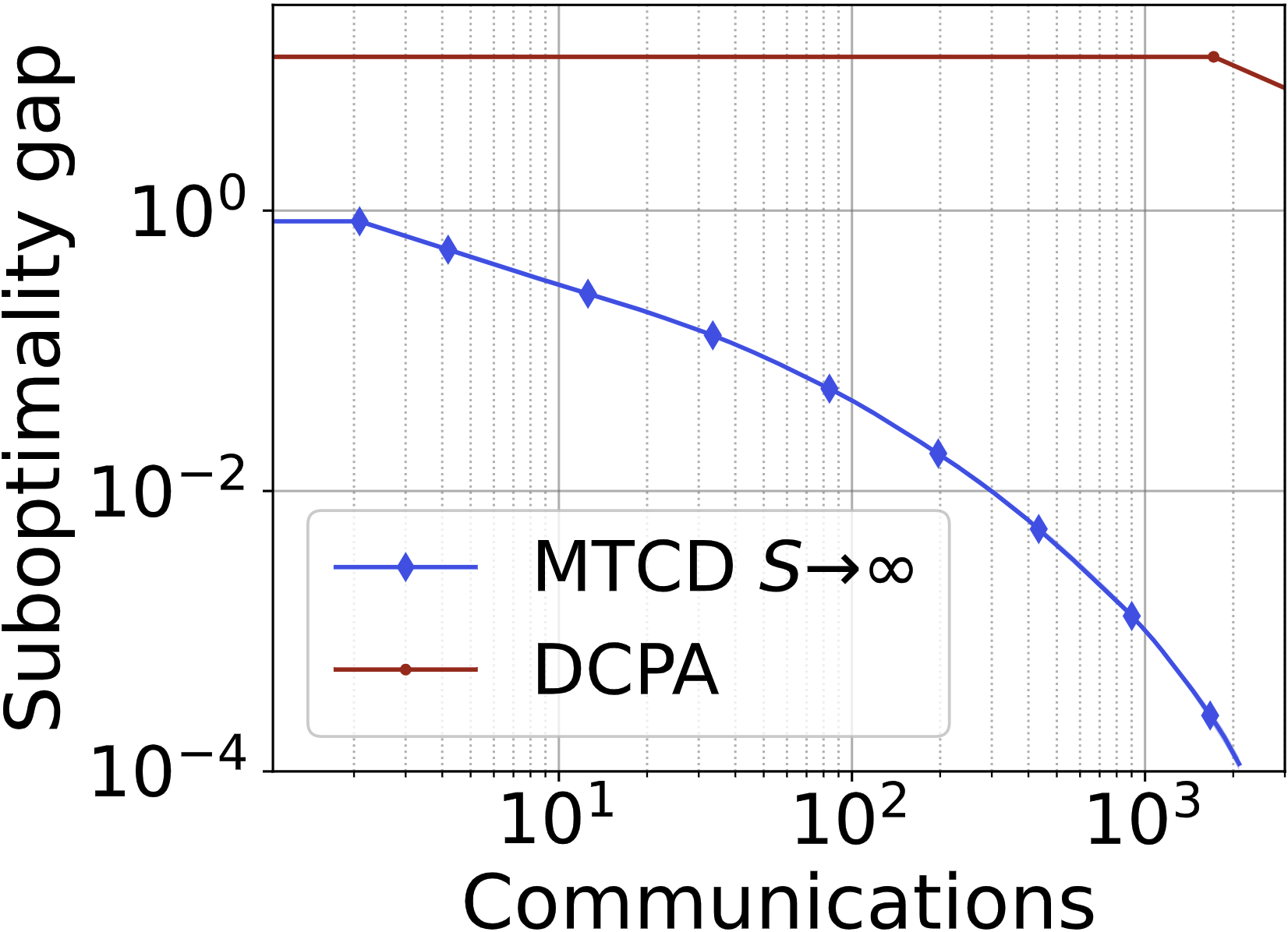}\label{fig:old_slr1}}%
		\hfil
		\subfloat[Path graph ($\alpha_{\mathcal{G}}=0.0061$)]{\includegraphics[width=0.24\textwidth]{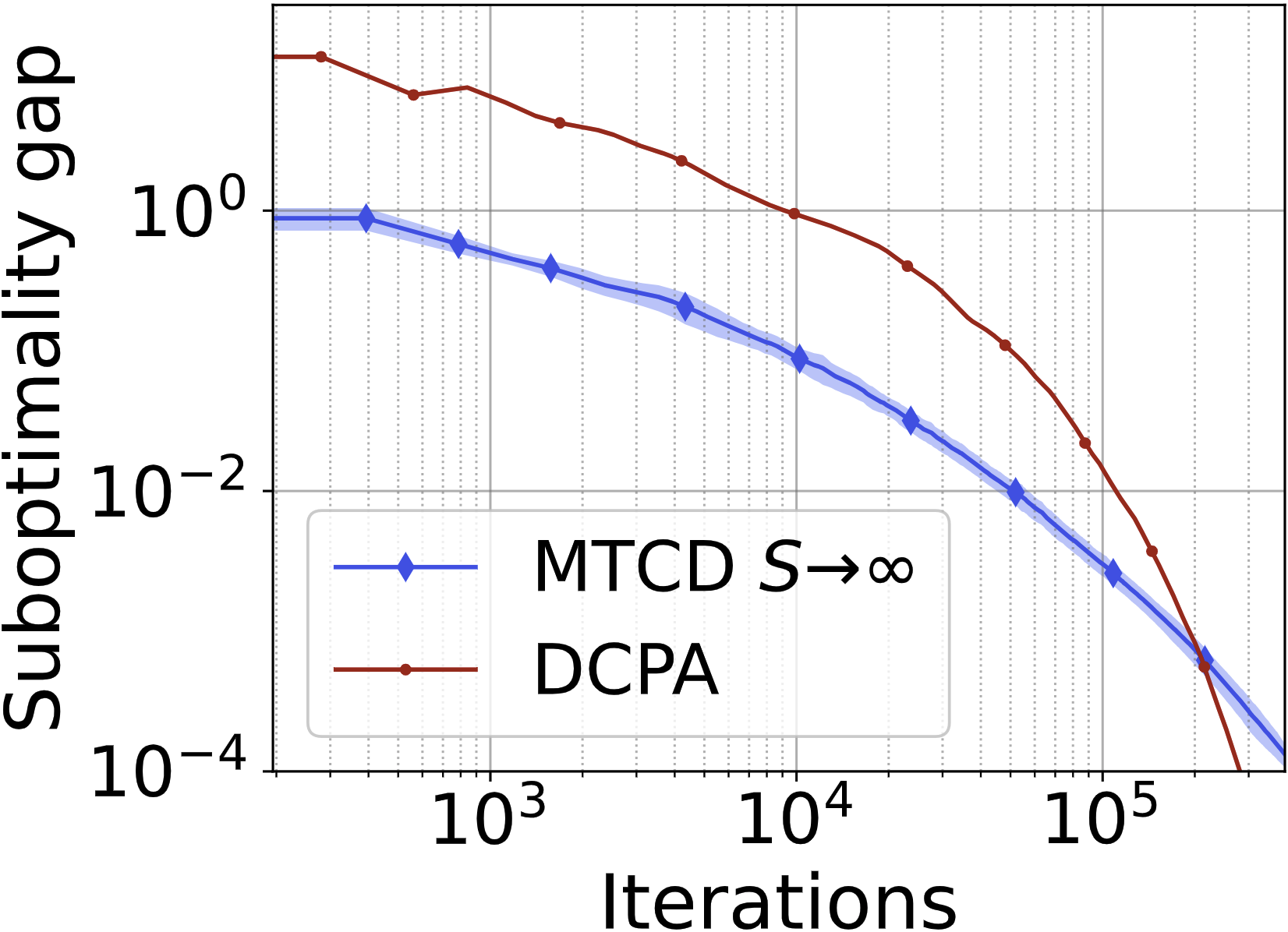}%
			\hspace{1mm}
			\includegraphics[width=0.24\textwidth]{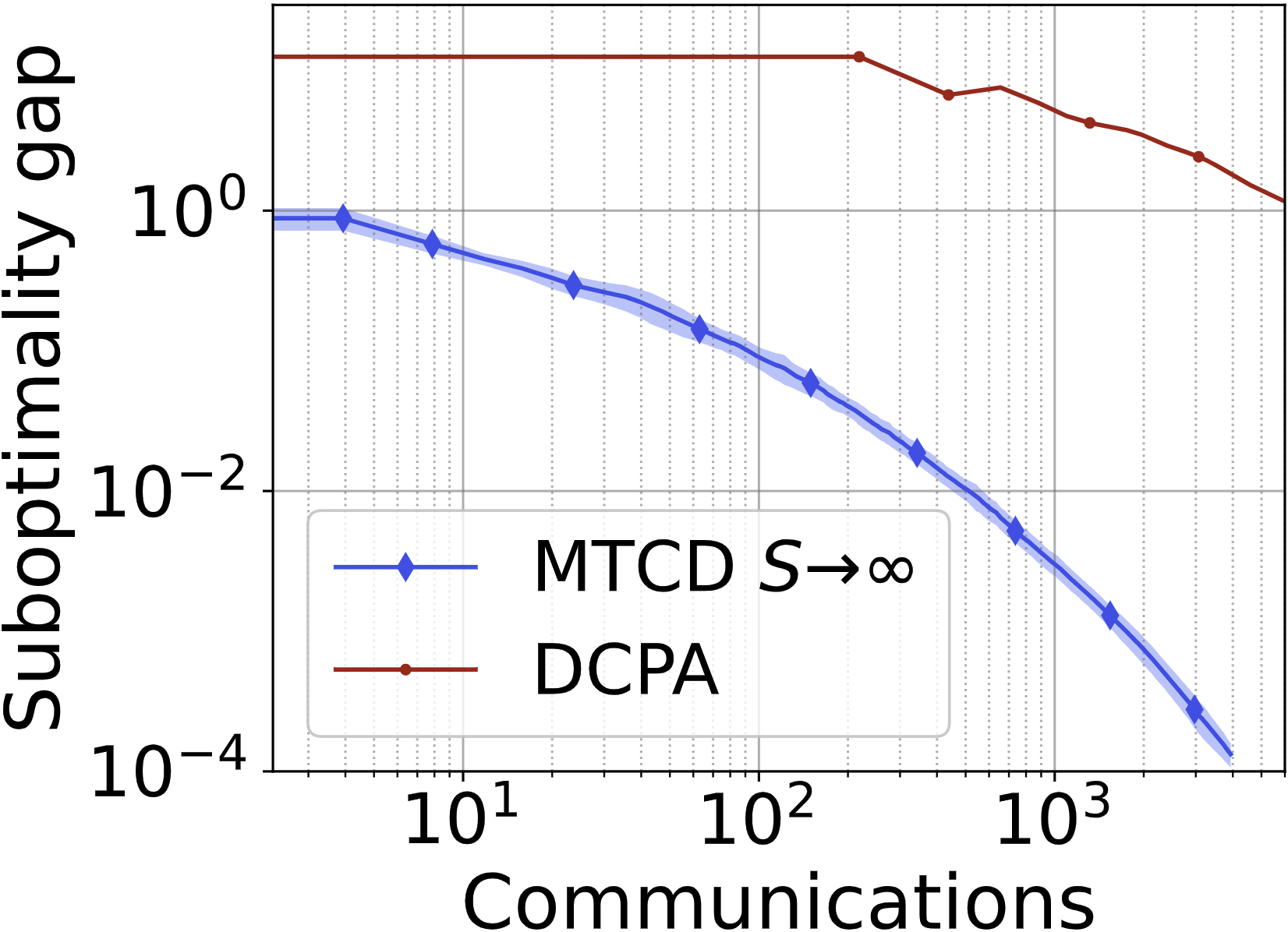}\label{fig:old_slr2}}%
		%
		\caption{
			Sparse logistic regression on a Erdős–Rényi graph and a path graph, both with $K=40$ clients.
			The $S\to\infty$ \texttt{MTCD} run has $\Gamma=1$ (i.e., it corresponds to \texttt{STCD}).
			Note that the flat \texttt{DCPA} trajectories with respect to communication cost have not plateaued, they simply take longer to see a drop in suboptimality.
		}
		\label{fig:stcd_dcpa}
	\end{figure*}
	
	\section{Experiments}
	\label{sec:experiments}
	We empirically evaluate our method, comparing it with \texttt{DCPA}~\citep{Alghunaim2021}, a fully decentralized approach, and with \texttt{SVFL}, a standard VFL method in the client-server setting as described in \citet{Liu2022}. Although low trajectory variances conceal some confidence intervals, all experiments are run over 5 seeds.
	
	\subsection{Convex problems}
	\label{sec:cvx_prob}
	
	In this section, to allow for a fair comparison between decentralized and SD approaches, we define, for all methods, an iteration as the cumulative number of hops. We use CVXPY~\citep{diamond2016} to obtain $f^\star$ and use the (relative) suboptimality gap $(f(\bm{\theta}^t)-f^\star)/f^\star$ as a metric.
	
	We denote the ratio of the cost of client-server to client-client communications as $R_C\coloneqq C_{CS}^Z/C_{CC}^Z=C_{CS}^h/C_{CC}^h$.\footnote{We assume, for simplicity, that client-server communications cost the same in both directions (uplink and downlink), although this does not always hold.} When plotting the suboptimality gap with respect to the communication cost, for $R_C=100$: we take $C_{CS}^Z$ as a unit cost, scale $C_{CC}^Z$ by $0.01$, and use these values to compute the total (weighted) communication cost. As explained in Section~\ref{sec:problem_setup}, for generalized linear models---such as the ones considered in this section---$ C_{CC}^h = C_{CC}^Z $ and $ C_{CS}^h = C_{CS}^Z $.
	
	We perform ridge regression on a dataset generated as in~\citet{Alghunaim2021},
	with $N=1000$ samples and $d=2000$ features split evenly across clients. Here, $f(\bm{\theta})=
	\left\Vert
	\bm{X}\bm{\theta}
	-\bm{y}
	\right\rVert_2^2/2
	+\alpha
	\Vert
	\bm{\theta}
	\rVert_2^2/2$, where $\bm{y}$ are the labels and $\alpha = 10$.
	For \texttt{MTCD}, $\eta=10^{-5}$ and $Q=20$.
	For \texttt{SVFL}, $\eta=5\times10^{-7}$ and $Q=20$.
	For \texttt{DCPA}, $\mu_w= 10^{-2}$, $\mu_y=3\times 10^{-4}$, and $\mu_x=3\times 10^{-2}$.
	
	We perform sparse logistic regression on Gisette~\citep{Guyon2004}, where $N=6000$ and $d=5000$. Again, the features are split evenly across clients. Let $s(z)\coloneqq(1+e^{-z})^{-1}$, $f(\bm{\theta})$ is given by:
	\[
	\sum_{n=1}^N
	\left[
	(1-y_n)
	\log(1-s(\bm{x}_n^\top\bm{\theta}))
	-
	y_n
	\log s(\bm{x}_n^\top\bm{\theta})
	\right]
	+\beta
	\lVert
	\bm{\theta}
	\rVert_1,
	\]
	where $\bm{x}_n$ and $y_n\in\{0,1\}$ denote sample~$n$ and its label, respectively, and $\beta = 1$. For \texttt{MTCD}, $\eta=10^{-4}$ and $Q=30$. For \texttt{DCPA}, $\mu_w= 10^{-3}, \mu_y=3\times 10^{-5}, \mu_x=3\times 10^{-3}$.\footnote{In \texttt{DCPA}, the lack of a closed-form solution for the proximal operator of the convex conjugate of the logistic regression loss forces each client to solve a local optimization problem at every iteration.}
	
	\paragraph{Fully decentralized setting.}
	In Figure~\ref{fig:stcd_dcpa_rr} and Figure~\ref{fig:stcd_dcpa}, we see that, while \texttt{MTCD} with $S\to\infty$ and $\Gamma=1$ (that is, \texttt{STCD}) does not consistently outperform \texttt{DCPA} in convergence per iteration, it is significantly more communication efficient. However, we also observe that \texttt{STCD} is sensitive to poorly connected networks, as seen when comparing the results in the better-connected graphs to those in the poorly-connected graph. Note that, for sparse logistic regression, while the proximal term used to handle the regularizer is not covered by our analysis, we observe that it performs well empirically. 
	
	\begin{figure*}[!t]
		\centering
		\includegraphics[width=0.25\textwidth]{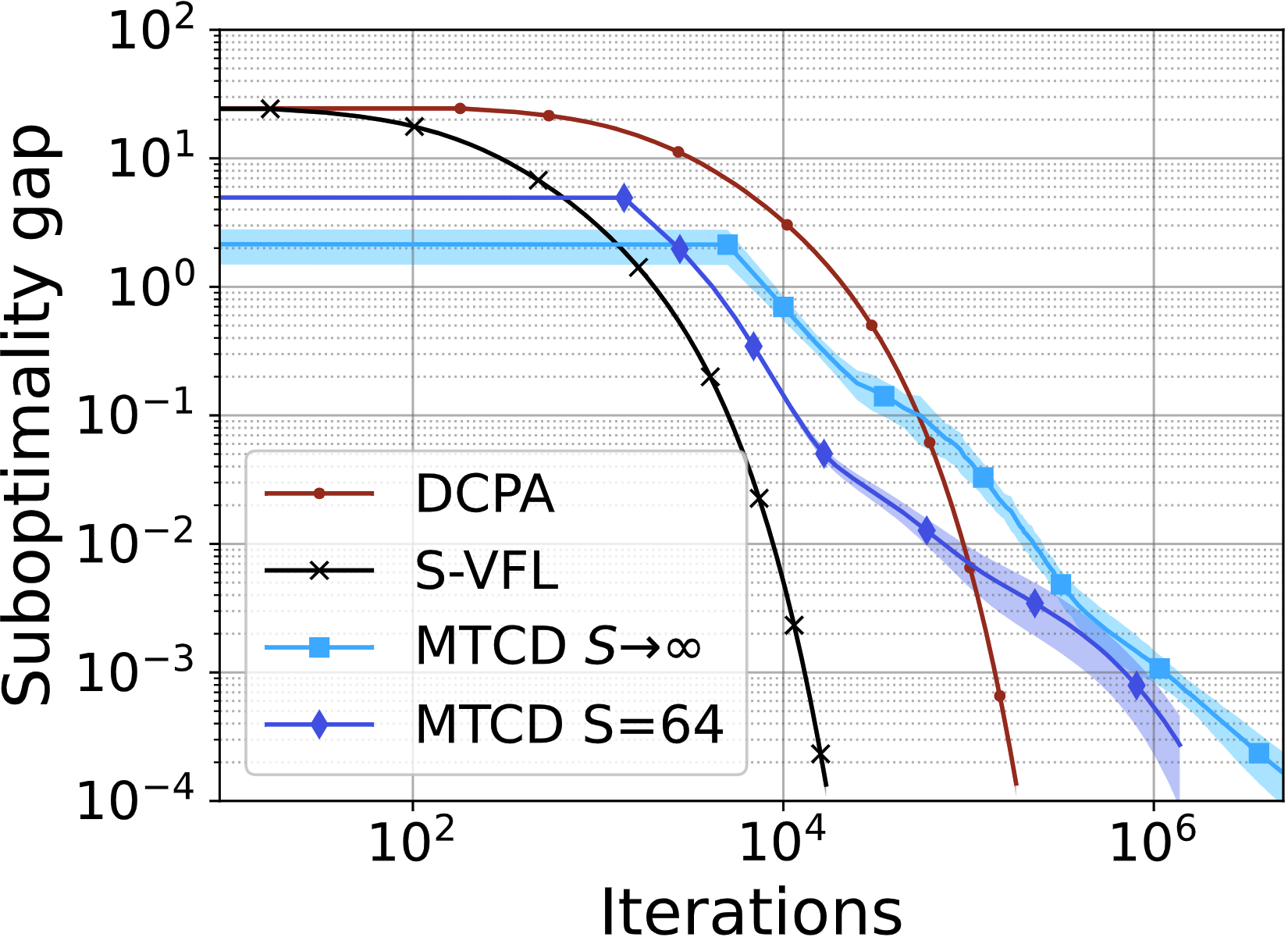}
		\hfil
		\includegraphics[width=0.25\textwidth]{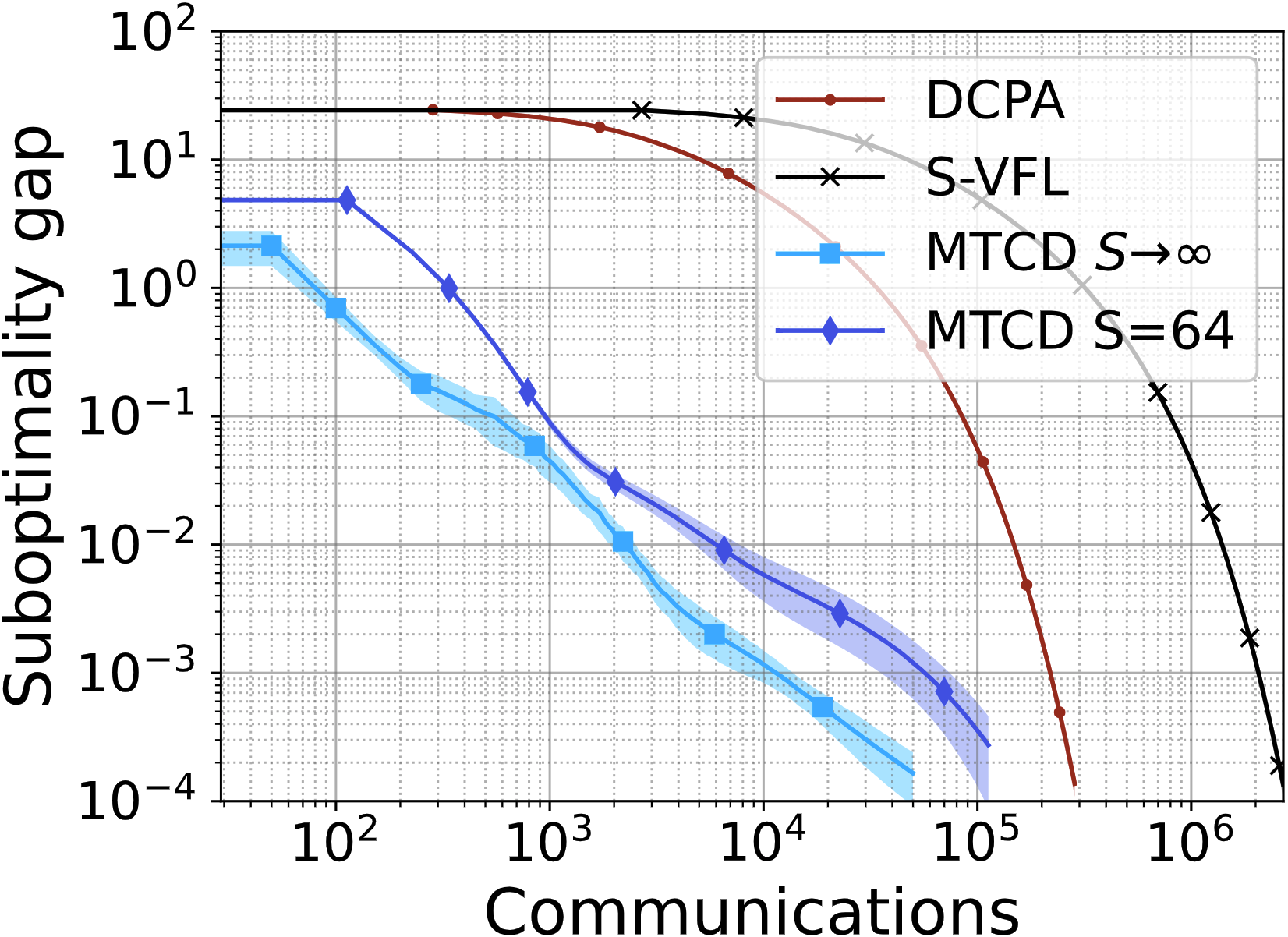}
		\hfil
		\raisebox{-1mm}{\includegraphics[width=0.25\textwidth]{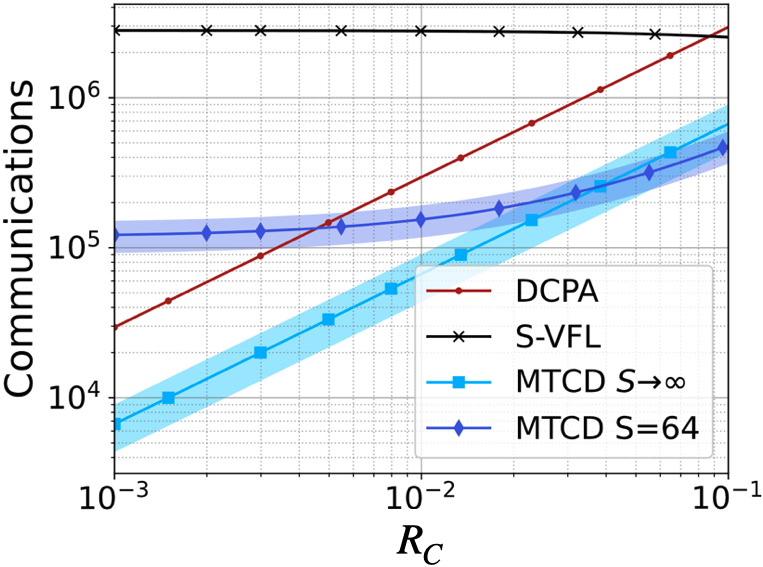}}
		\\
		\includegraphics[width=0.25\textwidth]{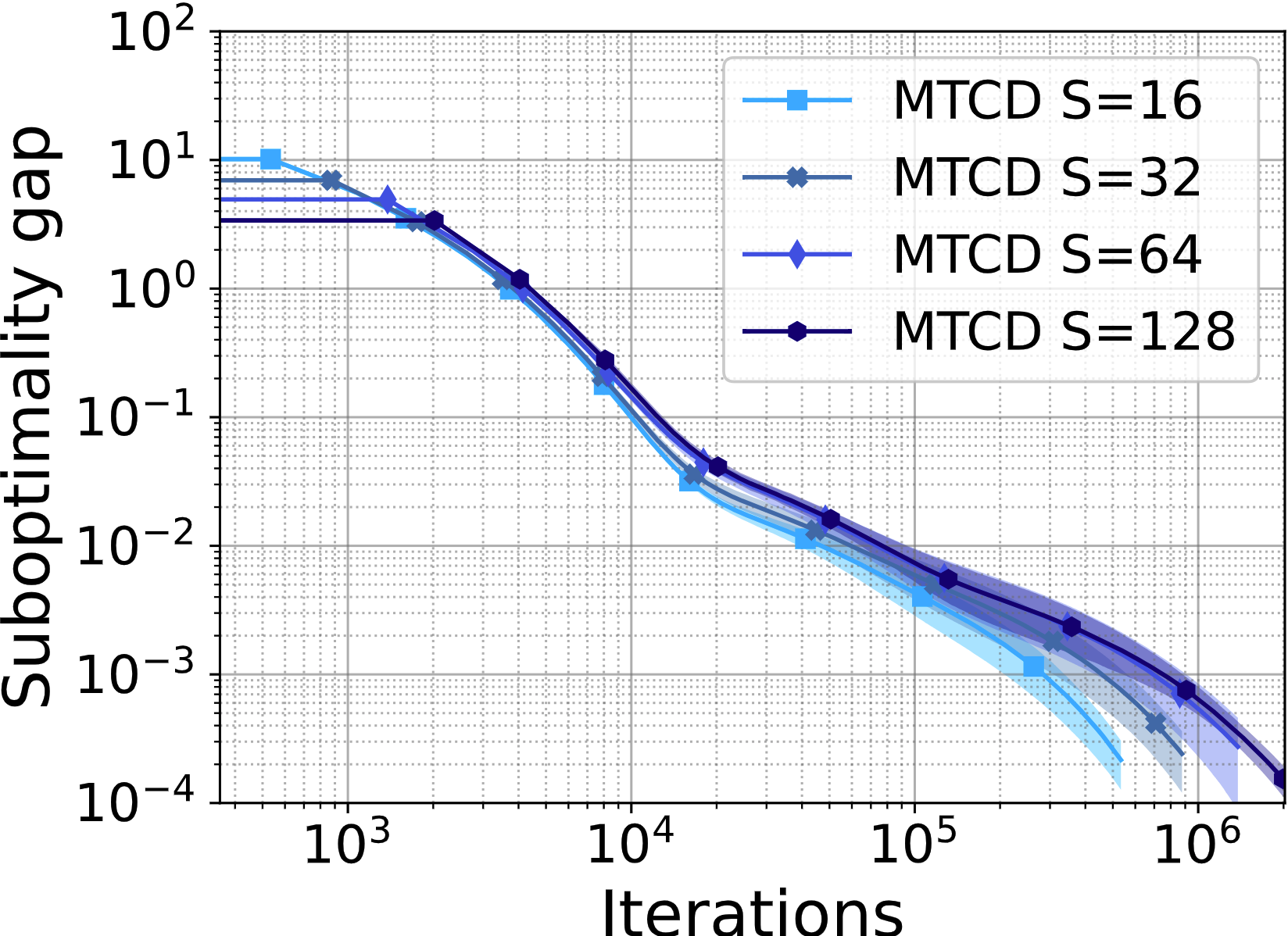}
		\hfil
		\includegraphics[width=0.25\textwidth]{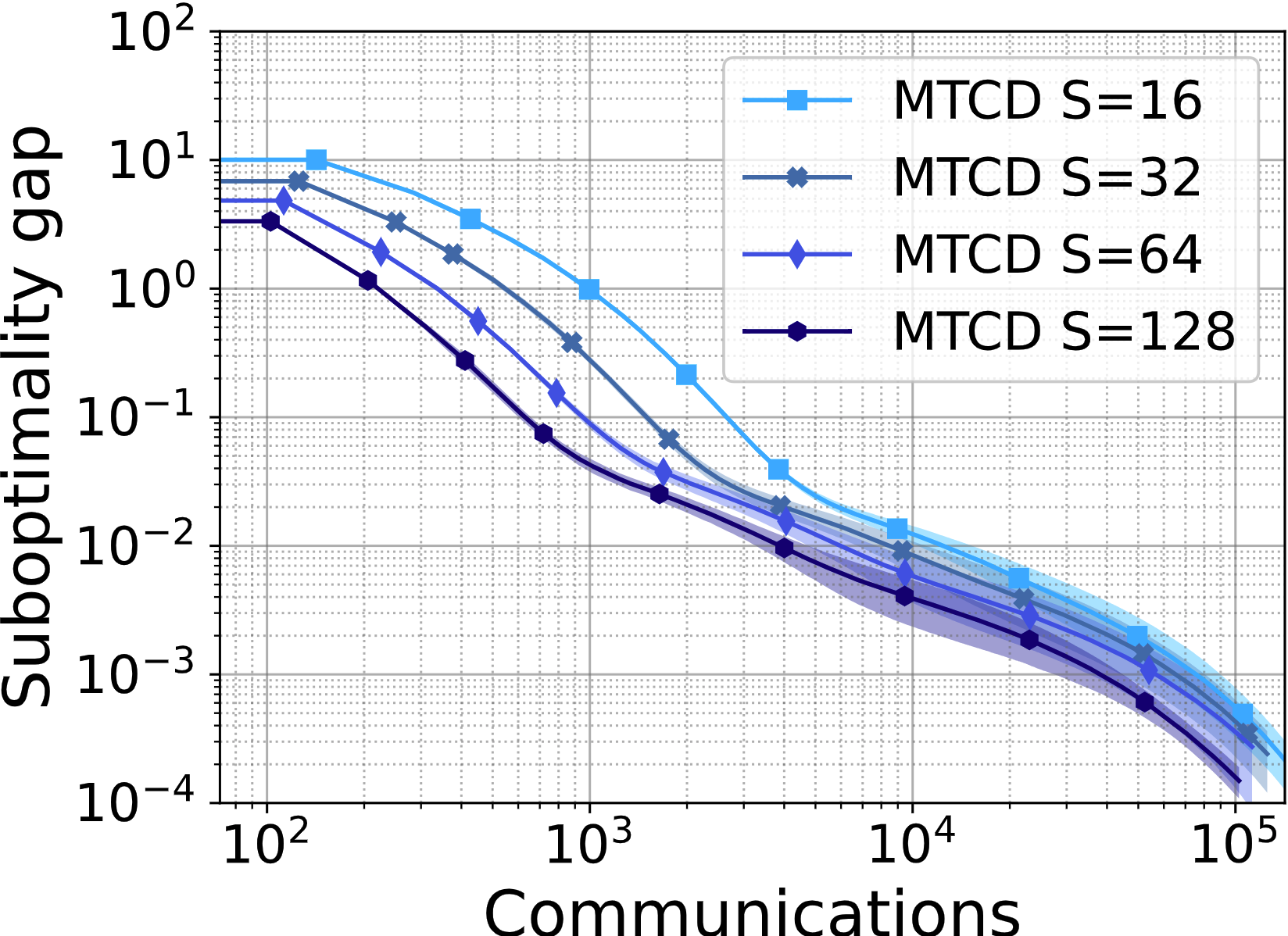}
		\hfil
		\raisebox{-1mm}{\includegraphics[width=0.25\textwidth]{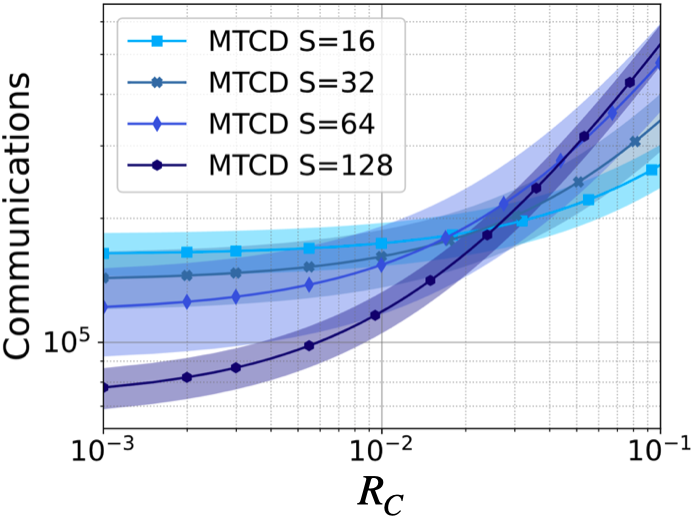}}
		\caption{
			We perform ridge regression on a path graph with $K=80$ nodes. We plot the suboptimality per iteration, the suboptimality per communication, and the number of communications needed to reach a given suboptimality for a range of ratios~$R_C$. In the top row, \texttt{MTCD} with $S\to\infty$ has $\Gamma=1$ (that is, it corresponds to \texttt{STCD}) and \texttt{MTCD} with $S=64$ have $\Gamma=2$ tokens. In the bottom row, all instances of \texttt{MTCD} have $\Gamma=2$ tokens.
		}
		\label{fig:k80_algo_comparison}
	\end{figure*}
	
	\begin{figure*}[!t]
		\centering
		\subfloat[$20\times$]{\includegraphics[width=0.32\textwidth]{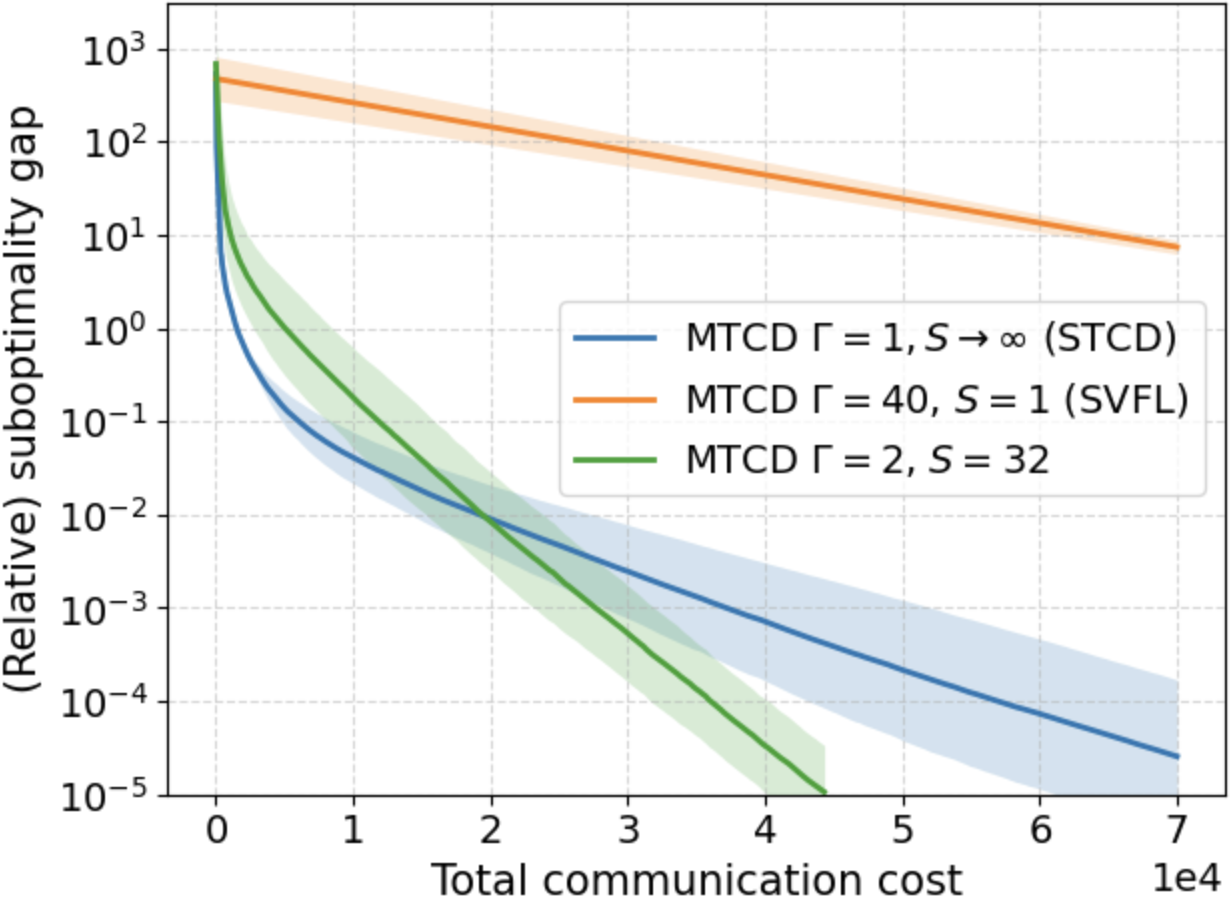}%
			\label{fig:my_plot_20x}}
		\hfil
		\subfloat[$10\times$]{\includegraphics[width=0.32\textwidth]{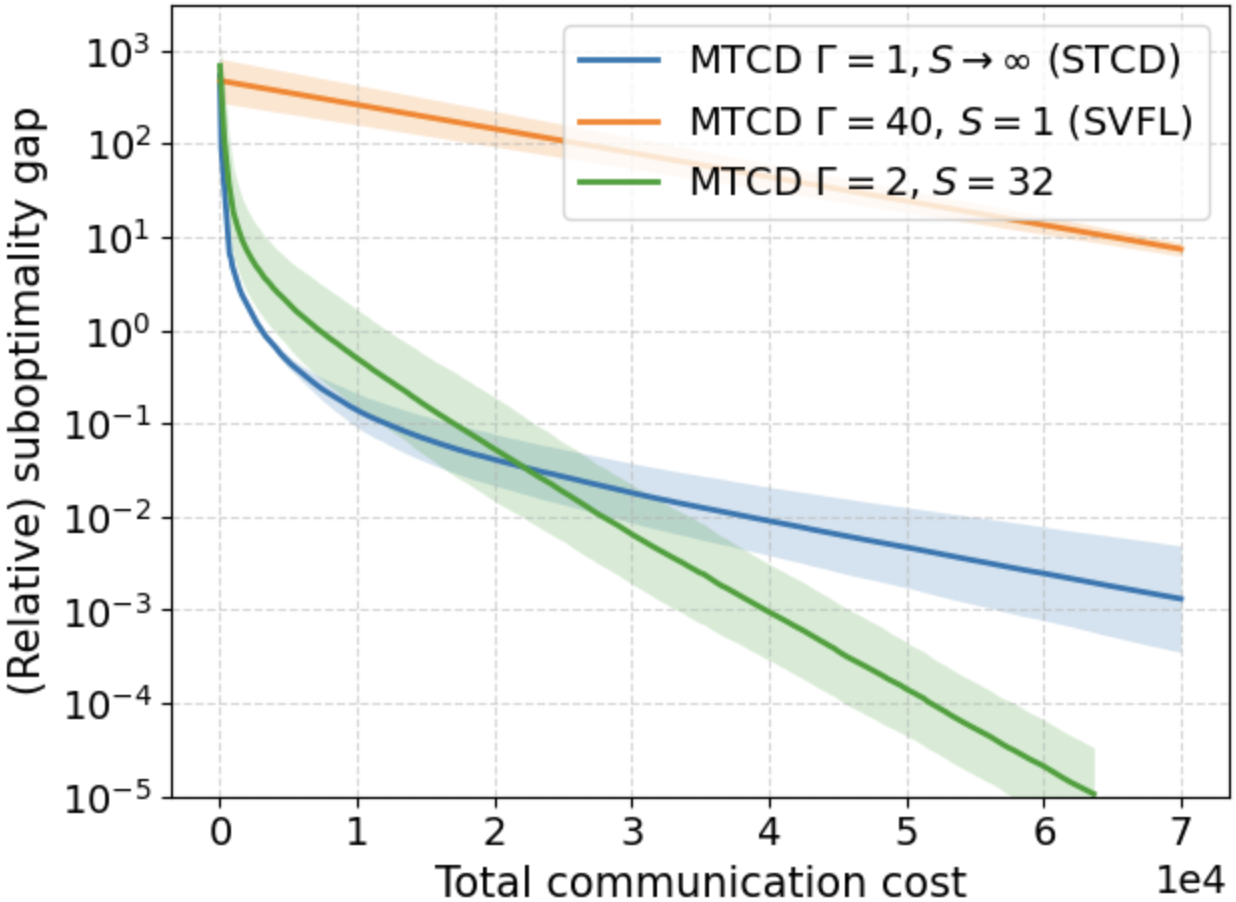}%
			\label{fig:my_plot_10x}}
		\hfil
		\subfloat[$5\times$]{\includegraphics[width=0.32\textwidth]{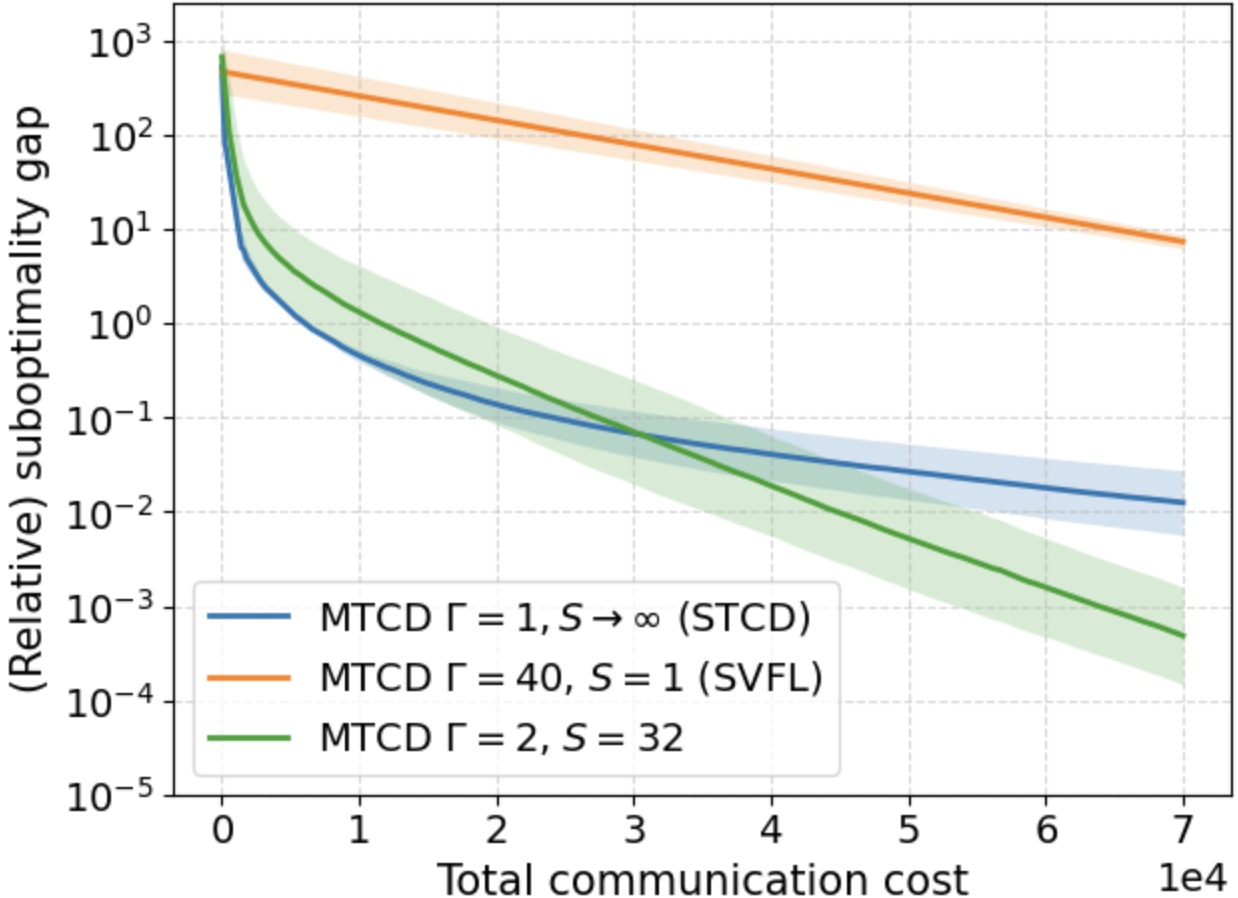}%
			\label{fig:my_plot_5x}}
		\caption{Ridge regression on a $K=40$ path graph, for a client–server communication cost of $20\times$, $10\times$, and $5\times$ larger than the client–client communication cost. (Single local update~$Q=1$.)}
		\label{fig:sdfl_best}
	\end{figure*}
	
	\begin{figure*}[!t]
		\centering
		\includegraphics[width=0.23\textwidth]{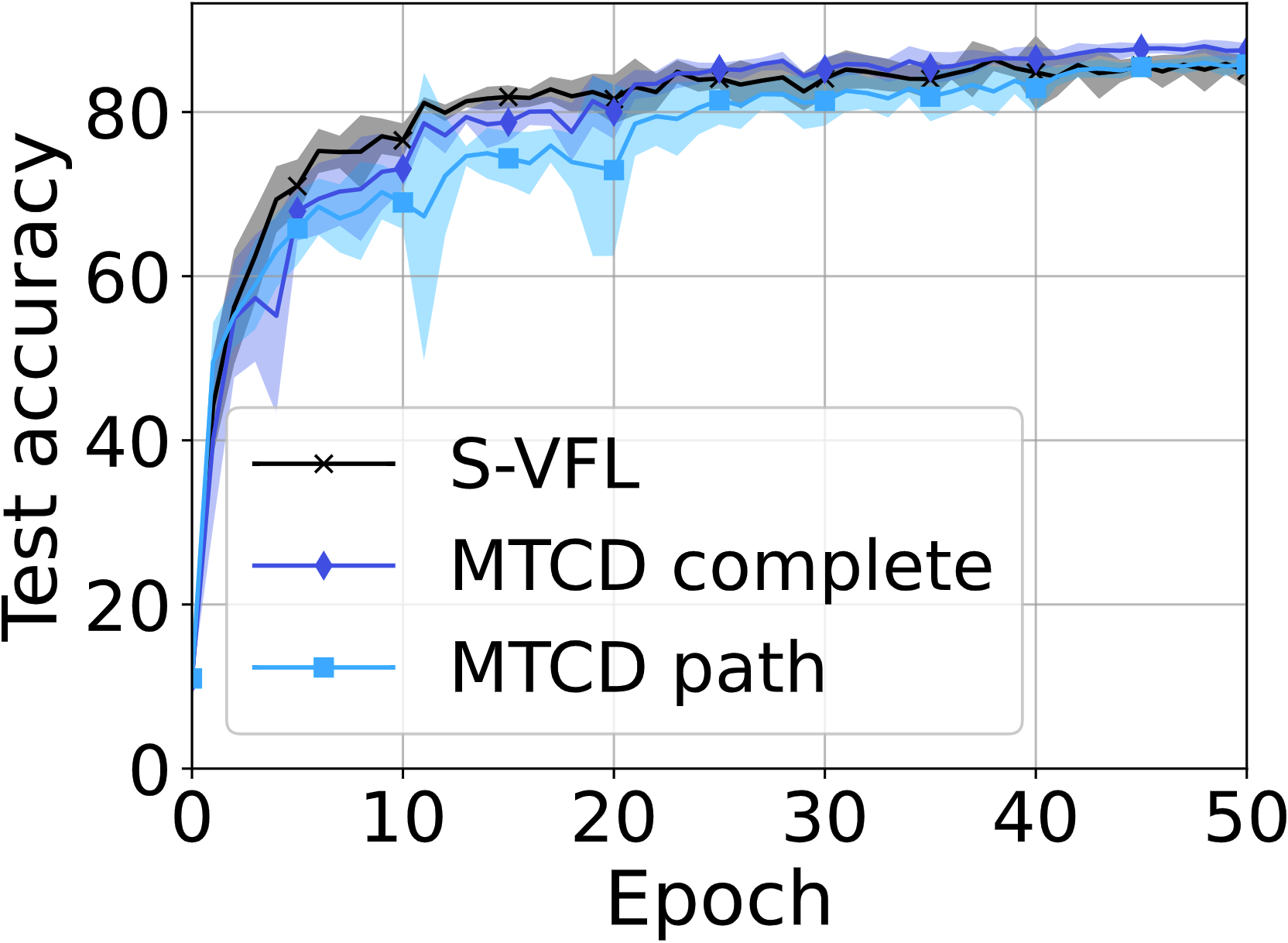}
		\hfil
		\includegraphics[width=0.23\textwidth]{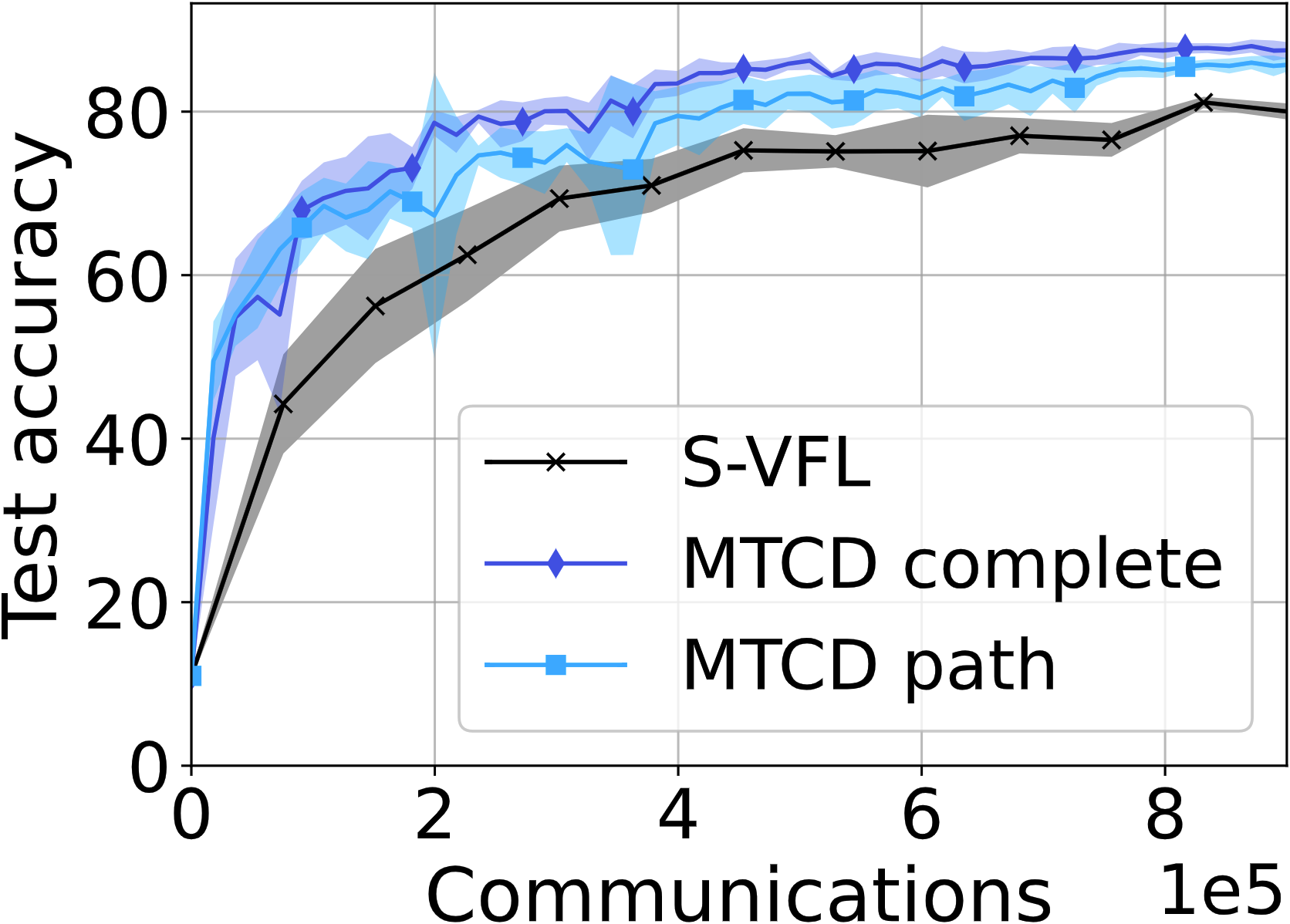}
		\hfil
		\includegraphics[width=0.23\textwidth]{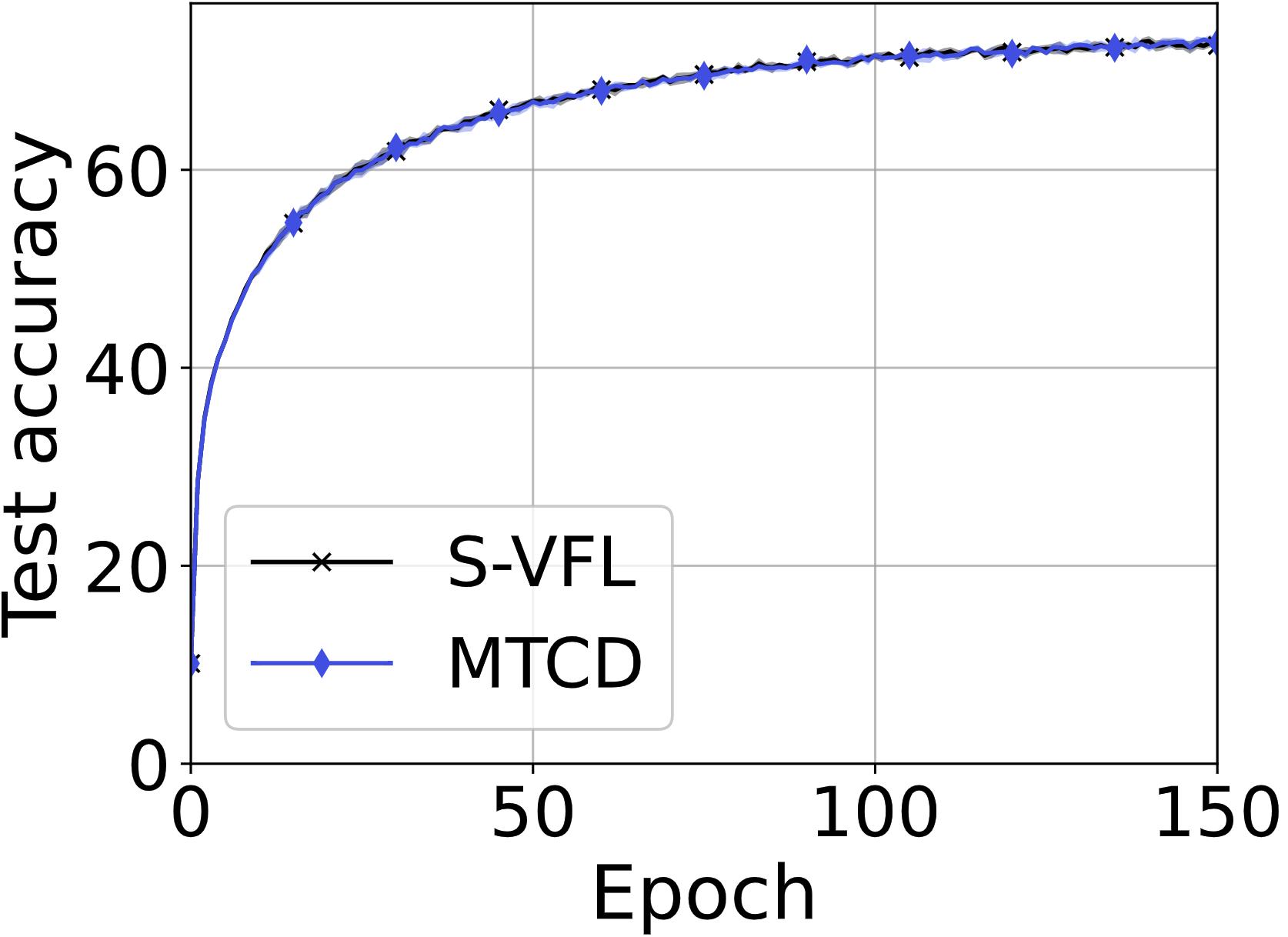}
		\hfil
		\includegraphics[width=0.23\textwidth]{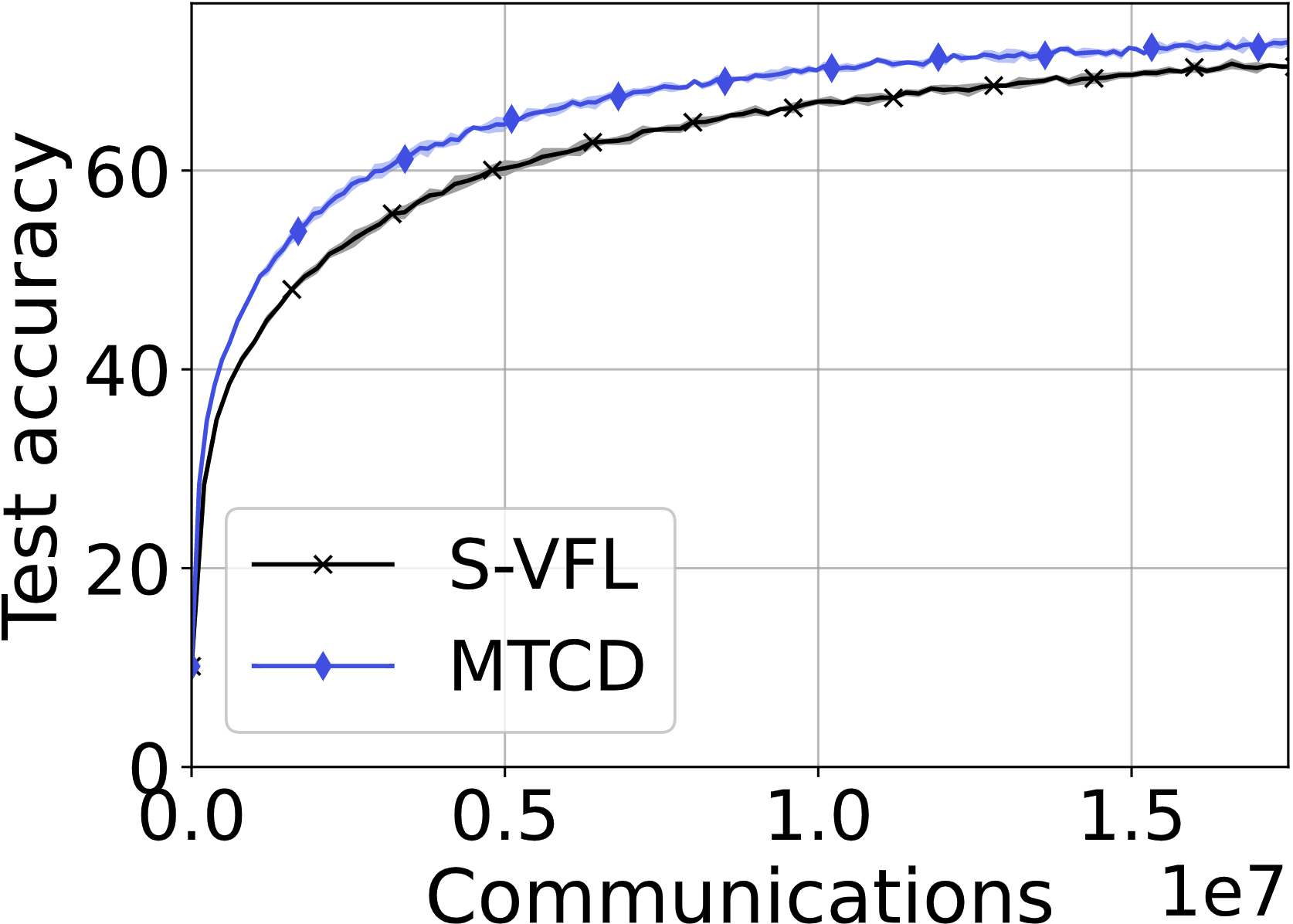}
		\caption{
			The plots on the left concern the training of an MVCNN on ModelNet10 and the ones on the right regard the training of a ResNet18 on CIFAR-10. 
		}
		\label{fig:nonconvex_plots}
	\end{figure*}
	
	\paragraph{Semi-decentralized setting.}
	We now consider the semi-decentralized setting, focusing on ridge regression and on path graphs, which are poorly-connected. The top row of Figure~\ref{fig:k80_algo_comparison} compares \texttt{MTCD} with the baselines. For \texttt{MTCD}, we assume that $P_\gamma$ is a uniform distribution over the clients and allow for overlapping token trajectories.
	
	The two leftmost plots show that, despite its slower convergence per iteration, \texttt{MTCD} is has a lower communication cost than the baselines (for $R_C=100$). Most interestingly, the plot on the right illustrates the communication cost required to achieve a suboptimality gap of $10^{-4}$ for various values of $R_C$---keeping the cost of client-server communications constant and varying that of client-client communications---demonstrating that the method that performs the best varies with $R_C$.
	In the bottom row, we present similar results, now comparing various \texttt{MTCD} instances with different numbers of hops $S$. As expected, we see that increasing the syncing frequency of \texttt{MTCD} leads to faster convergence, yet this comes at the cost of higher communication costs.
	
	In short, Figure~\ref{fig:k80_algo_comparison} highlights the advantage of having a flexible method enabling us to select the degree of reliance on the server depending on the application, and confirms that \texttt{MTCD} achieves this.
	
	In Figure~\ref{fig:sdfl_best}, we show the suboptimality gap with respect to the communication cost for the different methods, for different ratios of client-server to client-client communication cost (representing different application scenarios). These results demonstrate that the semi-decentralized setting can outperform both client-server and fully decentralized VFL for a range of setups.
	We present ablation studies in Appendix~\ref{app:ablations}.
	
	\subsection{Neural network training}
	
	We train an MVCNN~\citep{su2015multiview} on ModelNet10~\citep{wu2015}, a dataset of 3D CAD models. We consider $12$ clients split into two clusters with six clients each. Each client captures a different (2D) view of an object. We run \texttt{MTCD} for both complete graphs and path graphs, both with $S=6$. For \texttt{SVFL}, $\eta=0.001$. For \texttt{MTCD}, we use $\eta=0.001$ for the complete graph and $\eta=0.0005$ for the path graph, halving the learning rate every 20 epochs in both cases. For both \texttt{MTCD} and \texttt{SVFL}, $Q=10$ and $B=64$.
	
	We also train a ResNet18~\citep{he2015deep} on CIFAR-10~\citep{krizhevsky2009learning}. In this experiment, we consider $K=4$ clients split into two clusters, each with 2 clients. For both \texttt{SVFL} and \texttt{MTCD}, we have $\eta=0.0001$, $Q=10$, and $B=100$. For \texttt{MTCD}, we have $S=2$
	
	Figure~\ref{fig:nonconvex_plots} presents the results for the ModelNet10 and CIFAR-10 experiments, both in the token-per-cluster setting.
	In both experiments, we observe that \texttt{MTCD} obtains a similar performance to that of \texttt{SVFL} in terms of convergence per iteration, yet it achieves a lower communication cost.
	
	\section{Conclusions} \label{sec:conclusions}

	We formalize the semi-decentralized multi-token setup and propose \texttt{MTCD}, a multi-token method for semi-decentralized VFL. We show, both empirically and analytically, that, by leveraging both client-server and client-client communications, \texttt{MTCD} outperforms decentralized and client-server baselines in a range of relevant setups. A natural extension to this work is to combine \texttt{MTCD} with compression and privacy mechanisms, such as error-feedback compression and differential privacy.
	
	
	\section*{Acknowledgements}
	
	This work is supported in part by the Fundação para a Ciência e a Tecnologia through the Carnegie Mellon Portugal Program; by the grants U.S. National Science Foundation CCF-2007911 and ECCS-2318441/2537189; by the CMU-Portugal project CMU/TIC/0016/2021; by NOVA LINCS ref. UIDB/04516/2020 and ref. UIDP/04516/2020; by LARSyS FCT funding (DOI: 10.54499/LA/P/0083/2020); by PT Smart Retail project [PRR - 02/C05-i11/2024.C645440011-00000062], through IAPMEI - Agência para a Competitividade e Inovação; and by TaRDIS Horizon2020 Contract ID: 101093006.

	\bibliography{mtcd}

@article{polyak1963gradient,
	title={Gradient methods for the minimisation of functionals},
	author={Polyak, Boris T},
	journal={USSR Computational Mathematics and Mathematical Physics},
	volume={3},
	number={4},
	pages={864--878},
	year={1963},
	publisher={Elsevier}
}

@misc{ganguli2024faulttolerantserverlessvfl,
	title={Fault Tolerant Serverless VFL Over Dynamic Device Environment}, 
	author={Surojit Ganguli and Zeyu Zhou and Christopher G. Brinton and David I. Inouye},
	year={2024},
	eprint={2312.16638},
	archivePrefix={arXiv},
	primaryClass={cs.LG},
	url={https://arxiv.org/abs/2312.16638}, 
}

@ARTICLE{wang2025communication,
  author={Wang, He and Chi, Yuejie},
  journal={IEEE Transactions on Signal and Information Processing over Networks}, 
  title={Communication-Efficient Federated Optimization Over Semi-Decentralized Networks}, 
  year={2025},
  volume={11},
  number={},
  pages={147-160}
  }

@inproceedings{
	valdeira2025vertical,
	title={Vertical Federated Learning with Missing Features During Training and Inference},
	author={Pedro Valdeira and Shiqiang Wang and Yuejie Chi},
	booktitle={The Thirteenth International Conference on Learning Representations},
	year={2025},
	url={https://openreview.net/forum?id=OXi1FmHGzz}
}

@article{yang2023survey,
	title={A survey on vertical federated learning: From a layered perspective},
	author={Yang, Liu and Chai, Di and Zhang, Junxue and Jin, Yilun and Wang, Leye and Liu, Hao and Tian, Han and Xu, Qian and Chen, Kai},
	journal={arXiv preprint arXiv:2304.01829},
	year={2023}
}

@article{valdeira2025communication,
	title={Communication-efficient vertical federated learning via compressed error feedback},
	author={Valdeira, Pedro and Xavier, Jo{\~a}o and Soares, Cl{\'a}udia and Chi, Yuejie},
	journal={IEEE Transactions on Signal Processing},
	year={2025},
	publisher={IEEE}
}

@article{sharma2021machine,
	title={Machine learning in wireless sensor networks for smart cities: a survey},
	author={Sharma, Himanshu and Haque, Ahteshamul and Blaabjerg, Frede},
	journal={Electronics},
	volume={10},
	number={9},
	pages={1012},
	year={2021},
	publisher={Multidisciplinary Digital Publishing Institute}
}

@article{cheng2020federated,
	title={Federated learning for privacy-preserving AI},
	author={Cheng, Yong and Liu, Yang and Chen, Tianjian and Yang, Qiang},
	journal={Communications of the ACM},
	volume={63},
	number={12},
	pages={33--36},
	year={2020},
	publisher={ACM New York, NY, USA}
}

@incollection{robbins1971convergence,
	title={A convergence theorem for non negative almost supermartingales and some applications},
	author={Robbins, Herbert and Siegmund, David},
	booktitle={Optimizing methods in statistics},
	pages={233--257},
	year={1971},
	publisher={Elsevier}
}

@inproceedings{hui2007distributed,
	title={Distributed community detection in delay tolerant networks},
	author={Hui, Pan and Yoneki, Eiko and Chan, Shu Yan and Crowcroft, Jon},
	booktitle={Proceedings of 2nd ACM/IEEE international workshop on Mobility in the evolving internet architecture},
	pages={1--8},
	year={2007}
}

@inproceedings{
	valdeira2022a,
	title={A Multi-Token Coordinate Descent Method for Vertical Federated Learning},
	author={Pedro Valdeira and Yuejie Chi and Claudia Soares and Joao Xavier},
	booktitle={NeurIPS Workshop on Federated Learning: Recent Advances and New Challenges},
	year={2022}
}

@ARTICLE{Wu2018f,
	author={Wu, Tianyu and Yuan, Kun and Ling, Qing and Yin, Wotao and Sayed, Ali H.},
	journal={IEEE Transactions on Signal and Information Processing over Networks}, 
	title={Decentralized Consensus Optimization With Asynchrony and Delays}, 
	year={2018},
	volume={4},
	number={2},
	pages={293-307},
	doi={10.1109/TSIPN.2017.2695121}}

@article{yemini2023robust,
	title={Robust Semi-Decentralized Federated Learning via Collaborative Relaying},
	author={Yemini, Michal and Saha, Rajarshi and Ozfatura, Emre and G{\"u}nd{\"u}z, Deniz and Goldsmith, Andrea J},
	journal={IEEE Transactions on Wireless Communications},
	year={2023},
	publisher={IEEE}
}

@article{mota2013d,
	title={{D-ADMM}: A communication-efficient distributed algorithm for separable optimization},
	author={Mota, Joao FC and Xavier, Joao MF and Aguiar, Pedro MQ and P{\"u}schel, Markus},
	journal={IEEE Transactions on Signal processing},
	volume={61},
	number={10},
	pages={2718--2723},
	year={2013},
	publisher={IEEE}
}

@article{smith2018cocoa,
	title={{CoCoA}: A general framework for communication-efficient distributed optimization},
	author={Smith, Virginia and Forte, Simone and Ma, Chenxin and Tak{\'a}{\v{c}}, Martin and Jordan, Michael I and Jaggi, Martin},
	journal={Journal of Machine Learning Research},
	volume={18},
	number={230},
	pages={1--49},
	year={2018}
}

@article{zhang2021hybrid,
	title={Hybrid federated learning: Algorithms and implementation},
	author={Zhang, Xinwei and Yin, Wotao and Hong, Mingyi and Chen, Tianyi},
	journal={arXiv preprint arXiv:2012.12420},
	year={2020}
}

@article{lin2021semi,
  title={Semi-decentralized federated learning with cooperative D2D local model aggregations},
  author={Lin, Frank Po-Chen and Hosseinalipour, Seyyedali and Azam, Sheikh Shams and Brinton, Christopher G and Michelusi, Nicolo},
  journal={IEEE Journal on Selected Areas in Communications},
  volume={39},
  number={12},
  pages={3851--3869},
  year={2021},
  publisher={IEEE}
}

@inproceedings{alon2008many,
	title={Many random walks are faster than one},
	author={Alon, Noga and Avin, Chen and Koucky, Michal and Kozma, Gady and Lotker, Zvi and Tuttle, Mark R},
	booktitle={Proceedings of the twentieth annual symposium on parallelism in algorithms and architectures},
	pages={119--128},
	year={2008}
}

@misc{liu2022vertical,
	title={Vertical Federated Learning: Concepts, Advances and Challenges}, 
	author={Yang Liu and Yan Kang and Tianyuan Zou and Yanhong Pu and Yuanqin He and Xiaozhou Ye and Ye Ouyang and Ya-Qin Zhang and Qiang Yang},
	year={2023},
	eprint={2211.12814},
	archivePrefix={arXiv},
	primaryClass={cs.LG}
}

@inproceedings{su2015multiview,
	title={Multi-view convolutional neural networks for 3{D} shape recognition},
	author={Su, Hang and Maji, Subhransu and Kalogerakis, Evangelos and Learned-Miller, Erik},
	booktitle={Proceedings of the IEEE international conference on computer vision},
	year={2015}
}

@inproceedings{he2015deep,
	title={Deep residual learning for image recognition},
	author={He, Kaiming and Zhang, Xiangyu and Ren, Shaoqing and Sun, Jian},
	booktitle={Proceedings of the IEEE conference on computer vision and pattern recognition},
	year={2016}
}

@article{krizhevsky2009learning,
	title={Learning multiple layers of features from tiny images},
	author={Krizhevsky, Alex and Hinton, Geoffrey and others},
	year={2009},
	publisher={Toronto, ON, Canada}
}

@inproceedings{wu2015,
	title={3{D} shapenets: A deep representation for volumetric shapes},
	author={Wu, Zhirong and Song, Shuran and Khosla, Aditya and Yu, Fisher and Zhang, Linguang and Tang, Xiaoou and Xiao, Jianxiong},
	booktitle={Proceedings of the IEEE conference on computer vision and pattern recognition},
	year={2015}
}

@article{li2020communicationefficient, author = {Li, Boyue and Cen, Shicong and Chen, Yuxin and Chi, Yuejie}, title = {Communication-efficient distributed optimization in networks with gradient tracking and variance reduction}, year = {2020}, issue_date = {January 2020}, publisher = {JMLR.org}, volume = {21}, number = {1}, issn = {1532-4435}, journal = {J. Mach. Learn. Res.}, month = {jan}, articleno = {180}, numpages = {51} }

@article{ceballos2020splitnn,
	title={SplitNN-driven vertical partitioning},
	author={Ceballos, Iker and Sharma, Vivek and Mugica, Eduardo and Singh, Abhishek and Roman, Alberto and Vepakomma, Praneeth and Raskar, Ramesh},
	journal={arXiv preprint arXiv:2008.04137},
	year={2020}
}

@misc{paulin2015,
	title={Concentration inequalities for {M}arkov chains by {M}arton couplings and spectral methods}, 
	author={Daniel Paulin},
	year={2018},
	eprint={1212.2015},
	archivePrefix={arXiv},
	primaryClass={math.PR}
}

@ARTICLE{Alghunaim2021,
	author={Alghunaim, Sulaiman A. and Lyu, Qi and Yan, Ming and Sayed, Ali H.},
	journal={IEEE Transactions on Signal Processing}, 
	title={Dual Consensus Proximal Algorithm for Multi-Agent Sharing Problems}, 
	year={2021},
	volume={69},
	number={},
	pages={5568-5579}
}

@inproceedings{Zhao2022,
	author = {Zhao, Haoyu and Li, Boyue and Li, Zhize and Richtarik, Peter and Chi, Yuejie},
	booktitle = {Advances in Neural Information Processing Systems},
	editor = {S. Koyejo and S. Mohamed and A. Agarwal and D. Belgrave and K. Cho and A. Oh},
	pages = {31653--31667},
	publisher = {Curran Associates, Inc.},
	title = {{BEER}: Fast {O}(1/T) Rate for Decentralized Nonconvex Optimization with Communication Compression},
	volume = {35},
	year = {2022}
}

@article{He2018,
	title={{COLA}: Decentralized linear learning},
	author={He, Lie and Bian, An and Jaggi, Martin},
	journal={Advances in Neural Information Processing Systems},
	volume={31},
	year={2018}
}

@inproceedings{Hendrikx2022,
	title={A principled framework for the design and analysis of token algorithms},
	author={Hendrikx, Hadrien},
	booktitle={International Conference on Artificial Intelligence and Statistics},
	pages={470--489},
	year={2023},
	organization={PMLR}
}

@article{Chen2020,
	author    = {Tianyi Chen and
	Xiao Jin and
	Yuejiao Sun and
	Wotao Yin},
	title     = {{VAFL:} a Method of Vertical Asynchronous Federated Learning},
	journal = {arXiv:2007.06081},
	year      = {2020},
}

@article{Beck2013,
	author = {Beck, Amir and Tetruashvili, Luba},
	title = {On the Convergence of Block Coordinate Descent Type Methods},
	journal = {SIAM Journal on Optimization},
	volume = {23},
	number = {4},
	pages = {2037-2060},
	year = {2013},
}

@article{Wright2015,
	year = {2015},
	publisher = {Springer Science and Business Media {LLC}},
	volume = {151},
	number = {1},
	pages = {3--34},
	author = {Stephen J. Wright},
	title = {Coordinate descent algorithms},
	journal = {Mathematical Programming}
}

@article{Richtrik2012,
	year = {2012},
	publisher = {Springer Science and Business Media {LLC}},
	volume = {144},
	number = {1-2},
	pages = {1--38},
	author = {Peter Richt{\'{a}}rik and Martin Tak{\'{a}}{\v{c}}},
	title = {Iteration complexity of randomized block-coordinate descent methods for minimizing a composite function},
	journal = {Mathematical Programming}
}

@article{Nesterov2012,
	year = {2012},
	publisher = {Society for Industrial {\&} Applied Mathematics ({SIAM})},
	volume = {22},
	number = {2},
	pages = {341--362},
	author = {Yurii Nesterov},
	title = {Efficiency of Coordinate Descent Methods on Huge-Scale Optimization Problems},
	journal = {{SIAM} Journal on Optimization}
}

@article{Fercoq2015,
	author = {Fercoq, Olivier and Richt\'{a}rik, Peter},
	title = {Accelerated, Parallel, and Proximal Coordinate Descent},
	journal = {SIAM Journal on Optimization},
	volume = {25},
	number = {4},
	year = {2015},
}

@article{Sun2019,
	year = {2019},
	publisher = {Springer Science and Business Media {LLC}},
	volume = {75},
	number = {1},
	pages = {35--61},
	author = {Tao Sun and Yuejiao Sun and Yangyang Xu and Wotao Yin},
	title = {{M}arkov chain block coordinate descent},
	journal = {Computational Optimization and Applications}
}

@article{Guyon2004,
	title={Result Analysis of the {NIPS} 2003 Feature Selection Challenge},
	author={Guyon, Isabelle and Gunn, Steve and Ben-Hur, Asa and Dror, Gideon},
	journal={Advances in neural information processing systems},
	volume={17},
	year={2004}
}

@article{ram2009,
	title={Incremental stochastic subgradient algorithms for convex optimization},
	author={Ram, S Sundhar and Nedi{\'c}, A and Veeravalli, Venugopal V},
	journal={SIAM Journal on Optimization},
	volume={20},
	number={2},
	pages={691--717},
	year={2009},
	publisher={SIAM}
}

@article{Liu2022,
	year = {2022},
	publisher = {Institute of Electrical and Electronics Engineers ({IEEE})},
	volume = {70},
	pages = {4277--4290},
	author = {Yang Liu and Xinwei Zhang and Yan Kang and Liping Li and Tianjian Chen and Mingyi Hong and Qiang Yang},
	title = {{FedBCD}: A Communication-Efficient Collaborative Learning Framework for Distributed Features},
	journal = {{IEEE} Transactions on Signal Processing}
}

@article{Mao2020,
	year = {2020},
	publisher = {Institute of Electrical and Electronics Engineers ({IEEE})},
	volume = {68},
	pages = {2513--2528},
	author = {Xianghui Mao and Kun Yuan and Yubin Hu and Yuantao Gu and Ali H. Sayed and Wotao Yin},
	title = {Walkman: A Communication-Efficient Random-Walk Algorithm for Decentralized Optimization},
	journal = {{IEEE} Transactions on Signal Processing}
}

@inproceedings{Castiglia2022,
	title={Compressed-{VFL}: Communication-Efficient Learning with Vertically Partitioned Data},
	author={Castiglia, Timothy J and Das, Anirban and Wang, Shiqiang and Patterson, Stacy},
	booktitle={International Conference on Machine Learning},
	pages={2738--2766},
	year={2022},
	organization={PMLR}
}

@article{Lin2021,
	year = {2021},
	publisher = {Institute of Electrical and Electronics Engineers ({IEEE})},
	volume = {39},
	number = {12},
	pages = {3851--3869},
	author = {Frank Po-Chen Lin and Seyyedali Hosseinalipour and Sheikh Shams Azam and Christopher G. Brinton and Nicolo Michelusi},
	title = {Semi-Decentralized Federated Learning With Cooperative {D2D} Local Model Aggregations},
	journal = {{IEEE} Journal on Selected Areas in Communications}
}

@article{Chen2022,
	title={Asynchronous parallel incremental block-coordinate descent for decentralized machine learning},
	author={Chen, Hao and Ye, Yu and Xiao, Ming and Skoglund, Mikael},
	journal={IEEE Transactions on Big Data},
	year={2022},
	publisher={IEEE}
}

@inproceedings{guo2022,
	title={Hybrid Local {SGD} for Federated Learning with Heterogeneous Communications},
	author={Guo, Yuanxiong and Sun, Ying and Hu, Rui and Gong, Yanmin},
	booktitle={International Conference on Learning Representations},
	year={2021}
}

@inproceedings{McMahan2017,
	title={Communication-efficient learning of deep networks from decentralized data},
	author={McMahan, Brendan and Moore, Eider and Ramage, Daniel and Hampson, Seth and y Arcas, Blaise Aguera},
	booktitle={Artificial intelligence and statistics},
	pages={1273--1282},
	year={2017},
	organization={PMLR}
}

@ARTICLE{Nedic2009,
	author={Nedic, Angelia and Ozdaglar, Asuman},
	journal={IEEE Transactions on Automatic Control},
	title={Distributed Subgradient Methods for Multi-Agent Optimization},
	year={2009},
	volume={54},
	number={1},
	pages={48-61},
}

@inproceedings{Koloskova2020,
	title={A Unified Theory of Decentralized {SGD} with Changing Topology and Local Updates},
	author={Koloskova, Anastasia and Loizou, Nicolas and Boreiri, Sadra and Jaggi, Martin and Stich, Sebastian},
	booktitle={International Conference on Machine Learning},
	pages={5381--5393},
	year={2020},
	organization={PMLR}
}

@article{Duchi2012,
	year = 2012,
	publisher = {Institute of Electrical and Electronics Engineers ({IEEE})},
	volume = {57},
	number = {3},
	pages = {592--606},
	author = {J. C. Duchi and A. Agarwal and M. J. Wainwright},	
	title = {Dual Averaging for Distributed Optimization: Convergence Analysis and Network Scaling},
	journal = {{IEEE} Transactions on Automatic Control}
}

@article{Qu2018,
	year = 2018,
	publisher = {Institute of Electrical and Electronics Engineers ({IEEE})},
	volume = {5},
	number = {3},
	pages = {1245--1260},
	author = {Guannan Qu and Na Li},
	title = {Harnessing Smoothness to Accelerate Distributed Optimization},
	journal = {{IEEE} Transactions on Control of Network Systems}
}

@article{Bian2022,
	author = {Bian, Jieming and Xu, Jie},
	title = {Mobility Improves the Convergence of Asynchronous Federated Learning},
	journal = {arXiv:2206.04742},
	year = {2022}
}

@INPROCEEDINGS{Zhang2021,
	author={Zhang, Xueqing and Liu, Yanwei and Liu, Jinxia and Argyriou, Antonios and Han, Yanni},
	booktitle={2021 IEEE Wireless Communications and Networking Conference (WCNC)}, 
	title={{D2D}-Assisted Federated Learning in Mobile Edge Computing Networks}, 
	year={2021}
}

@article{Hosseinalipour2022,
	year = 2022,
	publisher = {Institute of Electrical and Electronics Engineers ({IEEE})},
	volume = {30},
	number = {4},
	pages = {1569--1584},
	author = {Seyyedali Hosseinalipour and Sheikh Shams Azam and Christopher G. Brinton and Nicolo Michelusi and Vaneet Aggarwal and David J. Love and Huaiyu Dai},
	title = {Multi-Stage Hybrid Federated Learning Over Large-Scale {D2D}-Enabled Fog Networks},	
	journal = {{IEEE}/{ACM} Transactions on Networking}
}

@article{johansson2010,
	title={A randomized incremental subgradient method for distributed optimization in networked systems},
	author={Johansson, Bj{\"o}rn and Rabi, Maben and Johansson, Mikael},
	journal={SIAM Journal on Optimization},
	volume={20},
	number={3},
	pages={1157--1170},
	year={2010},
	publisher={SIAM}
}

@article{bertsekas1997,
	title={A new class of incremental gradient methods for least squares problems},
	author={Bertsekas, Dimitri P},
	journal={SIAM Journal on Optimization},
	volume={7},
	number={4},
	pages={913--926},
	year={1997},
	publisher={SIAM}
}

@article{nedic2001,
	title={Incremental subgradient methods for nondifferentiable optimization},
	author={Nedic, Angelia and Bertsekas, Dimitri P},
	journal={SIAM Journal on Optimization},
	volume={12},
	number={1},
	pages={109--138},
	year={2001},
	publisher={SIAM}
}

@ARTICLE{Ye2020,
	author={Ye, Yu and Chen, Hao and Ma, Zheng and Xiao, Ming},
	journal={IEEE Communications Letters}, 
	title={Decentralized Consensus Optimization Based on Parallel Random Walk}, 
	year={2020},
	volume={24},
	number={2},
	pages={391-395}
}

@article{diamond2016,
	author  = {Steven Diamond and Stephen Boyd},
	title   = {{CVXPY}: {A} {P}ython-embedded modeling language for convex optimization},
	journal = {Journal of Machine Learning Research},
	year    = {2016},
	volume  = {17},
	number  = {83},
	pages   = {1--5},
}

@article{Lian2017,
	title={Can decentralized algorithms outperform centralized algorithms? {A} case study for decentralized parallel stochastic gradient descent},
	author={Lian, Xiangru and Zhang, Ce and Zhang, Huan and Hsieh, Cho-Jui and Zhang, Wei and Liu, Ji},
	journal={Advances in Neural Information Processing Systems},
	volume={30},
	year={2017}
}

@article{Nedic2018,
	year = {2018},
	publisher = {Institute of Electrical and Electronics Engineers ({IEEE})},
	volume = {106},
	number = {5},
	pages = {953--976},
	author = {Angelia Nedic and Alex Olshevsky and Michael G. Rabbat},
	title = {Network Topology and Communication-Computation Tradeoffs in Decentralized Optimization},
	journal = {Proceedings of the {IEEE}}
}
	
	\appendix
	
	\section{Proofs of the lemmas}
	\label{sec:proof-lemmas}
	
	Throughout the proofs, we slightly abuse notation for the sake of conciseness, denoting $\bm{\theta}_{c,:}$ as $\bm{\theta}_c$. (This collides with $\bm{\theta}_k$ in main paper, but it should still be clear which block we refer to when using each.)
	
	We use the following inequality, which follows from the Cauchy–Schwarz inequality, in our results:
	\begin{equation} \label{eq:youngs_inequality}
		\left\lVert
		\bm{v}_1+\cdots+\bm{v}_n
		\right\rVert^2
		\leq
		n
		\left(\left\lVert
		\bm{v}_1
		\right\rVert^2
		+\cdots+\left\lVert
		\bm{v}_1
		\right\rVert^2\right)
		.
	\end{equation}
	
	\subsection{Proof of Lemma~\ref{lemma:inner_product_bound}}
	\label{sec:proof-lemma1}
	
	We start by using the fact that the difference at block~$c$ between consecutive iterates, $\bm{\theta}_c^{t+1}-\bm{\theta}_c^{t}$, is given by the sum of the gradient-based updates used from $(t,0)$ to $(t,P-1)$:
	\[
	\begin{split}
		\mathbb{E}_{t^+}
		\left[\nabla_c f(\bm{\theta}^{t})^\top(\bm{\theta}_c^{t+1}-\bm{\theta}_c^{t})\right]
		&\;\;=
		\mathbb{E}_{t^+}
		\left[(\bm{\Pi}_c\nabla f(\bm{\theta}^{t}))^\top
		(\bm{\theta}^{t,P}(c)-\bm{\theta}^{t,0}(c)) \right]
		\\
		&\;\;=
		{\textstyle
			\mathbb{E}_{t^+}
			\left[
			\nabla f(\bm{\theta}^{t})^\top\bm{\Pi}_c
			\left(\sum_{p=0}^{P-1} \bm{\theta}^{t,p+1}(c)-\bm{\theta}^{t,p}(c)\right)\right]
		}
		\\
		&\;\;=
		{\textstyle
			\mathbb{E}_{t^+}
			\left[\nabla f(\bm{\theta}^{t})^\top\bm{\Pi}_c
			\left(\sum_{p=0}^{P-1} -\eta \bm{\Pi}_{c,i_c^{t,p}} \tilde{\nabla} f (\bm{\theta}^{t,p}(c);\mathcal{B}^{t}) \right) \right]
		}
		\\
		&\;\;=
		{\textstyle
			-\eta
			\sum_{p=0}^{P-1}
			\nabla f(\bm{\theta}^{t})^\top
			\bm{\Pi}_{c,i_c^{t,p}} \mathbb{E}_{t^+}
			\left[\tilde{\nabla} f (\bm{\theta}^{t,p}(c);\mathcal{B}^{t})
			\right]
		}
		\\
		&\;\,\stackrel{(a)}{=}
		{\textstyle
			-\eta
			\sum_{p=0}^{P-1}
			\nabla f(\bm{\theta}^{t})^\top
			\bm{\Pi}_{c,i_c^{t,p}} \nabla f (\bm{\theta}^{t,p}(c))
			,
		}
	\end{split}
	\]
	where $(a)$ follows from~\eqref{eq:unbiased}.
	
	We now separate the $p=0$ term from the remaining, $p>0$, terms of the sum, which correspond to updates using stale information, and use a polarization identity on the latter:
	\begin{equation} \label{eq:inner_product_decoupled}
		\begin{split}
			\mathbb{E}_{t^+}
			\left[\nabla_c f(\bm{\theta}^{t})^\top(\bm{\theta}_c^{t+1}-\bm{\theta}_c^{t})\right]
			&=
			-\eta
			\nabla f(\bm{\theta}^{t})^\top
			\bm{\Pi}_{c,i_c^{t,0}} \nabla f (\bm{\theta}^{t})
			\\
			&\quad
			{\textstyle
				-\eta
				\sum_{p=1}^{P-1}
				\nabla f(\bm{\theta}^{t})^\top
				\bm{\Pi}_{c,i_c^{t,p}} \nabla f (\bm{\theta}^{t,p}(c))
			}
			\\
			&\stackrel{(a)}{=}
			-\eta
			\nabla f(\bm{\theta}^{t})^\top
			\bm{\Pi}_{c,i_c^{t,0}} \nabla f (\bm{\theta}^{t})
			\\
			&\quad
			{\textstyle
				-
				\frac{\eta}{2}
				\sum_{p=1}^{P-1}
				\nabla f(\bm{\theta}^{t})^\top
				\bm{\Pi}_{c,i_c^{t,p}}
				\nabla f(\bm{\theta}^{t})
			}
			\\
			&\quad
			{\textstyle
				-
				\frac{\eta}{2}
				\sum_{p=1}^{P-1}
				\nabla f(\bm{\theta}^{t,p}(c))^\top
				\bm{\Pi}_{c,i_c^{t,p}}
				\nabla f(\bm{\theta}^{t,p}(c))
			}
			\\
			&\quad
			{\textstyle
				+
				\frac{\eta}{2} \sum_{p=1}^{P-1} \lVert \bm{\Pi}_{c,i_c^{t,p}}\nabla f(\bm{\theta}^{t,p}(c)) - \bm{\Pi}_{c,i_c^{t,p}}\nabla f(\bm{\theta}^{t}) \rVert^2
			}
			\\
			&\stackrel{(b)}{=}
			-\eta
			\lVert\nabla f(\bm{\theta}^{t})\rVert_{\bm{\Pi}_{c,i_c^{t,0}}}^{2}
			\\
			&\quad
			{\textstyle
				-
				\frac{\eta}{2} \sum_{p=1}^{P-1} \lVert \nabla f(\bm{\theta}^{t}) \rVert_{\bm{\Pi}_{c,i_c^{t,p}}}^{2}
			}
			\\
			&\quad
			{\textstyle
				-
				\frac{\eta}{2} \sum_{p=1}^{P-1} \lVert \nabla f(\bm{\theta}^{t,p}(c)) \rVert_{\bm{\Pi}_{c,i_c^{t,p}}}^{2}
			}
			\\
			&\quad
			{\textstyle
				+
				\frac{\eta}{2} \sum_{p=1}^{P-1} \lVert \nabla f(\bm{\theta}^{t,p}(c)) - \nabla f(\bm{\theta}^{t}) \rVert_{\bm{\Pi}_{c,i_c^{t,p}}}^2,
			}
		\end{split}
	\end{equation}
	where we use $\bm{u}^\top \bm{v}=\frac{1}{2}\lVert\bm{u}\rVert^2+\frac{1}{2}\lVert\bm{v}\rVert^2-\frac{1}{2}\lVert\bm{u}-\bm{v}\rVert^2$ in $ (a) $ and $ \lVert \bm{u} \rVert^2_{\bm{A}}= \bm{u}^\top\bm{A}\bm{u}$ in $ (b) $. Let us upper bound the offset term:
	\[
	\begin{split}	
		\lVert \nabla f(\bm{\theta}^{t,p}(c)) - \nabla f(\bm{\theta}^{t}) \rVert_{\bm{\Pi}_{c,i_c^{t,p}}}^2
		&\;\stackrel{(a)}{\leq}
		\lVert \nabla f(\bm{\theta}^{t,p}(c)) - \nabla f(\bm{\theta}^{t}) \rVert^2
		\\
		&\;\stackrel{(b)}{\leq}
		L^2\lVert  \bm{\theta}^{t,p}(c) - \bm{\theta}^{t,0}(c) \rVert^2
		\\
		&\;\,=
		{\textstyle
			L^2\lVert  -\eta \bm{\Pi}_{c,i_c^{t,0}} \nabla f (\bm{\theta}^{t,0}(c)) -\eta\sum_{r}^{p-1} \bm{\Pi}_{c,i_c^{t,r}} \nabla f (\bm{\theta}^{t,r}(c)) \rVert^2,
		}
	\end{split}
	\]
	where $ (a) $ follows from $\bm{\Pi}_{c,i_c^{t,p}} \preceq \bm{I}$ and $ (b) $ from~\eqref{eq:L-smoothness}.
	
	\noindent
	Now, using~\eqref{eq:youngs_inequality} and the fact that $p\leq P-1$, we get that:
	\[
	\begin{split}
		\lVert \nabla f(\bm{\theta}^{t,p}(c)) - \nabla f(\bm{\theta}^{t}) \rVert_{\bm{\Pi}_{c,i_c^{t,p}}}^2
		&\leq
		L^2\eta^2p
		\Big(\lVert  \nabla f (\bm{\theta}^{t}) \rVert^2_{\bm{\Pi}_{c,i_c^{t,0}}}
		+ \sum_{r}^{p-1} \lVert  \nabla f (\bm{\theta}^{t,r}(c)) \rVert^2_{\bm{\Pi}_{c,i_c^{t,r}}}\Big)
		\\
		&\leq
		L^2\eta^2(P-1)
		\Big(\lVert  \nabla f (\bm{\theta}^{t}) \rVert^2_{\bm{\Pi}_{c,i_c^{t,0}}}
		+ \sum_{r}^{P-2} \lVert  \nabla f (\bm{\theta}^{t,r}(c)) \rVert^2_{\bm{\Pi}_{c,i_c^{t,r}}}\Big).
	\end{split}
	\]
	Plugging this inequality into~\eqref{eq:inner_product_decoupled}, we arrive at~\eqref{eq:inner_product_bound}.
	
	\subsection{Proof of Lemma~\ref{lemma:norm_bound}}
	\label{sec:proof-lemma2}
	
	{From the fact that $||\bm{\theta}_i||\leq||\bm{\theta}||$ and from~\eqref{eq:youngs_inequality}, we have that:
		\[
		\begin{split}
			\mathbb{E}_{t^+}
			\lVert\bm{\theta}_c^{t+1}-\bm{\theta}_c^{t}\rVert_{}^{2}
			&=
			\mathbb{E}_{t^+}
			\lVert\bm{\theta}_c^{t,P}(c)-\bm{\theta}_c^{t,0}(c)\rVert_{}^{2}
			\\
			&\leq
			\mathbb{E}_{t^+}
			\lVert\bm{\theta}^{t,P}(c)-\bm{\theta}^{t,0}(c)\rVert_{}^{2}
			\\
			&=
			{\textstyle
				\mathbb{E}_{t^+}
				\left\lVert
				\sum_{p=0}^{P-1}
				\bm{\theta}^{t,p+1}(c)-\bm{\theta}^{t,p}(c)
				\right\rVert_{}^{2}
			}
			\\
			&\leq
			{\textstyle
				P\cdot
				\sum_{p=0}^{P-1}
				\mathbb{E}_{t^+}
				\left\lVert
				\bm{\theta}^{t,p+1}(c)-\bm{\theta}^{t,p}(c)
				\right\rVert_{}^{2}
			}
			.
		\end{split}
		\]
		Now, from our gradient-based update,
		we get that:
		\[
		\begin{split}
\mathbb{E}_{t^+}
			\lVert\bm{\theta}_c^{t+1}-\bm{\theta}_c^{t}\rVert_{}^{2}
			&\leq
			{\textstyle
				P
				\sum_{p=0}^{P-1}
				\mathbb{E}_{t^+}
				\left\lVert
				- \eta \bm{\Pi}_{c,i_c^{t,p}} \tilde{\nabla} f (\bm{\theta}^{t,p}(c);\mathcal{B}^{t})
				\right\rVert_{}^{2}
			}
			\\
			&=
			{\textstyle
				\eta^2P
				\sum_{p=0}^{P-1}
				\mathbb{E}_{t^+}
				\left\lVert
				\tilde{\nabla} f (\bm{\theta}^{t,p}(c);\mathcal{B}^{t})
				\right\rVert_{\bm{\Pi}_{c,i_c^{t,p}}}^{2}
			}
			\\
			&\stackrel{(a)}{\leq}
			{\textstyle
				\eta^2P
				\sum_{p=0}^{P-1}
				\left[
				\left\lVert
				\nabla f (\bm{\theta}^{t,p}(c))
				\right\rVert_{\bm{\Pi}_{c,i_c^{t,p}}}^{2}
				+
				\frac{\sigma^2}{B}
				\right]
			}
			\\
			&=
			\eta^2P
			\Big(
			\frac{\sigma^2P}{B}
			+
			\lVert \nabla f(\bm{\theta}^{t}) \rVert_{\bm{\Pi}_{c,i_c^{t,0}}}^2
			+
			\sum_{p=1}^{P-1}
			\lVert \nabla f(\bm{\theta}^{t,p}(c)) \rVert_{\bm{\Pi}_{c,i_c^{t,p}}}^2
			\Big),
		\end{split}
		\]
		where $(a)$ follows from \eqref{eq:unbiased} and \eqref{eq:bounded_var}. We thus arrive at~\eqref{eq:norm_bound}.}
	
	\section{Proof of Theorem~\ref{thm:mtcd_tpc_convergence}} \label{sec:thm1_proof}
	
	It follows from the $L$-smoothness assumption~\eqref{eq:L-smoothness} that
	\[
	{\textstyle
		f(\bm{\theta}^{t+1})
		\leq
		f(\bm{\theta}^{t})
		+
		\nabla f(\bm{\theta}^{t})^\top(\bm{\theta}^{t+1}-\bm{\theta}^{t})
		+
		\frac{L}{2}
		||\bm{\theta}^{t+1}-\bm{\theta}^{t}||^2.
	}
	\]
	We now let $\delta^t\coloneqq f(\bm{\theta}^{t+1})-f(\bm{\theta}^{t})$, split the upper bound across cluster-specific terms, and take the conditional expectation:
	\[
	\mathbb{E}_{t^+} \delta^t
	\leq
	\sum_{c=1}^{C}
	\mathbb{E}_{t^+}
	\left[
	\nabla_c f(\bm{\theta}^{t})^\top(\bm{\theta}_c^{t+1}-\bm{\theta}_c^{t})
	+
	\frac{L}{2}
	||\bm{\theta}_c^{t+1}-\bm{\theta}_c^{t}||^2
	\right]
	.
	\]
	Using Lemma~\ref{lemma:inner_product_bound} and Lemma~\ref{lemma:norm_bound} on the bound above, we get:
	\[
	\begin{split}
		\mathbb{E}_{t^+}\delta^t &\leq
		-\eta\left(1-\frac{\eta^2L^2(P-1)^2}{2}\right)\cdot\sum_{c=1}^{C}
		\lVert\nabla f(\bm{\theta}^{t})\rVert_{\bm{\Pi}_{c,i_c^{t,0}}}^{2}
		\\
		&\quad
		-
		\frac{\eta}{2}\cdot\sum_{c=1}^{C} \lVert \nabla f(\bm{\theta}^{t}) \rVert_{\sum_{p=1}^{P-1}\bm{\Pi}_{c,i_c^{t,p}}}^{2}
		\\
		&\quad
		-
		\frac{\eta}{2}
		(1-\eta^2L^2(P-1)^2)\cdot\sum_{c=1}^{C}
		\sum_{p=1}^{P-1} \lVert \nabla f(\bm{\theta}^{t,p}(c)) \rVert_{\bm{\Pi}_{c,i_c^{t,p}}}^{2}
		\\
		&\quad+\frac{\eta^2P^2\sigma^2L}{2B}
		+
		\frac{\eta^2LP}{2}\cdot\sum_{c=1}^{C}
		\lVert \nabla f(\bm{\theta}^{t}) \rVert_{\bm{\Pi}_{c,i_c^{t,0}}}^2
		\\
		&\quad
		+
		\frac{\eta^2LP}{2}\cdot\sum_{c=1}^{C}
		\sum_{p=1}^{P-1}
		\lVert \nabla f(\bm{\theta}^{t,p}(c)) \rVert_{\bm{\Pi}_{c,i_c^{t,p}}}^2
		.
	\end{split}
	\]
	Or, equivalently,
	\begin{equation} \label{eq:descent_lemma}
		\begin{split}
			&\mathbb{E}_{t^+}\delta^t
			\\
			&\;\leq
			-\eta\left(1-\frac{\eta PL}{2}-\frac{\eta^2L^2(P-1)^2}{2}\right)\sum_{c=1}^{C}
			\lVert\nabla f(\bm{\theta}^{t})\rVert_{\bm{\Pi}_{c,i_c^{t,0}}}^{2}
			\\
			&\;-
			\frac{\eta}{2}\sum_{c=1}^{C} \lVert \nabla f(\bm{\theta}^{t}) \rVert_{\sum_{p=1}^{P-1}\bm{\Pi}_{c,i_c^{t,p}}}^{2}
			+\frac{\eta^2P^2\sigma^2L}{2B}
			\\
			&\;-
			\frac{\eta}{2}
			(1-\eta LP-\eta^2L^2(P-1)^2)\sum_{c=1}^{C}
			\sum_{p=1}^{P-1} \lVert \nabla f(\bm{\theta}^{t,p}(c)) \rVert_{\bm{\Pi}_{c,i_c^{t,p}}}^{2}.
		\end{split}
	\end{equation}
	Now, as long as $1-\eta LP-\eta^2L^2(P-1)^2>0$, which we can ensure by having $\eta \in \big(0,\frac{1}{2LP}\big]$, we can drop the $ \sum_{c=1}^{C}
	\sum_{p=1}^{P-1} \lVert \nabla f(\bm{\theta}^{t,p}(c)) \rVert_{\bm{\Pi}_{c,i_c^{t,p}}}^{2} $ term, arriving at:
	\[
	\begin{split}
		&\mathbb{E}_{t^+} \delta^t
		\leq
		{\textstyle
			-\mu(\eta)\cdot\sum_{c=1}^{C}
			\lVert\nabla f(\bm{\theta}^{t})\rVert_{\bm{\Pi}_{c,i_c^{t,0}}}^{2}
		}
		\\
		&\qquad\qquad-
		\nu(\eta)\cdot\sum_{c=1}^{C} \lVert \nabla f(\bm{\theta}^{t}) \rVert_{\sum_{p=1}^{P-1}\bm{\Pi}_{c,i_c^{t,p}}}^{2}
		+\frac{\eta^2P^2\sigma^2L}{2B}.
	\end{split}
	\]
	where $\mu(\eta)\coloneqq\eta\left(1-\frac{\eta PL}{2}-\frac{\eta^2L^2(P-1)^2}{2}\right)$ and $\nu(\eta)\coloneqq\frac{\eta}{2}$. Further, let $\tilde{\tau}(\eta)\coloneqq\min\{\mu(\eta),\nu(\eta)\}$, we get that
	\[
	\mathbb{E}_{t^+} \delta^t
	\leq
	-
	\tilde{\tau}(\eta)
	\sum_{c=1}^{C}
	\lVert\nabla f(\bm{\theta}^{t})\rVert_{\bm{\Pi}_{c,i_c^{t,0}}+\dots+\bm{\Pi}_{c,i_c^{t,P-1}}}^{2}
	+\frac{\eta^2P^2\sigma^2L}{2B}.
	\]
	We now take the conditional expectation with respect to $ \bm{p}^t $,
	\[
	\mathbb{E}_{t} \delta^t
	\leq -\tilde{\tau}(\eta)
	\sum_{c=1}^{C}
	\mathbb{E}_{t} \lVert\nabla f(\bm{\theta}^{t})\rVert_{\bm{\Pi}_{c,i_c^{t,0}}+\dots+\bm{\Pi}_{c,i_c^{t,P-1}}}^{2}
	+\frac{\eta^2P^2\sigma^2L}{2B},
	\]
	and, since $ \bm{\theta}^t $ is fixed when conditioned on $ \mathcal{F}^t $, we get that:
	\[
	\mathbb{E}_{t} \delta^t
	\leq
	-\tilde{\tau}(\eta)
	\cdot\sum_{c=1}^{C}
	\lVert \nabla f(\bm{\theta}^{t}) \rVert_{\bar{\bm{\Pi}}_{c}^{t}}^{2}
	+\frac{\eta^2P^2\sigma^2L}{2B}
	,
	\]
	where $\bar{\bm{\Pi}}_{c}^{t}\coloneqq\mathbb{E}\left[\bm{\Pi}_{c,i_c^{t,0}}+\dots+\bm{\Pi}_{c,i_c^{t,P-1}}
	\mid\mathcal{F}^t\right]$. From~\eqref{eq:min_prob_token_visit}, we have that $\bar{\bm{\Pi}}_{c}^{t}\succeq\pi\bm{\Pi}_c$.
	From this, we have that:
	\begin{equation} \label{eq:cond_exp_descent_lemma}
		\begin{split}
			&\mathbb{E}_{t} \delta^t
			\leq
			-\tilde{\tau}(\eta)\cdot \pi
			\cdot\sum_{c=1}^{C}
			\lVert \nabla f(\bm{\theta}^{t}) \rVert_{\bm{\Pi}_c}^{2}
			+\frac{\eta^2P^2\sigma^2L}{2B}
			\\
			&\quad=
			-\tilde{\tau}(\eta)\cdot \pi
			\cdot
			\lVert \nabla f(\bm{\theta}^{t}) \rVert_{}^{2}
			+\frac{\eta^2P^2\sigma^2L}{2B}
			.
		\end{split}
	\end{equation}
	Thus, since $\tilde{\tau}(\eta)>0$ for $\eta \in \big(0,\frac{1}{2LP}\big]$, it follows from the classic result by Robbins and Siegmund~\citep{robbins1971convergence} that, in the full-batch case, $\nabla f \left( \bm{\theta}^{t}\right)\to0$ almost surely. Taking the unconditional expectation and letting $ {\tau}(\eta) \coloneqq {\tilde{\tau}(\eta)}/{\eta} $, we get:
	\[
	\mathbb{E}
	\delta^t
	\leq
	-\eta\pi\cdot\tau(\eta)
	\cdot \mathbb{E} \lVert \nabla f(\bm{\theta}^{t}) \rVert_{}^{2}
	+\frac{\eta^2P^2\sigma^2L}{2B}
	.
	\]
	Rearranging the terms and averaging over $t$, we have that:
	\[
	\frac{1}{T}
	\sum_{t=0}^{T-1}
	\mathbb{E} \lVert \nabla f(\bm{\theta}^{t}) \rVert_{}^{2}
	\leq
	\frac{f(\bm{\theta}^{0})
		-
		\mathbb{E}
		f(\bm{\theta}^{T})}{\eta\pi T \cdot \tau(\eta) }
	+\frac{\eta^2P^2\sigma^2L}{2\eta\pi B \cdot  \tau(\eta)}
	.
	\]
	Thus, from the finite infimum assumption in~\eqref{eq:L-smoothness}, we get:
	\[
	{
		\frac{1}{T}
		\sum_{t=0}^{T-1}
		\mathbb{E} \lVert \nabla f(\bm{\theta}^{t}) \rVert_{}^{2}
		\leq
		\frac{f(\bm{\theta}^{0})
			-
			f^\star}{\eta\pi T \cdot \tau(\eta) }
		+\frac{\eta^2P^2\sigma^2L}{2\eta\pi B \cdot  \tau(\eta)}
		.
	}
	\]
	Finally, since $\tau(\eta)\geq1/2$ for $\eta \in \big(0,\frac{1}{2LP}\big]$, we arrive at~\eqref{eq:main_theorem}.
	
	\section{Proof of Theorem~\ref{thm:mtcd_convergence}}
	\label{sec:thm2_proof}
	
	We now consider the case where tokens roam over overlapping sets of clients. We define $\bm{\Pi}_\gamma$ analogously to $\bm{\Pi}_c$. That is, the nonzero entries along the diagonal correspond to the blocks that can be visited by token $\gamma$. This means that $\bm{\Pi}_\gamma=\bm{I}$, yet we stick to this notation for consistency, and because the $\bm{\Pi}_{\gamma,i_\gamma^{t,p}}$ notation will remain useful.
	In this setting, \eqref{eq:min_prob_token_visit} holds for a single ``cluster'': the whole graph.
	Lemma~\ref{lemma:inner_product_bound} and Lemma~\ref{lemma:norm_bound} also continue to hold, sufficing to adjust their notation:
	\[
	\begin{split}
		\mathbb{E}_{t^+}&
		\left[\nabla_\gamma f(\bm{\theta}^{t})^\top(\bm{\theta}_\gamma^{t+1}-\bm{\theta}_\gamma^{t}) \right]
		\\
		&\leq
		-\eta\left(1-\frac{\eta^2L^2(P-1)^2}{2}\right)
		\lVert\nabla f(\bm{\theta}^{t})\rVert_{\bm{\Pi}_{\gamma,i_\gamma^{t,0}}}^{2}
		\\
		&\quad-
		\frac{\eta}{2} \lVert \nabla f(\bm{\theta}^{t}) \rVert_{\sum_{p=1}^{P-1}\bm{\Pi}_{\gamma,i_c^{t,p}}}^{2}\\
		&\quad
		-
		\frac{\eta}{2}
		(1-\eta^2L^2(P-1)^2)
		\sum_{p=1}^{P-1} \lVert \nabla f(\bm{\theta}^{t,p}(\gamma)) \rVert_{\bm{\Pi}_{\gamma,i_\gamma^{t,p}}}^{2}
	\end{split}
	\]
	and
	\[
	\begin{split}
		&\mathbb{E}_{t^+}
		||\bm{\theta}_\gamma^{t+1}-\bm{\theta}_\gamma^{t}||^2
		\leq
		\\
		&\;\;
		\eta^2P
		\left(
		\frac{\sigma^2P}{B}
		+
		\lVert \nabla f(\bm{\theta}^{t}) \rVert_{\bm{\Pi}_{\gamma,i_\gamma^{t,0}}}^2
		+
		\sum_{p=1}^{P-1}
		\lVert \nabla f(\bm{\theta}^{t,p}(\gamma)) \rVert_{\bm{\Pi}_{\gamma,i_\gamma^{t,p}}}^2
		\right)
		,
	\end{split}
	\]
	respectively. From~\eqref{eq:convexity}, we have that
	\[
	{\textstyle
		f(\bm{\theta}^{t+1})
		=
		f\left(\frac{1}{\Gamma}\sum_{\gamma=1}^{\Gamma}\bm{\theta}^{t,P}(\gamma)\right)
		\leq
		\frac{1}{\Gamma}\sum_{\gamma=1}^{\Gamma}
		f(\bm{\theta}^{t,P}(\gamma)).
	}
	\]
	Thus, it follows from $L$-smoothness~\eqref{eq:L-smoothness} that
	\[
	\delta^t
	\leq
	\frac{1}{\Gamma}\sum_{\gamma=1}^{\Gamma}
	\left[
	\nabla f(\bm{\theta}^{t})^\top\left(\bm{\theta}^{t,P}(\gamma)-\bm{\theta}^{t}\right)
	+
	\frac{L}{2}
	\left\lVert\bm{\theta}^{t,P}(\gamma)-\bm{\theta}^{t}\right\rVert^2
	\right]
	.
	\]
	Taking the conditional expectation, we get that:
	{
		\[
		\mathbb{E}_{t^+}\delta^t
		\leq
		\frac{1}{\Gamma}
		\sum_{\gamma=1}^{\Gamma}
		\mathbb{E}_{t^+}
		\left[
		\nabla_\gamma f(\bm{\theta}^{t})^\top(\bm{\theta}_\gamma^{t+1}-\bm{\theta}_\gamma^{t})
		+
		\frac{L}{2}||\bm{\theta}_\gamma^{t+1}-\bm{\theta}_\gamma^{t}||^2
		\right]
		.
		\]
		We bound the inner product term using Lemma~\ref{lemma:inner_product_bound} and the norm term using Lemma~\ref{lemma:norm_bound}, arriving at:
		\[
		\begin{split}
			\mathbb{E}_{t^+} \delta^t
			&\leq
			{\textstyle
				-
				\frac{\eta}{\Gamma}\left(1-\frac{\eta^2L^2(P-1)^2}{2}\right)\cdot\sum_{\gamma=1}^{\Gamma}
				\lVert\nabla f(\bm{\theta}^{t})\rVert_{\bm{\Pi}_{\gamma,i_\gamma^{t,0}}}^{2}
			}
			\\
			&\quad-
			{\textstyle
				\frac{\eta}{2\Gamma}\cdot\sum_{\gamma=1}^{\Gamma} \lVert \nabla f(\bm{\theta}^{t}) \rVert_{\sum_{p=1}^{P-1}\bm{\Pi}_{\gamma,i_\gamma^{t,p}}}^{2}
			}
			\\
			&\quad-
			\frac{\eta}{2\Gamma}
			(1-\eta^2L^2(P-1)^2)\cdot\sum_{\gamma=1}^{\Gamma}
			\sum_{p=1}^{P-1} \lVert \nabla f(\bm{\theta}^{t,p}(\gamma)) \rVert_{\bm{\Pi}_{\gamma,i_\gamma^{t,p}}}^{2}
			\\
			&\quad
			+
			\frac{\eta^2P^2\sigma^2L}{2B}
			+
			\frac{\eta^2LP}{2\Gamma}\cdot\sum_{\gamma=1}^{\Gamma}
			\lVert \nabla f(\bm{\theta}^{t}) \rVert_{\bm{\Pi}_{\gamma,i_\gamma^{t,0}}}^2
			\\
			&\quad+
			\frac{\eta^2LP}{2\Gamma}\cdot\sum_{\gamma=1}^{\Gamma}
			\sum_{p=1}^{P-1}
			\lVert \nabla f(\bm{\theta}^{t,p}(\gamma)) \rVert_{\bm{\Pi}_{\gamma,i_\gamma^{t,p}}}^2
			.
		\end{split}
		\]
		Thus, we have that
		\[
		\begin{split}
			\mathbb{E}_{t^+} \delta^t
			&\leq -\frac{\eta}{\Gamma}\left(1-\frac{\eta PL}{2}-\frac{\eta^2L^2(P-1)^2}{2}\right)\sum_{\gamma=1}^{\Gamma}
			\lVert\nabla f(\bm{\theta}^{t})\rVert_{\bm{\Pi}_{\gamma,i_\gamma^{t,0}}}^{2}
			\\
			&\quad -
			\frac{\eta}{2\Gamma}\sum_{\gamma=1}^{\Gamma} \lVert \nabla f(\bm{\theta}^{t}) \rVert_{\sum_{p=1}^{P-1}\bm{\Pi}_{\gamma,i_\gamma^{t,p}}}^{2}
			+\frac{\eta^2P^2\sigma^2L}{2B}
			\\
			&\quad-
			\frac{\eta}{2\Gamma}
			(1-\eta LP-\eta^2L^2(P-1)^2)
			\sum_{\gamma=1}^{\Gamma}
			\sum_{p=1}^{P-1} \lVert \nabla f(\bm{\theta}^{t,p}(\gamma)) \rVert_{\bm{\Pi}_{\gamma,i_\gamma^{t,p}}}^{2}.
		\end{split}
		\]
		Now, if $\eta \in \big(0,\frac{1}{2LP}\big]$, we can drop the last term, and get that:
		\[
		\begin{split}
			\mathbb{E}_{t^+} \delta^t
			& \leq
			-\frac{\mu(\eta)}{\Gamma}\cdot\sum_{\gamma=1}^{\Gamma}
			\lVert\nabla f(\bm{\theta}^{t})\rVert_{\bm{\Pi}_{\gamma,i_\gamma^{t,0}}}^{2}
			-
			\frac{\nu(\eta)}{\Gamma}\cdot\sum_{\gamma=1}^{\Gamma} \lVert \nabla f(\bm{\theta}^{t}) \rVert_{\sum_{p=1}^{P-1}\bm{\Pi}_{\gamma,i_\gamma^{t,p}}}^{2}
			+\frac{\eta^2P^2\sigma^2L}{2B}.
		\end{split}
		\]
		Thus, using the fact that $\tilde{\tau}(\eta)=\min\{\mu(\eta),\nu(\eta)\}$, we have:
		\[
		\mathbb{E}_{t^+} \delta^t
		\leq
		-\frac{\tilde{\tau}(\eta)}{\Gamma} \sum_{\gamma=1}^{\Gamma}
		\lVert\nabla f(\bm{\theta}^{t})\rVert_{\sum_{p=0}^{P-1}\bm{\Pi}_{\gamma,i_\gamma^{t,p}}}^{2}
		+
		\frac{\eta^2P^2\sigma^2L}{2B}.
		\]
		We now take the conditional expectation $\mathbb{E}_{t}$,
		\[
		\begin{split}
			\mathbb{E}_{t} \delta^t
			\leq
			-\frac{\tilde{\tau}(\eta)}{\Gamma} \sum_{\gamma=1}^{\Gamma}
			\mathbb{E}_{t}
			\lVert\nabla f(\bm{\theta}^{t})\rVert_{\sum_{p=0}^{P-1}\bm{\Pi}_{\gamma,i_\gamma^{t,p}}}^{2}
			+
			\frac{\eta^2P^2\sigma^2L}{2B},
		\end{split}
		\]
		and, since $ \bm{\theta}^t $ is fixed when conditioned on $ \mathcal{F}^t $, we get that:
		\[
		\mathbb{E}_{t} \delta^t
		\leq
		-\frac{\tilde{\tau}(\eta)}{\Gamma}\cdot\sum_{\gamma=1}^{\Gamma}
		\lVert \nabla f(\bm{\theta}^{t}) \rVert_{\bar{\bm{\Pi}}_{\gamma}^{t}}^{2}
		+\frac{\eta^2P^2\sigma^2L}{2B}
		,
		\]
		where $\bar{\bm{\Pi}}_{\gamma}^{t}\coloneqq\mathbb{E}_t\left[\bm{\Pi}_{\gamma,i_\gamma^{t,0}}+\dots+\bm{\Pi}_{\gamma,i_\gamma^{t,P-1}} \right]$. From~\eqref{eq:min_prob_token_visit}, we have that $\bar{\bm{\Pi}}_{\gamma}^{t}\succeq\pi \bm{I}$. This can be achieved through various communication schemes, as discussed below, in Appendix~\ref{app:lb-pi}. Therefore
		\[
		\begin{split}
			\mathbb{E}_{t} \delta^t
			&\leq
			-\frac{\tilde{\tau}(\eta)}{\Gamma}\cdot \pi\cdot\sum_{\gamma=1}^{\Gamma}
			\lVert \nabla f(\bm{\theta}^{t}) \rVert^{2}
			+\frac{\eta^2P^2\sigma^2L}{2B}
			\\
			&=
			-\tilde{\tau}(\eta)\cdot \pi
			\cdot
			\lVert \nabla f(\bm{\theta}^{t}) \rVert_{}^{2}
			+\frac{\eta^2P^2\sigma^2L}{2B}
			,
		\end{split}
		\]
		which matches inequality~\eqref{eq:cond_exp_descent_lemma} exactly. We can therefore continue the proof as in Appendix~\ref{sec:thm1_proof}, arriving the result in~\eqref{eq:overlapping-tokens-theorem}.}
	
	\paragraph{On the derivation of suboptimality results.}
	Note that, assuming only convexity, it is not straightforward to derive a suboptimality result, as our update vector is not an unbiased estimate of the gradient, and thus the standard procedure to derive such results from a
	descent lemma does not apply. Yet, it would be straightforward to obtain a suboptimality result (and prove linear convergence) if we further assumed the Polyak-Łojasiewicz inequality~\citep{polyak1963gradient} to hold.

	\begin{figure}[t]
		\centering
		\subfloat[$K=40$ path experiments, $Q=1$, $S=1$, varying $\Gamma$.]{\includegraphics[width=0.24\textwidth]{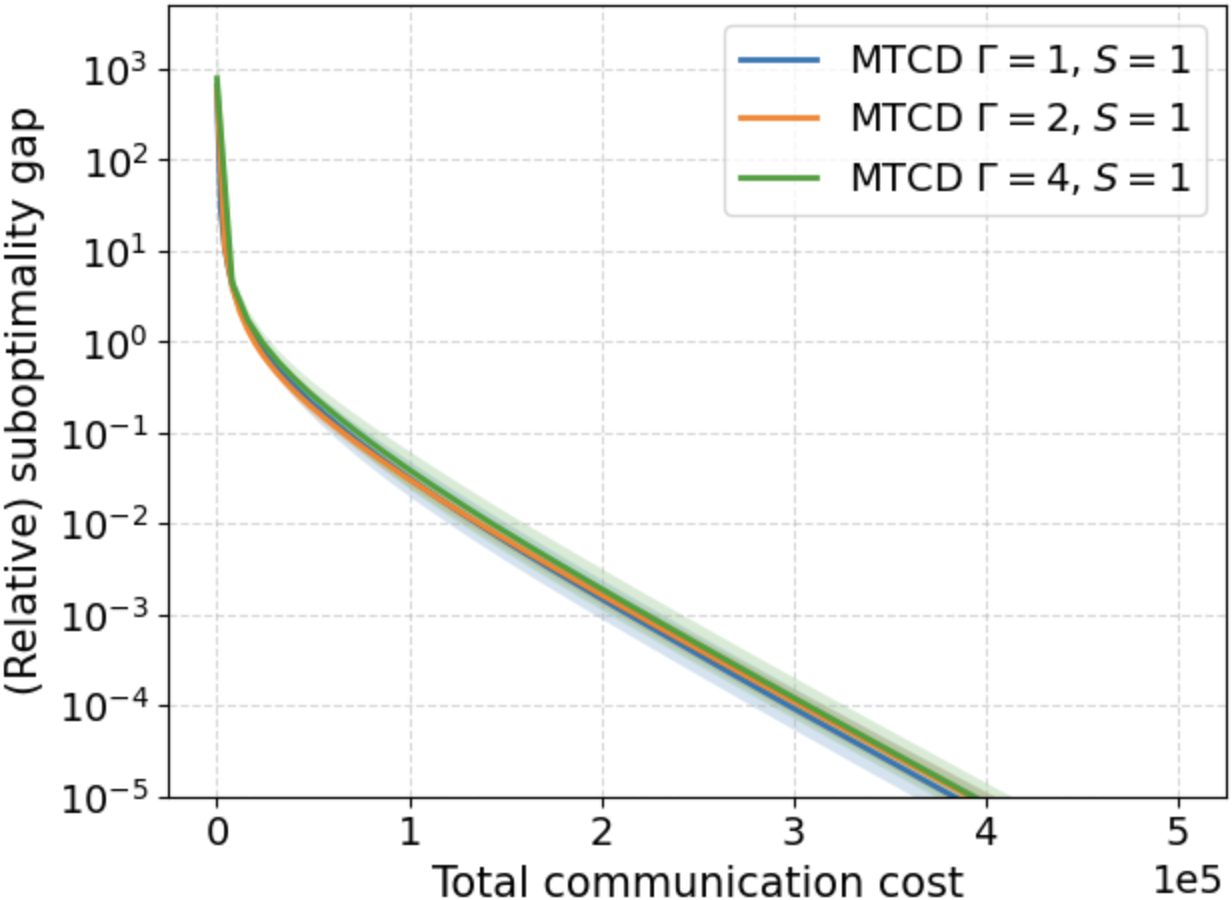}%
			\includegraphics[width=0.24\textwidth]{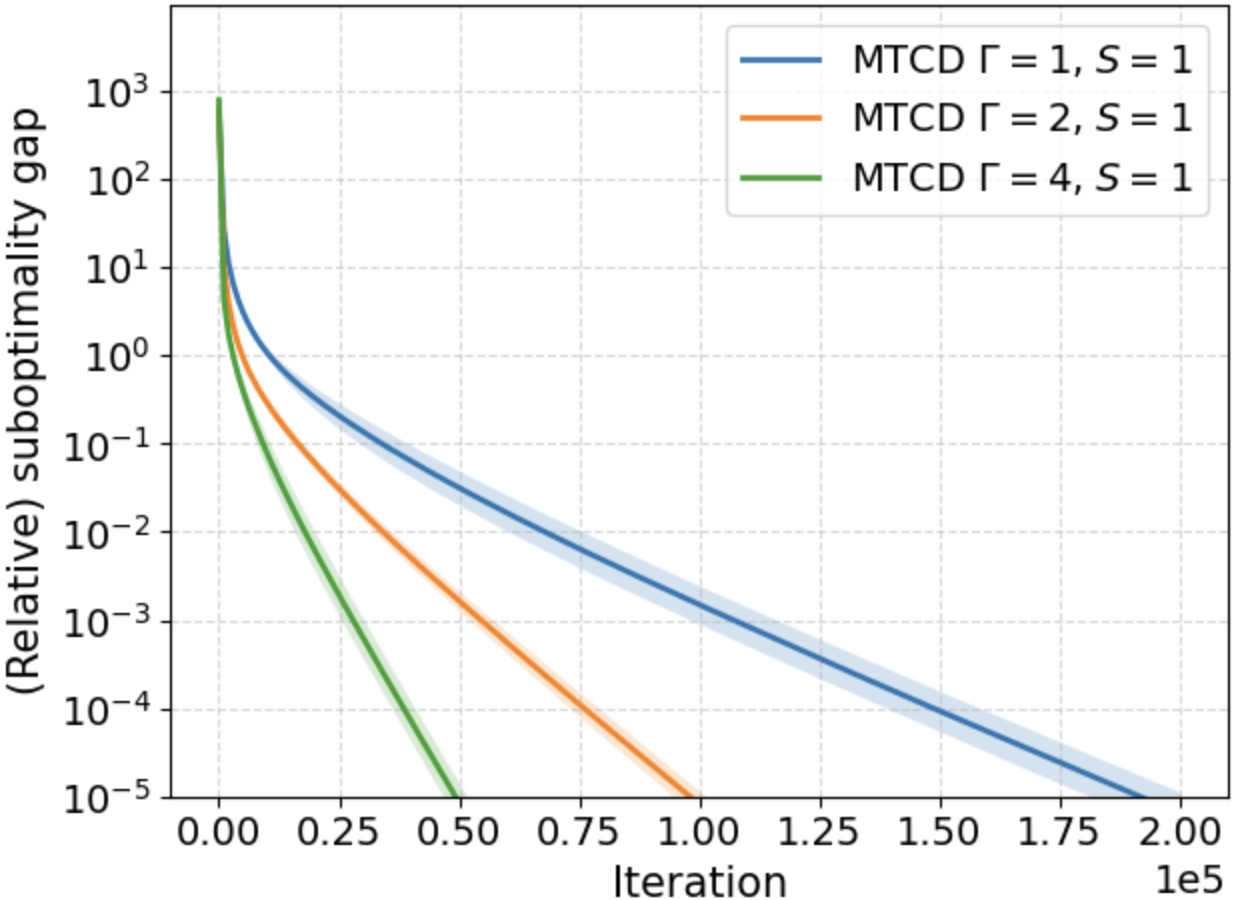}%
			\label{fig:mtcd_G_S1Q1}}
		
		\subfloat[$K=40$ path experiments, $Q=1$, $S=2$, varying $\Gamma$.]{\includegraphics[width=0.24\textwidth]{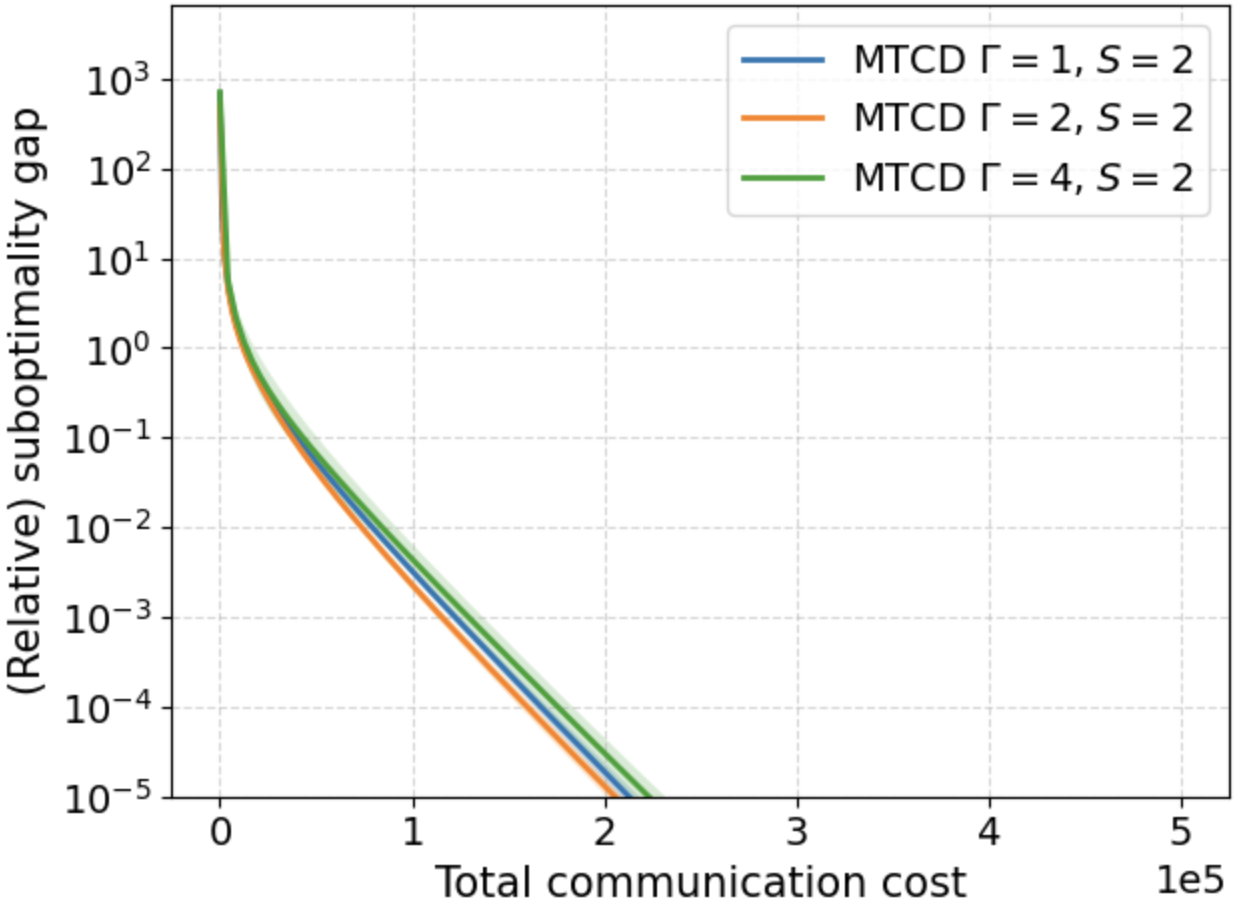}%
			\includegraphics[width=0.24\textwidth]{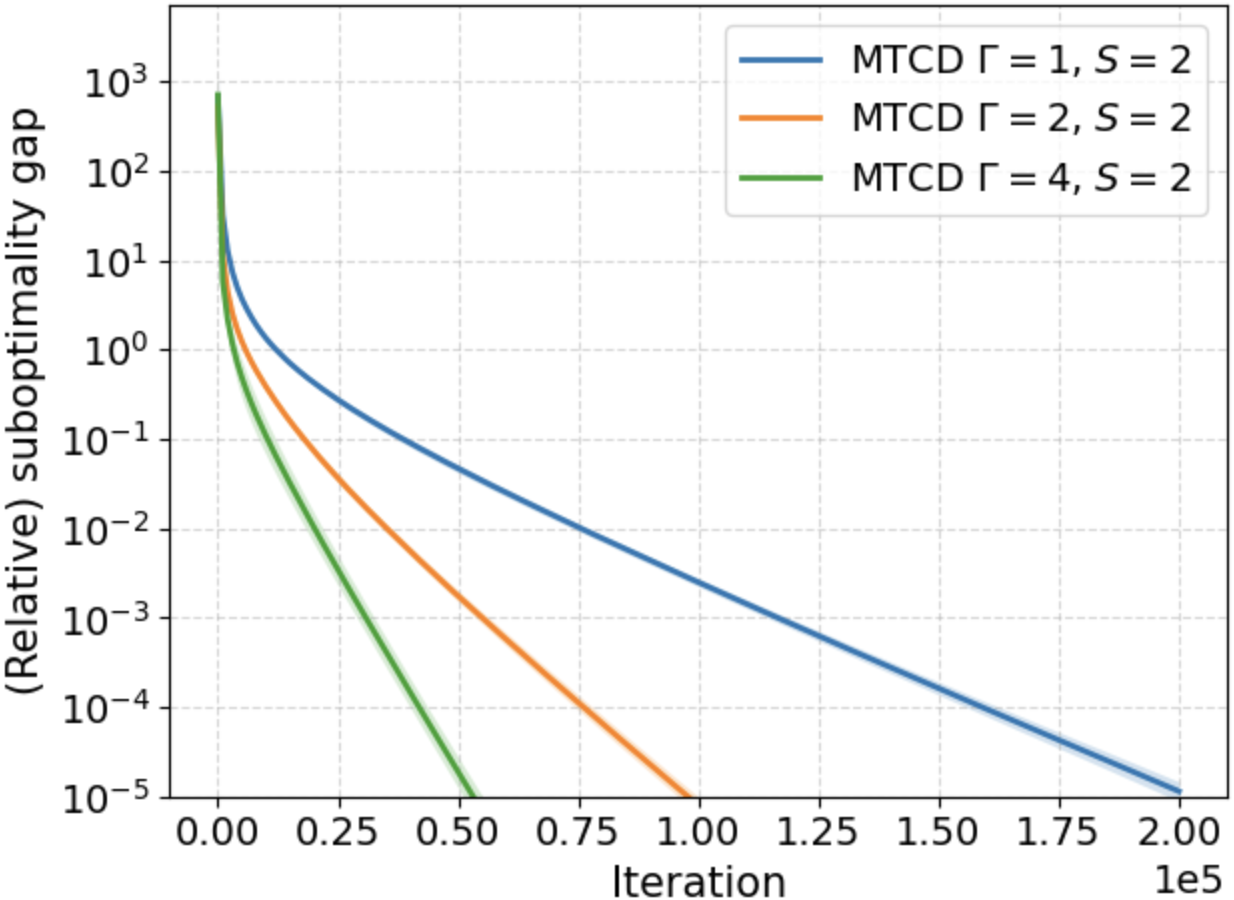}%
			\label{fig:mtcd_G_S2Q1}}
		
		\subfloat[$K=40$ path experiments, $Q=1$, $S=4$, varying $\Gamma$.]{\includegraphics[width=0.24\textwidth]{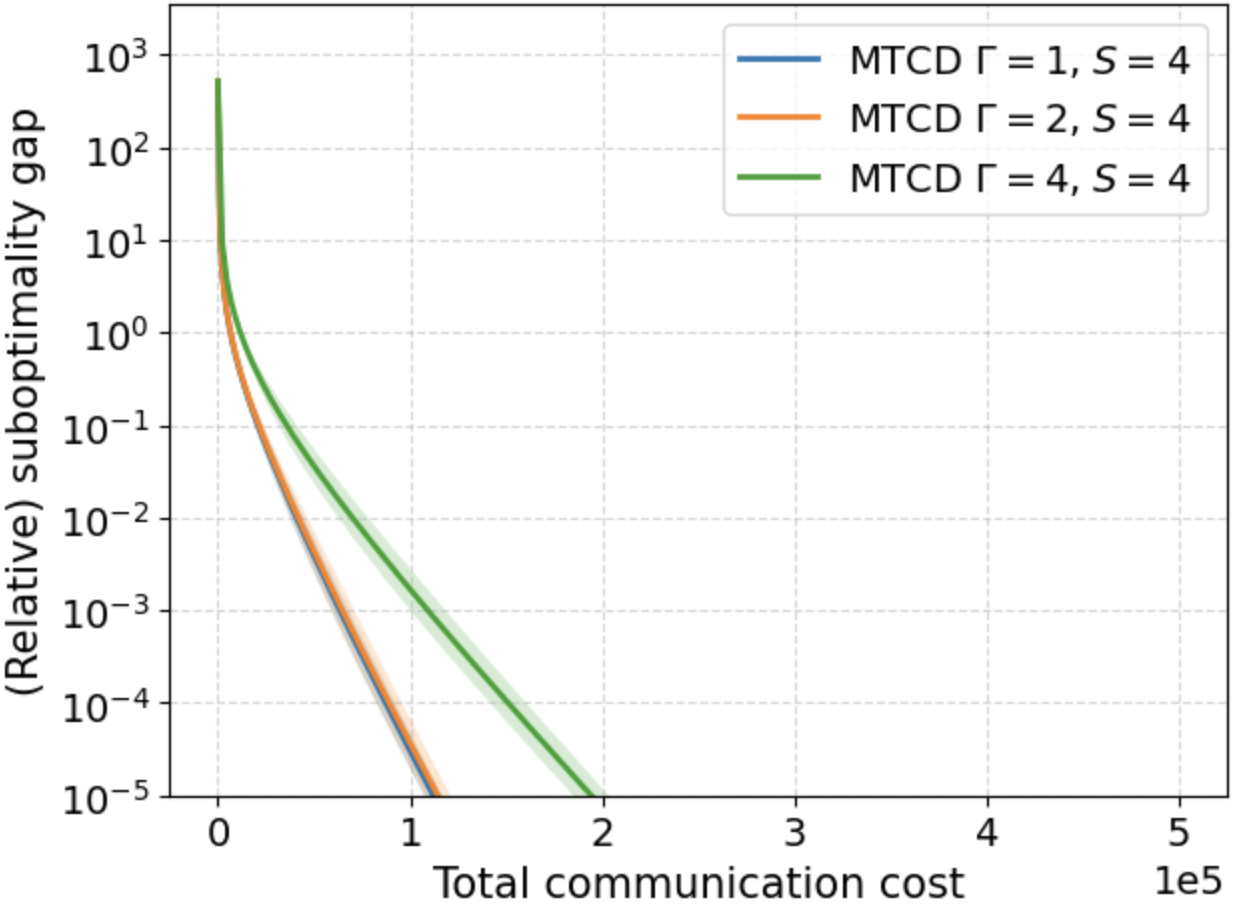}%
			\includegraphics[width=0.24\textwidth]{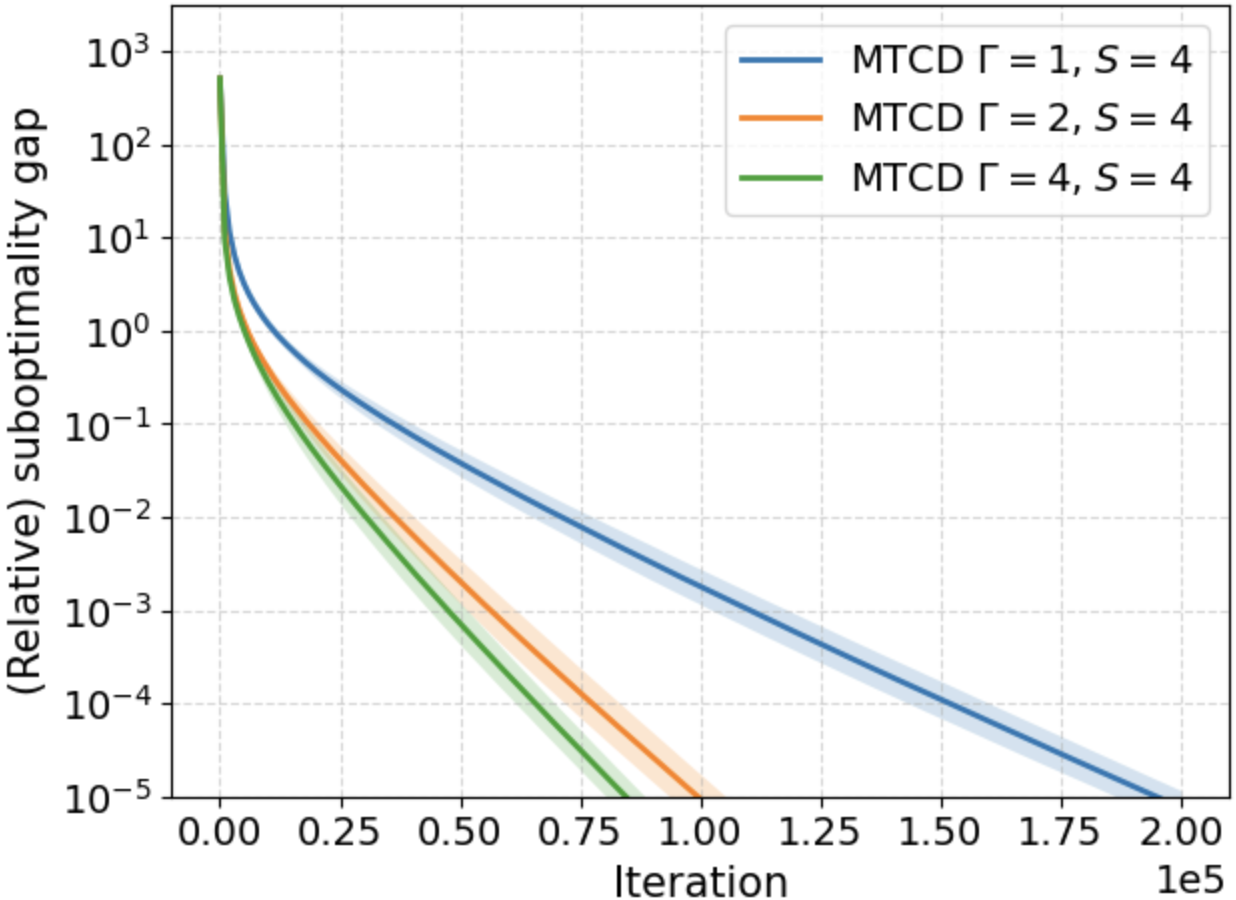}%
			\label{fig:mtcd_G_S4Q1}}
		
		\caption{$K=40$ path experiments with $Q=1$ and $S\in\{1,2,4\}$, varying $\Gamma$. Each row shows (left) communication cost and (right) iterations.}
		\label{fig:mtcd_G_Q1_allS}
	\end{figure}
	
	\begin{figure}[t]
		\centering
		
		\subfloat[$K=40$ path experiments, $Q=1$, $\Gamma=1$, varying $S$.]{\includegraphics[width=0.24\textwidth]{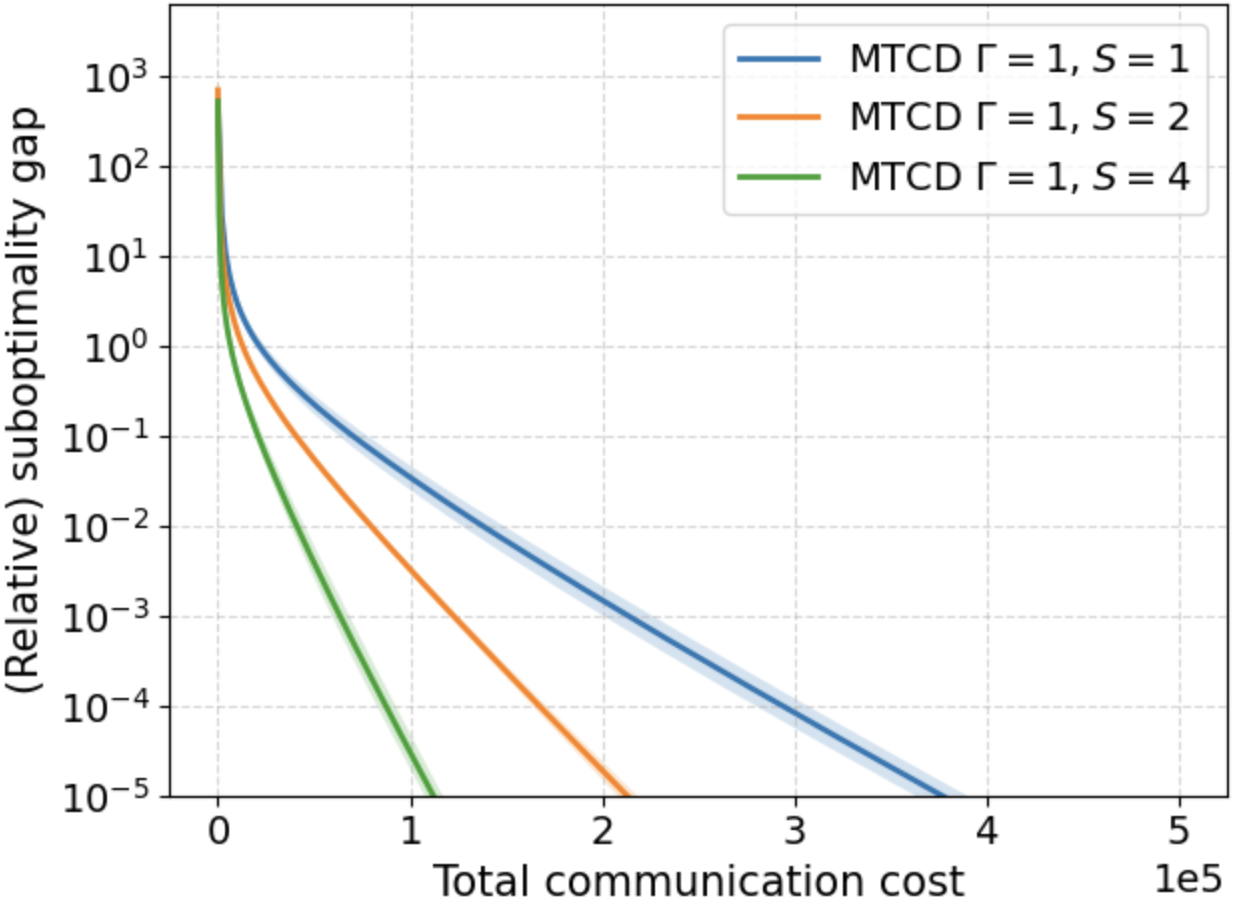}%
			\includegraphics[width=0.24\textwidth]{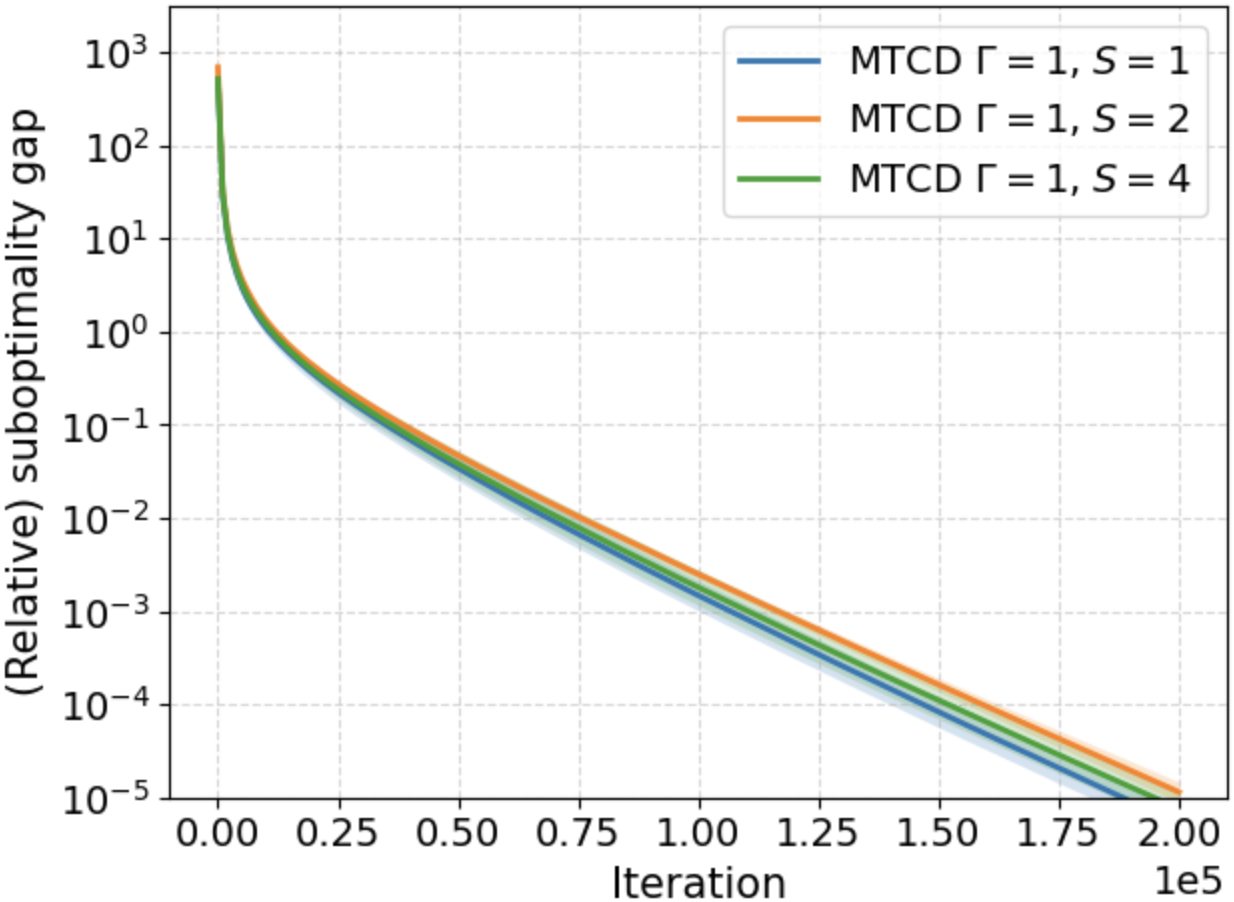}%
			\label{fig:mtcd_S_G1Q1}}
		
		\subfloat[$K=40$ path experiments, $Q=1$, $\Gamma=2$, varying $S$.]{\includegraphics[width=0.24\textwidth]{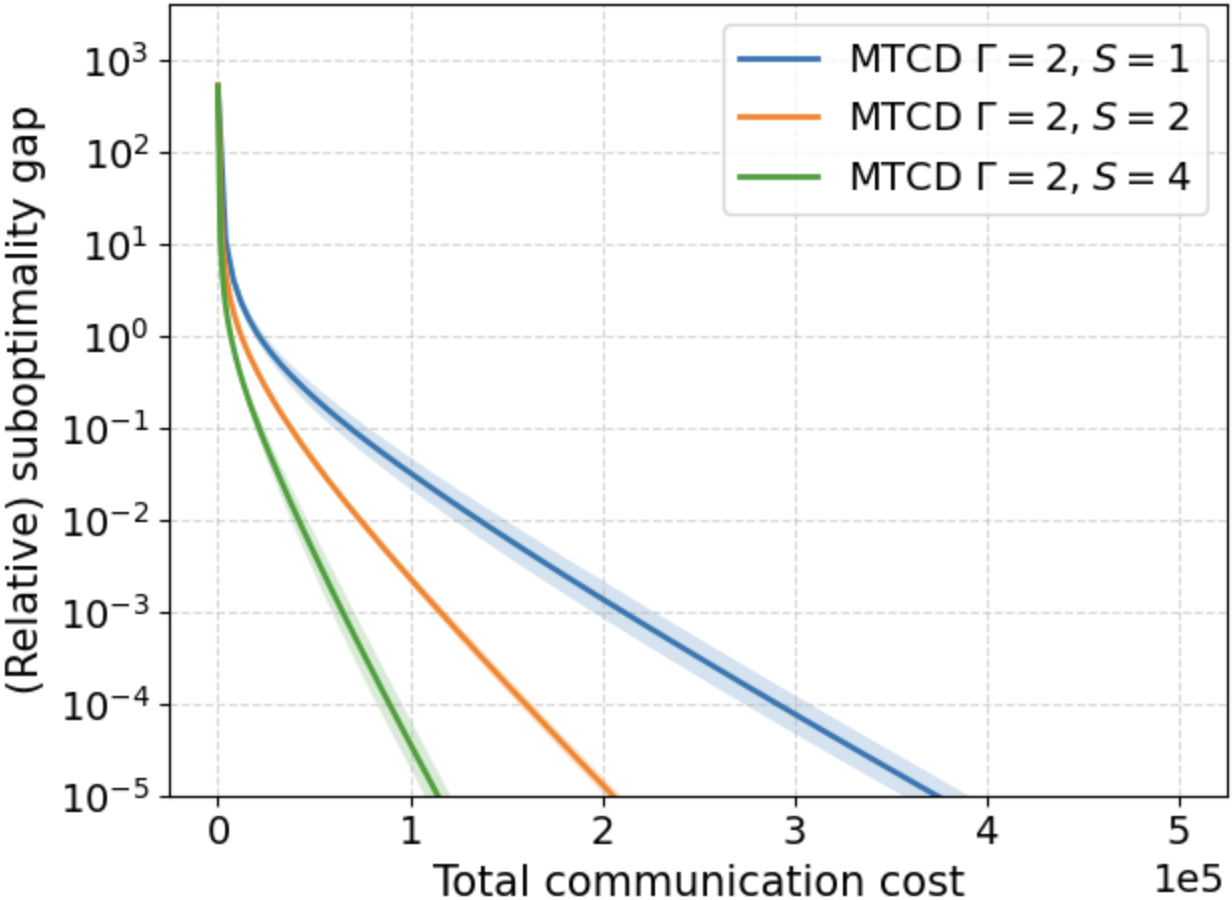}%
			\includegraphics[width=0.24\textwidth]{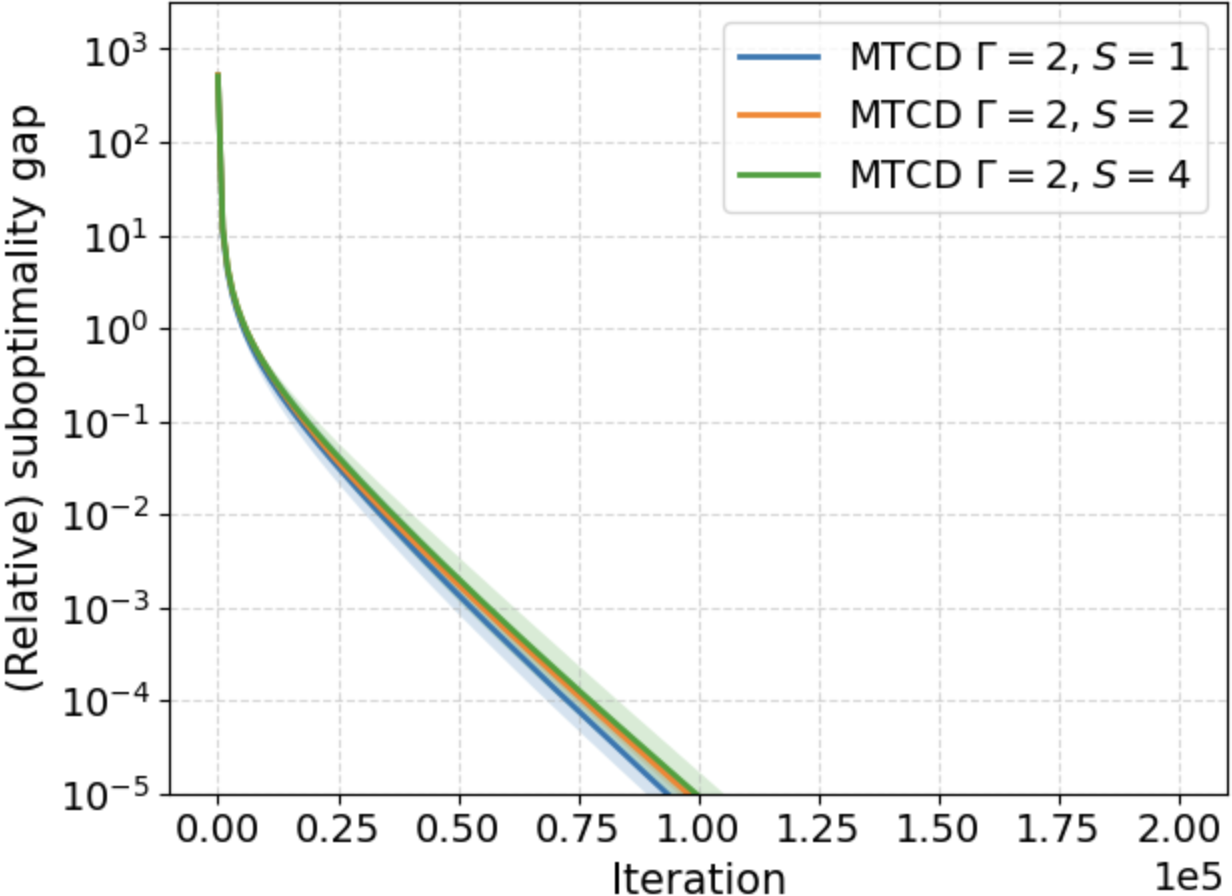}%
			\label{fig:mtcd_S_G2Q1}}
		
		\subfloat[$K=40$ path experiments, $Q=1$, $\Gamma=4$, varying $S$.]{\includegraphics[width=0.24\textwidth]{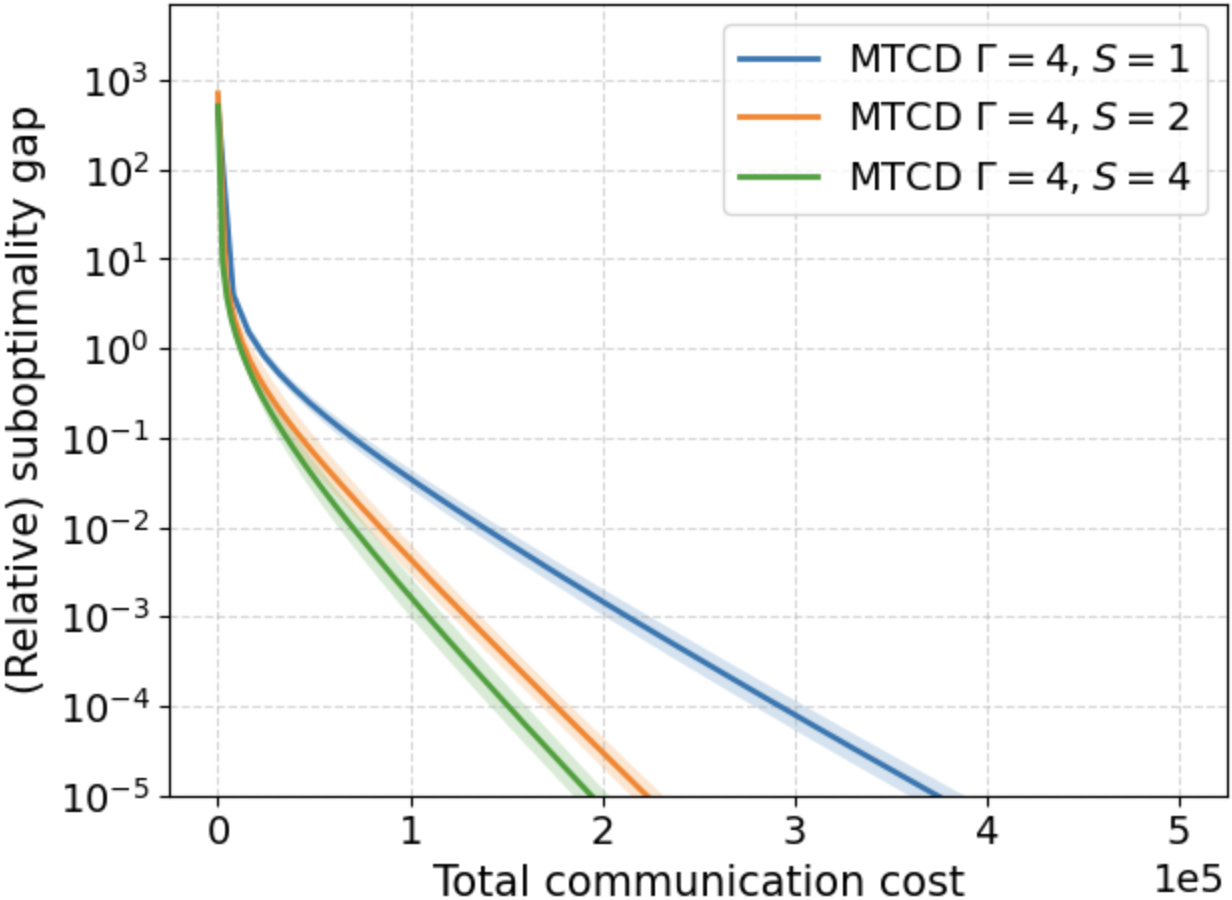}%
			\includegraphics[width=0.24\textwidth]{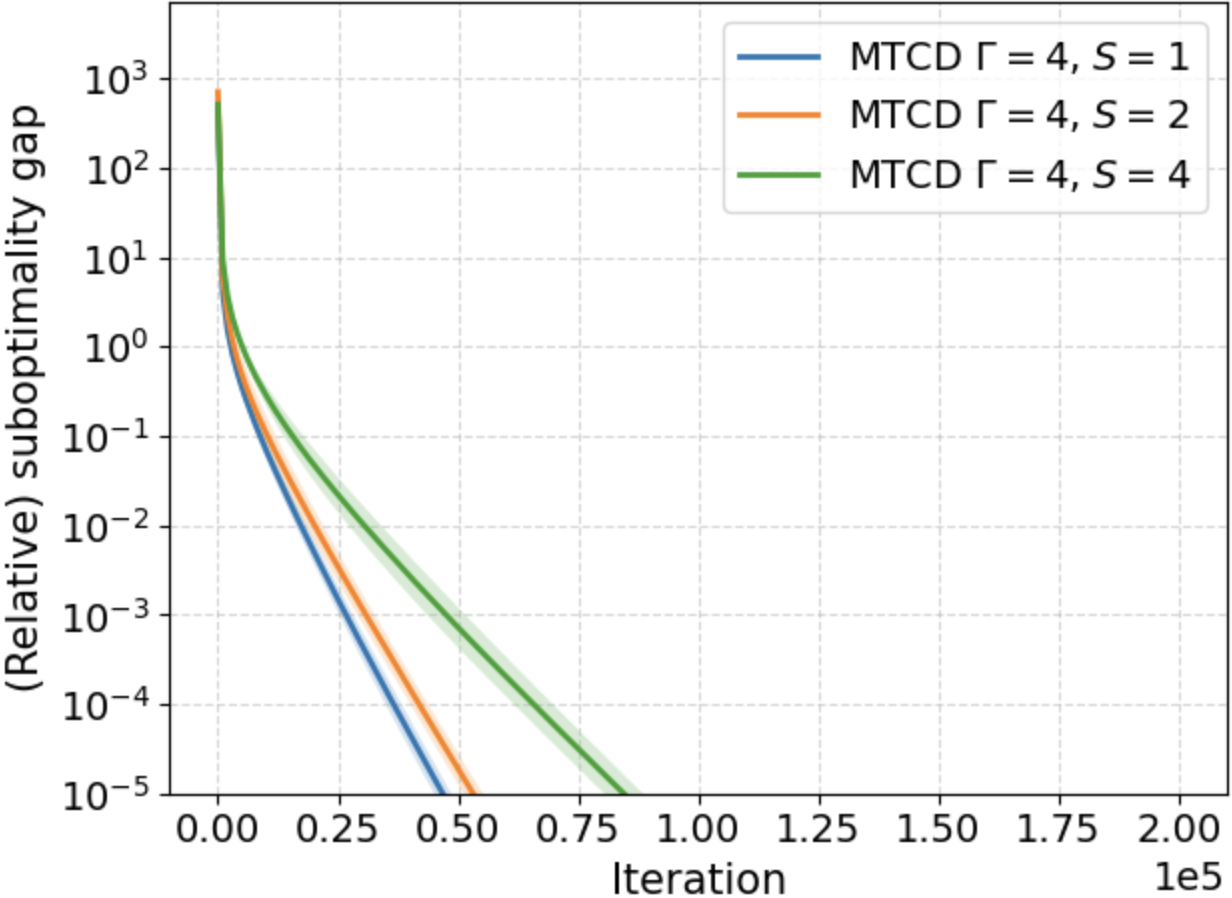}%
			\label{fig:mtcd_S_G4Q1}}
		
		\caption{$K=40$ path experiments with $Q=1$ and $\Gamma\in\{1,2,4\}$, varying $S$. Each row shows (left) communication cost and (right) iterations.}
		\label{fig:mtcd_S_Q1_allG}
	\end{figure}
	
	\begin{figure}[ht]
		\centering
		
		\subfloat[$K=40$ path experiments, $\Gamma=1$, $S=1$, varying $Q$.]{\includegraphics[width=0.24\textwidth]{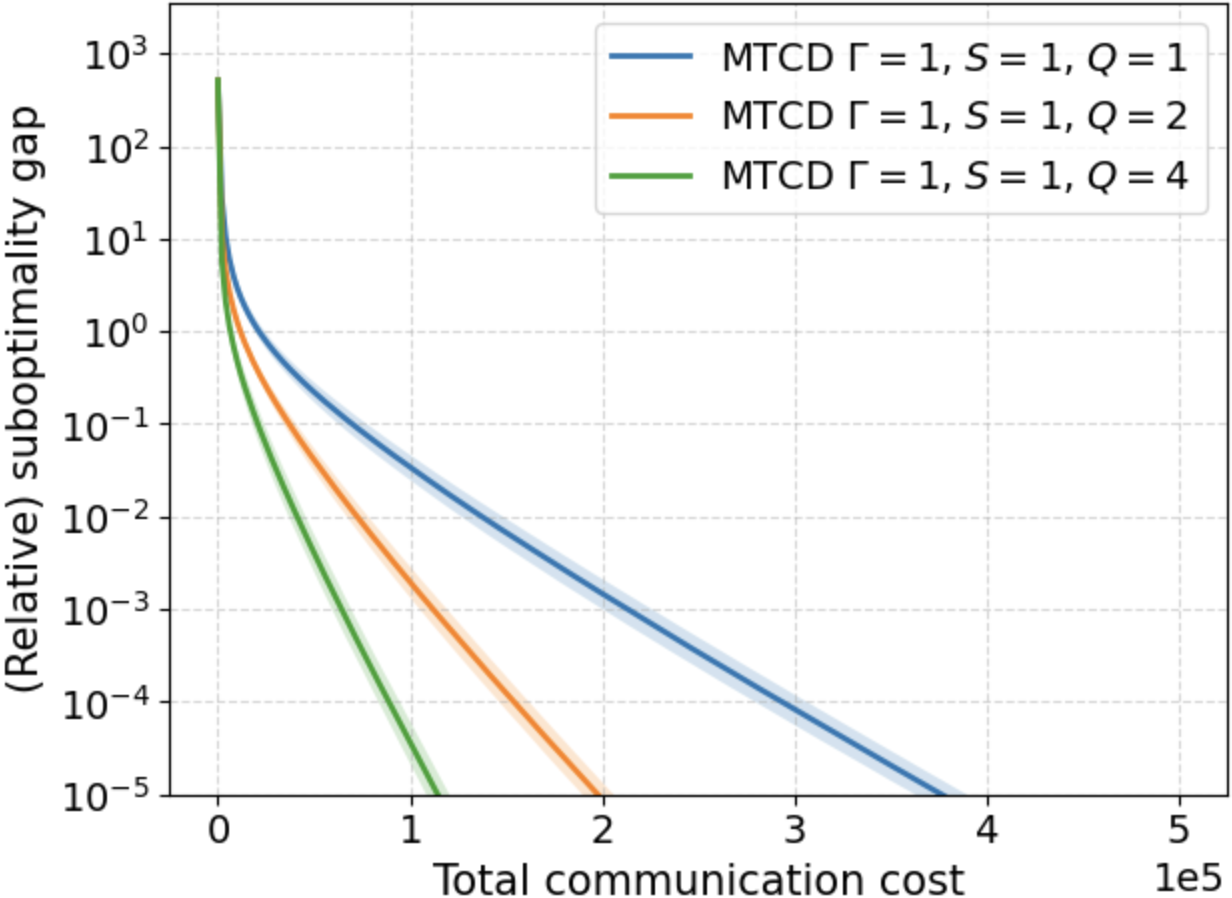}%
			\includegraphics[width=0.24\textwidth]{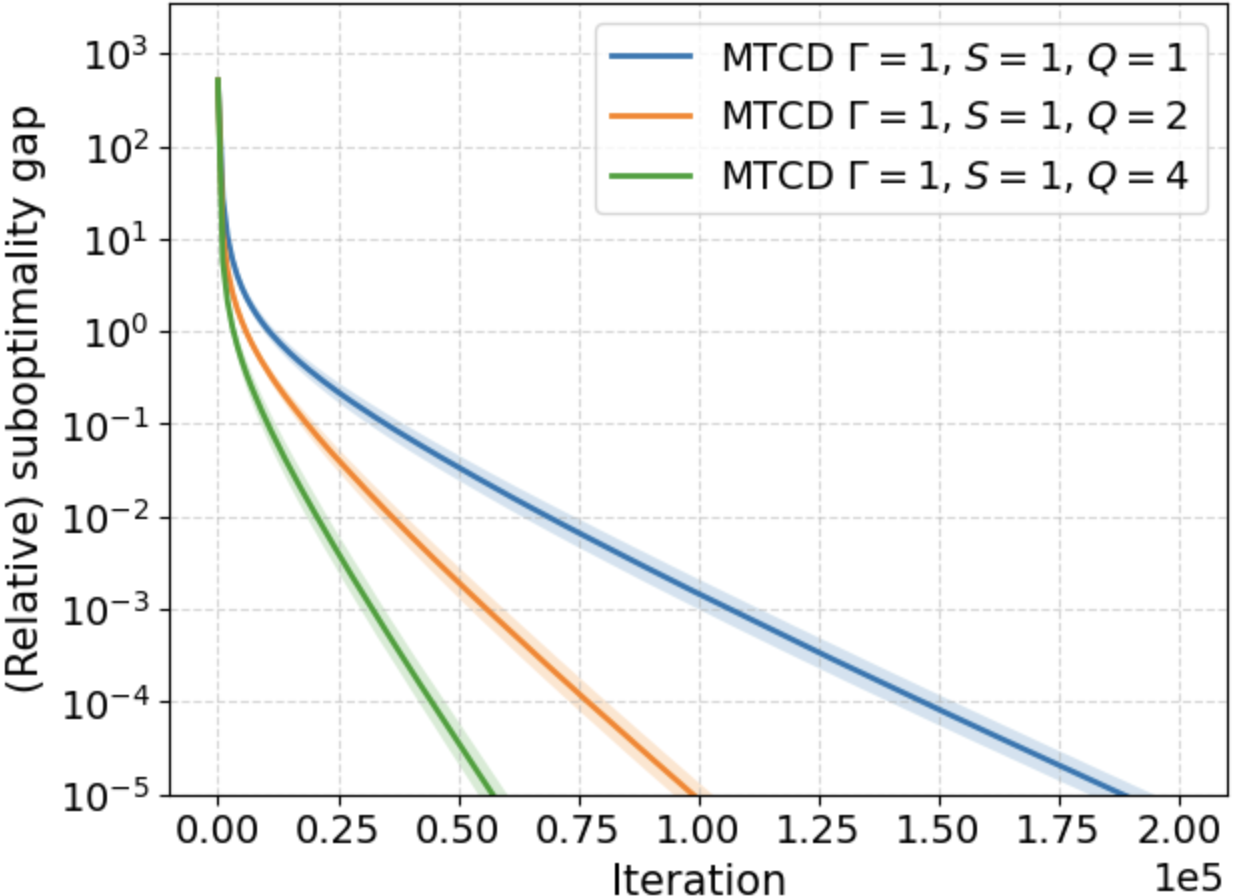}%
			\label{fig:mtcd_Q_G1S1}}
		
		\subfloat[$K=40$ path experiments, $\Gamma=2$, $S=1$, varying $Q$.]{\includegraphics[width=0.24\textwidth]{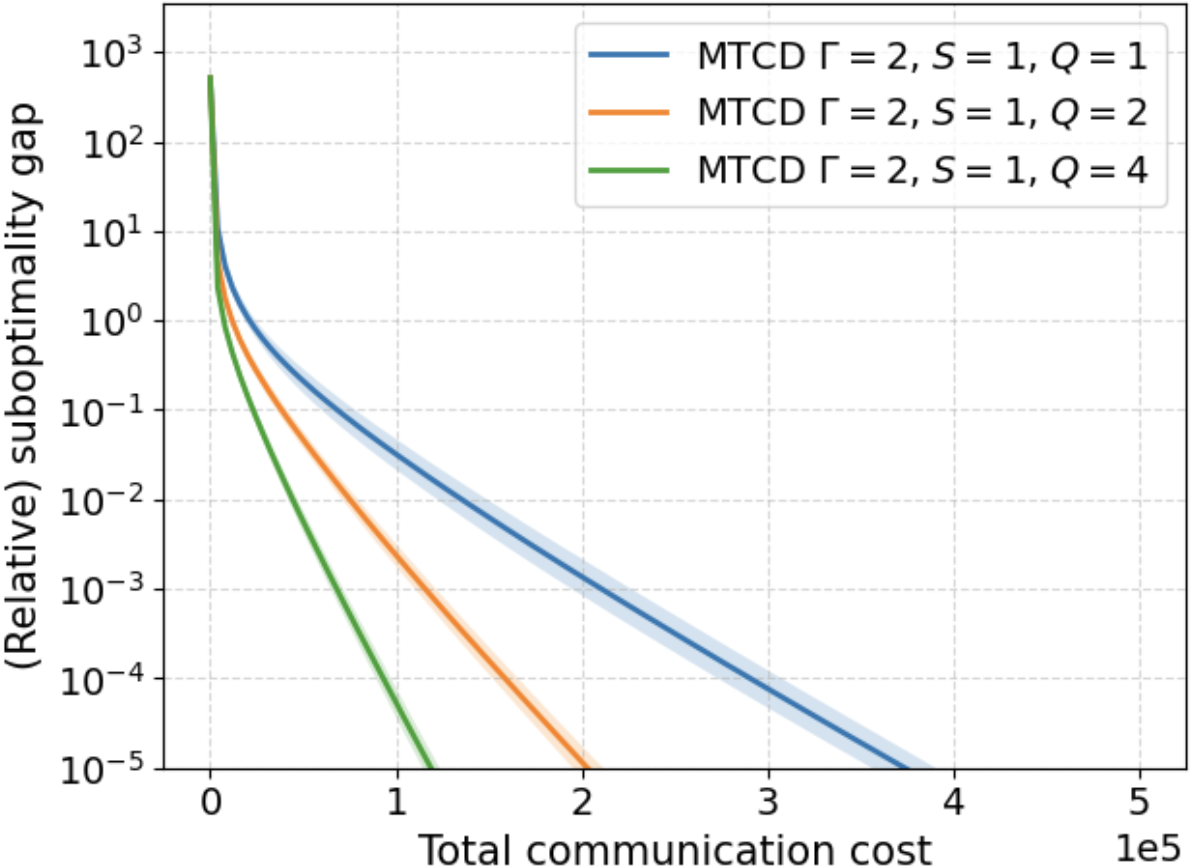}%
			\includegraphics[width=0.24\textwidth]{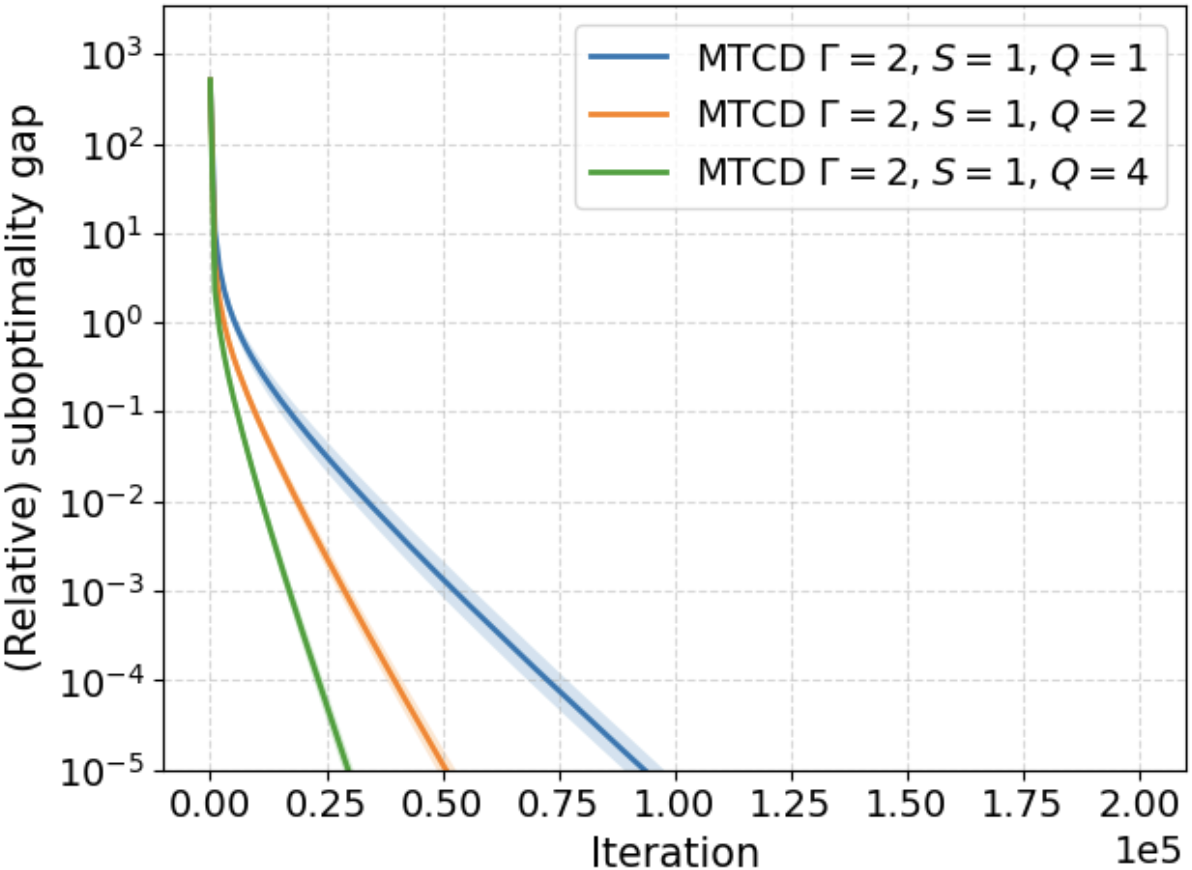}%
			\label{fig:mtcd_Q_G2S1}}
		
		\subfloat[$K=40$ path experiments with $\Gamma=4$, $S=1$, varying $Q$.]{\includegraphics[width=0.24\textwidth]{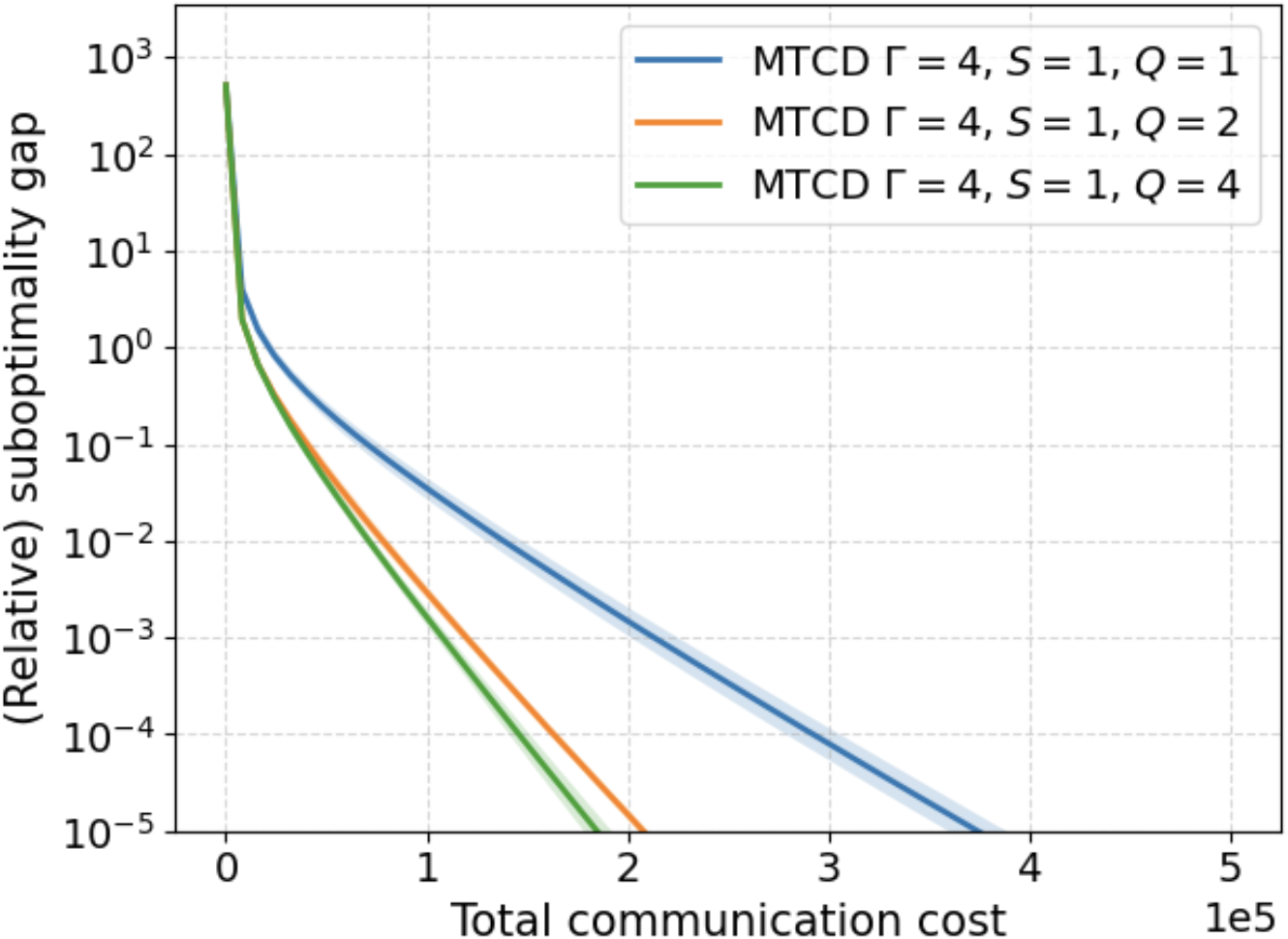}%
			\includegraphics[width=0.24\textwidth]{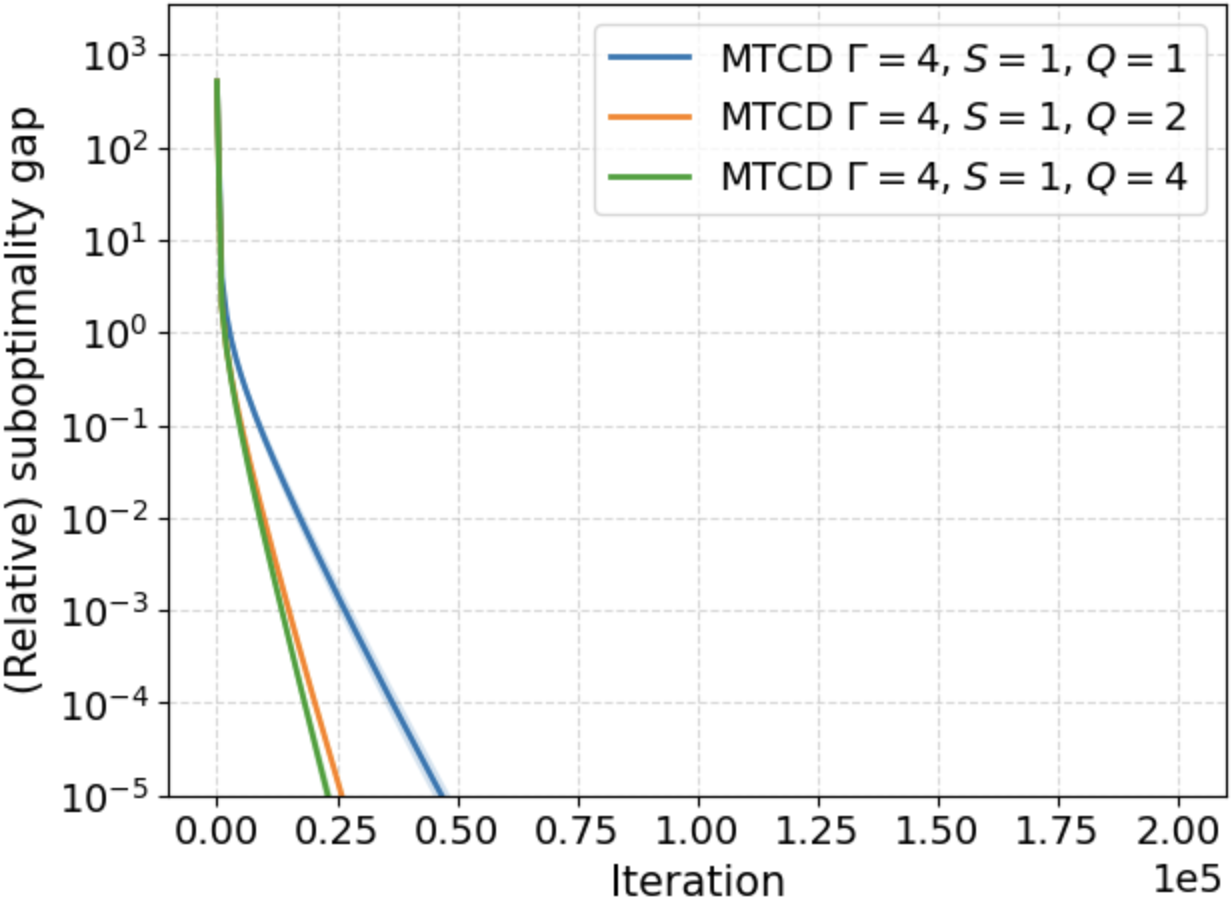}%
			\label{fig:mtcd_Q_G4S1}}
		
		\caption{$K=40$ path experiments with $S=1$ and $\Gamma\in\{1,2,4\}$, varying $Q$. Each row shows (left) communication cost and (right) iterations.}
		\label{fig:mtcd_Q_S1_allGamma}
	\end{figure}

	{\section{Lower bounding $\pi$} \label{app:lb-pi}
		As mentioned earlier, we can achieve a positive lower bound for $\pi>0$ through a variety of communication schemes. Some examples of such schemes include:
		\begin{itemize}[leftmargin=1.0em]
			\item If all clients have a probability $p_0$ of receiving the token directly from the server, then~\eqref{eq:min_prob_token_visit} holds for $\pi=p_0$ for any $P\geq1$, since, for all $c\in[C]$, $j\in[K_c]$, and $t = 0,\dots,T-1$:
			\[
			{\textstyle
				\mathbb{P}\left(\bigcup_{p=0}^{P-1} \{i_c^{t,p}=j\}\right) \geq
				\mathbb{P}\left( i_c^{t,0}=j \right)
				=p_0.
			}
			\]
			\item If at least one client $i$ in each cluster $c$ has a probability $p_0$ of receiving the token directly from the server and $S$ is greater than or equal to the diameter---the greatest distance between any pair of nodes in a graph---of the cluster. Then, it follows from the lazy random walk of the token that, for all $c\in[C]$, $j\in[K_c]$, and $t = 0,\dots,T-1$:
			\[
			{\textstyle
				\mathbb{P}\left(\bigcup_{p=0}^{P-1} \{i_c^{t,p}=j\}\right)
				\geq
				p_0 \cdot (1+d_{\max})^{-S}
				,
			}
			\]
			where $d_{\max}$ is the maximum node degree across all clusters. Thus, we have that~\eqref{eq:min_prob_token_visit} holds for $\pi=p_0 \cdot \frac{1}{(1+d_{\max})^S}$.
			\item We can also leverage the fact that the token roaming follows a Markov chain to arrive at a tighter bound than the previous one. In particular, if $S$ is greater than the mixing time of the Markov chain, if we make some mild assumptions on the associated transition matrix, we can leverage the results in~\citet{paulin2015} to drop the exponential dependency on $S$ above.
	\end{itemize}}
	
	\section{Ablation studies} \label{app:ablations}
	In Figure~\ref{fig:mtcd_G_Q1_allS}, Figure~\ref{fig:mtcd_S_Q1_allG}, and Figure~\ref{fig:mtcd_Q_S1_allGamma}, we show how key algorithmic parameters influence the performance of \texttt{MTCD}. In general, these results show us that, as we increase $\Gamma$, $S$, and $Q$, we observe a speedup, until a saturation point is reached, due to staleness.
	Note that, in Figure~\ref{fig:mtcd_S_Q1_allG}, although we do not get a faster convergence per iteration as we increase $S$, that is simply because (recall) for the experiments, we define iteration as a hop (for consistency with the fully decentralized version of the method), so in fact this means that doubling $S$ without seeing a speedup corresponds to a speedup according to our analytical definition of iteration. This improvement can further be seen in the convergence per communication cost plot.
		
\end{document}